\documentclass[journal]{IEEEtran}
\usepackage{amsmath,amsfonts}
\usepackage{algorithmic}
\usepackage{algorithm}
\usepackage{array}
\usepackage[caption=false,font=normalsize,labelfont=sf,textfont=sf]{subfig}
\usepackage{textcomp}
\usepackage{stfloats}
\usepackage{url}
\usepackage{verbatim}
\usepackage{graphicx}
\usepackage{cite}
\usepackage{multirow}
\usepackage{multicol}
\usepackage{longtable}
\usepackage{caption} 
\usepackage{forest}
\usepackage{tikz}
\forestset{parent color/.style args={#1}{
    {fill=#1},
    for tree={fill/.wrap pgfmath arg={#1!##1}{1/level()*80},draw=#1!80!darkgray}},
    root color/.style args={#1}{fill={{#1!60!gray!25},draw=#1!80!darkgray}}
}

\usepackage{xurl} 
\PassOptionsToPackage{hyphens}{url}\usepackage{hyperref} 

\allowdisplaybreaks 

\def\BibTeX{{\rm B\kern-.05em{\sc i\kern-.025em b}\kern-.08em
    T\kern-.1667em\lower.7ex\hbox{E}\kern-.125emX}}

\begin{document}

\title{Deep Crowd Anomaly Detection: State-of-the-Art, Challenges, and Future Research Directions}

\author{Md. Haidar Sharif, Lei Jiao, Christian W. Omlin \\~\IEEEmembership{Department of ICT, University of Agder,  Norway}

\thanks{Manuscript received \today; revised Month Day, 2022.}}

\markboth{Sharif \MakeLowercase{\textit{et al.}}: Deep Crowd Anomaly Detection: State-of-the-Art, Challenges, and Future Research Directions}
{Sharif \MakeLowercase{\textit{et al.}}: Deep Crowd Anomaly Detection: State-of-the-Art, Challenges, and Future Research Directions}

\maketitle

\begin{abstract}
Crowd anomaly detection is one of the most popular topics in computer vision in the context of smart cities. A plethora of deep learning methods have been proposed that generally outperform other machine learning solutions. Our review primarily discusses algorithms that were published in mainstream conferences and journals between 2020 and 2022. We present datasets that are typically used for benchmarking, produce a taxonomy of the developed algorithms, and discuss and compare their performances. Our main findings are that the heterogeneities of pre-trained convolutional models have a negligible impact on crowd video anomaly detection performance. We conclude our discussion with fruitful directions for future research.
\end{abstract}


\section{Used Acronyms}
\small\footnotesize
\begin{align}
Abbreviation &\Rightarrow  \emph{Expansion}  \nonumber\\
\emph{NA} &\Rightarrow \emph{Not Available or No Answer from the authors} \nonumber\\
\emph{BN} &\Rightarrow \emph{Batch Normalization} \nonumber\\
\emph{ReLU}   &\Rightarrow  \emph{Rectified Linear Unit}  \nonumber\\
\emph{CNN}  &\Rightarrow \emph{Convolutional Neural Network} \nonumber\\
\emph{AE}   &\Rightarrow \emph{Autoencoder} \nonumber\\
\emph{DAE}  &\Rightarrow \emph{Denoising Autoencoder} \nonumber\\
\emph{VAE}  &\Rightarrow \emph{Variational Autoencoder} \nonumber\\
\emph{AEVB} &\Rightarrow \emph{AutoEncoding Variational Bayesian} \nonumber\\
\emph{DpAE} &\Rightarrow \emph{Deep Autoencoder} \nonumber\\
\emph{AAE}  &\Rightarrow \emph{Attention-Based Autoencoder} \nonumber\\
\emph{2D-CAE}  &\Rightarrow \emph{Two-Dimensional Convolutional Autoencoder} \nonumber\\
\emph{3D-CAE}  &\Rightarrow \emph{Three-Dimensional Convolutional Autoencoder} \nonumber\\
\emph{TS-AE}  &\Rightarrow \emph{Two-Stream-Based Autoencoder} \nonumber\\
\emph{MoAE}     &\Rightarrow \emph{Motion Autoencoder} \nonumber\\
\emph{A3D-CAE} &\Rightarrow \emph{Adversarial 3D-CAE} \nonumber\\
\emph{CdAE}    &\Rightarrow \emph{Cascaded Autoencoder} \nonumber\\
\emph{RNN}  &\Rightarrow  \emph{Recurrent Neural Network} \nonumber\\
\emph{RNN-AE}  &\Rightarrow \emph{RNN-Based Autoencoder} \nonumber\\
\emph{AN}   &\Rightarrow \emph{Attention Network} \nonumber\\
\emph{CAE} &\Rightarrow \emph{Convolutional Autoencoder} \nonumber\\
\emph{TH-AE} &\Rightarrow \emph{Top-Heavy autoencoder} \nonumber\\
\emph{ST-U-Net}  &\Rightarrow \emph{Spatiotemporal U-Net} \nonumber\\
\emph{VQ-U-Net}  &\Rightarrow \emph{Vector-quantized U-Net} \nonumber\\
\emph{TS-AE} &\Rightarrow \emph{Two-Stream-Based Autoencoder} \nonumber\\
\emph{LSTM} &\Rightarrow  \emph{Long Short-Term Memory}  \nonumber\\
\emph{ConvLSTM} &\Rightarrow  \emph{Convolutional LSTM}  \nonumber\\
\emph{GCN}  &\Rightarrow  \emph{Graph Convolutional Network} \nonumber\\
\emph{Q-CNN}  &\Rightarrow  \emph{Quantum Convolutional Neural Network} \nonumber\\
\emph{GAN}  &\Rightarrow  \emph{Generative Adversarial Network}  \nonumber\\
\emph{C-GAN}  &\Rightarrow  \emph{Conditional GAN}  \nonumber\\
\emph{NM-GAN}  &\Rightarrow  \emph{Noise-Modulated GAN}  \nonumber\\
\emph{DE-GAN}  &\Rightarrow  \emph{Double-Encoder GAN}  \nonumber\\
\emph{BR-GAN}  &\Rightarrow  \emph{Bidirectional Retrospective GAN}  \nonumber\\
\emph{RPN}  &\Rightarrow  \emph{Region Proposal Network}  \nonumber\\
\emph{TNN}  &\Rightarrow  \emph{Transformer Neural Network}  \nonumber\\
\emph{MLAD} &\Rightarrow  \emph{Multi-Level Anomaly Detector}  \nonumber\\
\emph{YOLO} &\Rightarrow  \emph{You Only Look Once}  \nonumber\\
\emph{HTM}  &\Rightarrow  \emph{Hierarchical Temporal Memory}  \nonumber\\
\emph{SFA}  &\Rightarrow  \emph{Scale Factor Attentive}  \nonumber\\
\emph{MLP}  &\Rightarrow  \emph{Multi-Layer Perceptron} \nonumber\\
\emph{DAE}  &\Rightarrow  \emph{Denoising Autoencoder} \nonumber\\
\emph{SCN}  &\Rightarrow  \emph{Sparse Coding Network}  \nonumber\\
\emph{3DCNN}  &\Rightarrow  \emph{Three-Dimensional CNN} \nonumber\\
\emph{RCNN }  &\Rightarrow  \emph{Recurrent CNN}  \nonumber\\
\emph{DenseNet} &\Rightarrow  \emph{Dense Convolutional Network}  \nonumber\\
\emph{ResNet}   &\Rightarrow  \emph{Residual Network}  \nonumber\\
\emph{VGGNet}   &\Rightarrow  \emph{Visual Geometry Group Network}  \nonumber\\
\emph{SVM} &\Rightarrow  \emph{Support Vector Machine}  \nonumber\\
\emph{OCSVM}  &\Rightarrow  \emph{One-Class SVM}  \nonumber\\
\emph{SGD} &\Rightarrow  \emph{Stochastic Gradient Descent}  \nonumber\\
\emph{GRU} &\Rightarrow  \emph{Gated Recurrent Unit}  \nonumber\\
\emph{ConvGRU} &\Rightarrow  \emph{Convolutional GRU}  \nonumber\\
\emph{PSNR} &\Rightarrow  \emph{Peak Signal-to-Noise Ratio}  \nonumber\\
\emph{CPU} &\Rightarrow  \emph{Central Processing Unit}  \nonumber\\
\emph{GPU} &\Rightarrow  \emph{Graphics Processing Unit}  \nonumber\\
\emph{ACC}  &\Rightarrow  \emph{Accuracy}  \nonumber\\
\emph{ROC}  &\Rightarrow  \emph{Receiver Operating Characteristic}  \nonumber\\
\emph{AUC} &\Rightarrow  \emph{Area Under the ROC Curve}  \nonumber\\
\emph{F-AUC} &\Rightarrow  \emph{Frame-Level AUC}  \nonumber\\
\emph{P-AUC} &\Rightarrow  \emph{Pixel-Level AUC}  \nonumber\\
\emph{EER}  &\Rightarrow  \emph{Equal Error Rate}  \nonumber\\
\emph{F-EER} &\Rightarrow  \emph{Frame-Level EER}  \nonumber\\
\emph{P-EER} &\Rightarrow  \emph{Pixel-Level EER}  \nonumber\\
\emph{EDR} &\Rightarrow  \emph{Equal Detected Rate}  \nonumber\\
\emph{RTM}  &\Rightarrow  \emph{Run Time}  \nonumber\\
\emph{PRS}  &\Rightarrow  \emph{Precision Score}  \nonumber\\
\emph{RES} &\Rightarrow  \emph{Recall Score}  \nonumber\\
\emph{F1S} &\Rightarrow  \emph{F1 Score} \nonumber\\
\emph{IoU}  &\Rightarrow  \emph{Intersection over Union} \nonumber\\
\emph{APD}  &\Rightarrow  \emph{Average Precision Delay} \nonumber\\
\emph{mAP}  &\Rightarrow  \emph{Mean Average Precision} \nonumber\\
\emph{ELR}  &\Rightarrow  \emph{Event-Level Rating} \nonumber\\
\emph{PLR}   &\Rightarrow  \emph{Pixel-Level Rating} \nonumber\\
\emph{DPLR}  &\Rightarrow  \emph{Dual PLR} \nonumber\\
\emph{RBDR}  &\Rightarrow  \emph{Region-Based Detection Rate} \nonumber\\
\emph{TBDR}  &\Rightarrow  \emph{Track-Based Detection Rate} \nonumber\\
\emph{MSE}   &\Rightarrow  \emph{Mean Square Error} \nonumber\\
\emph{MAE}  &\Rightarrow  \emph{Mean Absolute Error} \nonumber\\
\emph{MISE}  &\Rightarrow  \emph{Misclassification Severity} \nonumber\\
\emph{AOS}  &\Rightarrow  \emph{Anomaly Score} \nonumber\\
\emph{RGS}  &\Rightarrow  \emph{Regularity Score} \nonumber\\
\emph{FAR}  &\Rightarrow  \emph{False Alarm Rate} \nonumber\\
\emph{SGAP}  &\Rightarrow  \emph{Score Gap} \nonumber\\
\emph{DOF}  &\Rightarrow  \emph{Degrees of Freedom} \nonumber\\
\emph{$d\prime$}     &\Rightarrow  \emph{Decidability Index} \nonumber\\
\emph{S4}   &\Rightarrow  \emph{S4 Score} \nonumber\\
\emph{fps}  &\Rightarrow  \emph{Frames per Second} \nonumber\\
\eta &\Rightarrow  \emph{Learning Rate} \nonumber\\
\chi^2 &\Rightarrow  \emph{Chi-squared Frequency Distribution} \nonumber\\
\sigma  &\Rightarrow  \emph{Standard Deviation} \nonumber
\end{align}
\normalsize

\section{Introduction}\label{Introduction}
\IEEEPARstart{T}{he} number of closed-circuit television surveillance cameras deployed in urban environments was estimated to have surpassed 1 billion in 2021~\cite{BahramiPVS21}. Contemporary automated crowd surveillance requires both privacy preservation and automation, and deep learning has become the method of choice for both~\cite{Sharif21,ZhaoCZ19,KassaniKWSD19}.
What constitutes an anomaly greatly depends on the context. For instance, people running out of a bank would typically be considered an anomaly, whereas a group of running people would be considered normal during a marathon. Furthermore, anomalies are few and far between, which means a lack of labeled data is a further challenge. Thus, the detection of anomalies is typically treated as an unsupervised rather than supervised learning problem: a model (e.g., an autoencoder) is trained on a video frame sequence to capture what is essentially normal activity, and any divergence is viewed as an anomaly~\cite{RodriguesEtAlcorr2019}.

Before the advent of deep learning models, researchers typically focused on the extraction of handcrafted spatiotemporal features and traditional image processing techniques~\cite{IhaddadeneSD08,SharifD09,FontugneHF08}. Although deep learning has made feature extraction by and large superfluous, there are remaining challenges related to anomaly inconsistency, the cost of obtaining labeled data, and the need for robustness in the face of multiple view angles and varying illumination and weather conditions~\cite{JaouediBB20,DoshiY20a,SindagiEtAl2020,RezaeeEtAl2021,YuanWZS22}.

Several surveys regarding crowd anomaly detection methods already exist. For example,
Borja et al.~\cite{BorjaEtAl2018} presented a short review of deep learning methods aimed at understanding group and crowd behaviors. Similarly, Afiq et al.~\cite{AfiqZSNKFJGIF19} offered a review of algorithms published between 2013 and 2018.
Khan et al.~\cite{KhanAKQD20} summarized seminal research works on crowd management published between 2010 and 2020.
Suarez et al.~\cite{SuarezEtAlcorr2020}  compiled a survey of deep learning solutions for anomaly detection in surveillance videos, considering articles published between 2016 and 2020.
Braham et al.~\cite{BrahamEtAl2021} performed a comparative study of crowd analysis algorithms, taking into account some previous reviews and various studies published between 2017 and 2020.
Elbishlawi et al.~\cite{ElbishlawiAESM20} assembled deep learning-based methods for crowd scene analysis methods that were published up to the time of writing (until 2020).
Mohammadi et al.~\cite{MohammadiEtAl2021corr}  conducted an in-depth literature survey into deep learning-based anomaly detection methods for both images and video (the studies were mainly published between 2010 and 2020).
Mu et al.~\cite{MuSYW21ReviewPaper} focused their review on deep learning methods published until 2020.
Sharma et al.~\cite{SharmaEtAl2021}  contrasted the deep learning literature published between 2017 and 2020 with earlier research published between 2011 and 2017 by studying 93 research articles from reputed databases published between 2011 and 2020.
Sanchez et al.~\cite{SanchezHTH20} reviewed articles on deep learning-based models published between 2011 and 2019.
Yuan et al.~\cite{YuanWZS22}  chronicled the evolution of human behavior-recognition methods, starting from standard manual feature extraction to deep learning methods.
Rezaee et al.~\cite{RezaeeEtAl2021} focused their survey on methods for secure distributed video analytics published between 2016 and 2020. Currently, deep learning algorithms are the standard approach for automatic crowd anomaly detection, with recent progress including the transition from two- to three-dimensional convolutional neural networks, the introduction of the attention mechanisms~\cite{ZhangWHWW21} now used in transformers~\cite{VaswaniSPUJGKP17,DosovitskiyEtAlCorr2020,TouvronCDMSJ21}, and excursions into the application of quantum computing-based machine learning~\cite{SchuldSP14,Tang2018corr,BlekosK21}.
While the aforementioned surveys were comprehensive and offered useful insights into open problems, they cannot bear witness to the tremendous advances that have been made since 2020. In our paper, we offer a survey that focuses on the literature published between 2020 and 2022. Our contribution can be summarized as follows:

\begin{itemize}
  \item A summary, taxonomy, and comparative study of the performance indicators of contemporary deep models;
  \item A collation of crowd datasets to gauge their popularity and relevance;
  \item A rigorous statistical analysis of data taken from experimental setups to assess the impact of the architectural heterogeneities of pre-trained two-dimensional convolutional neural networks.
\end{itemize}

The rest of the survey is organized as follows:
Section~\ref{EarlyWorks} summarizes several seminal works up to the period where our review begins.
Section~\ref{VariousFeatureSelectionMethods} discusses common crowd features and their extraction methods.
Section~\ref{PopularPerformanceMetricsforCrowdAnomalyDetection} compares the performance metrics of various crowd anomaly detection models.
Section~\ref{DatasetBriefDescriptionCrowdDatasets} briefly summarizes crowd datasets.
Section~\ref{DeepMethodsForCrowdAnomalyDetection} compiles a succinct survey of deep crowd anomaly detection methods.
Section~\ref{ArchImpactsOf2DCNNOnCrowdAnomalyDetection} illustrates the architectural impacts of pre-trained two-dimensional convolutional neural network models on the performances of crowd anomaly detection methods.
Section~\ref{FutureProspectResearchChallenge} identifies current research challenges and future prospects. Finally,
Section~\ref{Conclusion} concludes the paper.

\section{Early Works} \label{EarlyWorks}
This section summarizes the most influential early crowd anomaly detection methods.

\begin{figure*}
\centering
\includegraphics[width=0.89\textwidth]{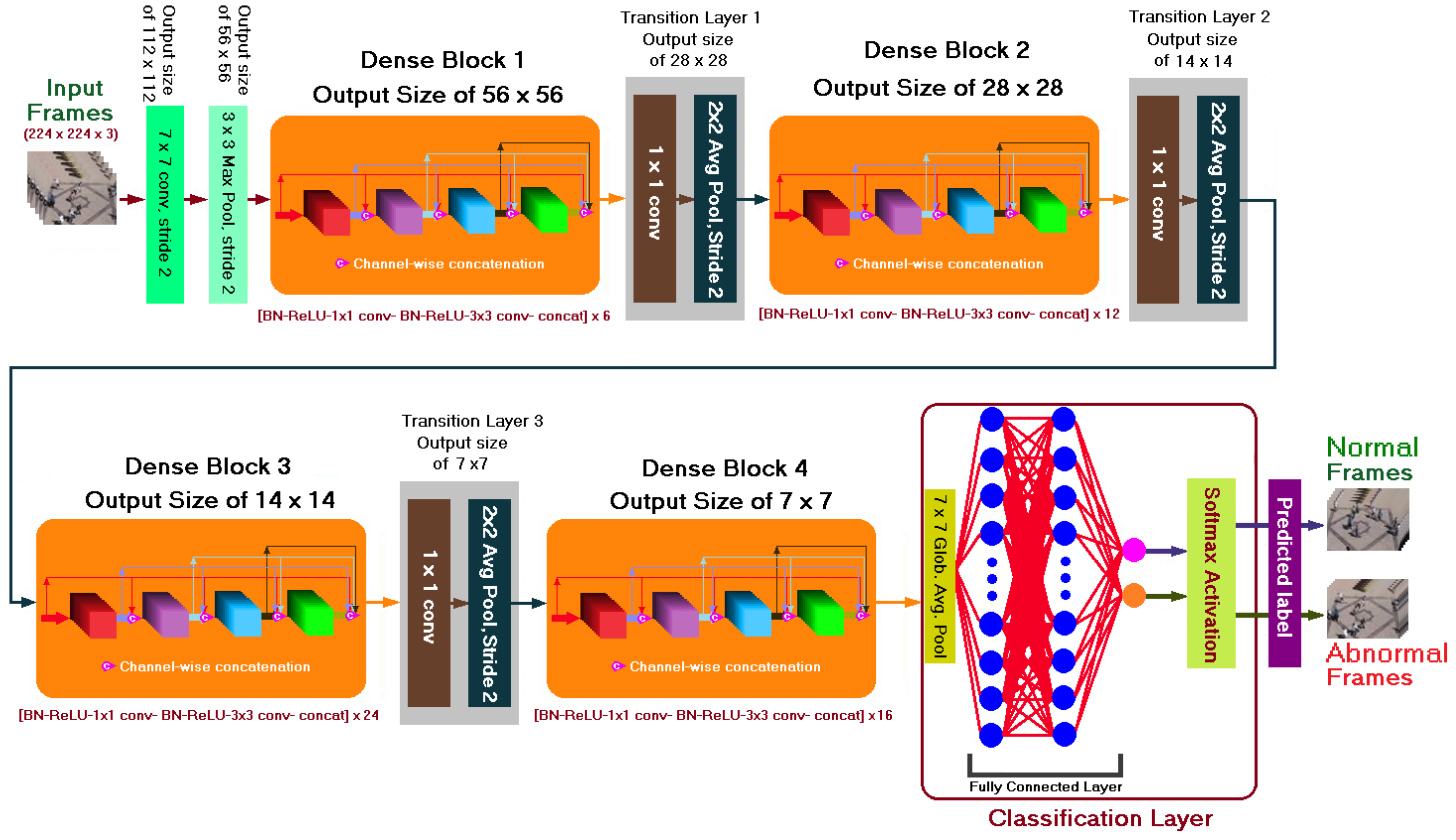}
\caption{Theoretical view of normal--abnormal frame classification using UMN~\cite{UMNdataset2021} dataset and DenseNet121~\cite{HuangLMW17,WakiliSSSUKIU2022}.}
\label{ArchDensenet121fc}
\end{figure*}

Detecting the abnormal activities or anomalies of a crowd in real-world surveillance videos is considered an important yet challenging task because prior knowledge about anomalies is usually very limited or unavailable.  Moreover, in real-life situations, both abnormal events and behaviors may take place at once in the crowd, and they both need to be detected.
Fundamentally, the crowd anomaly detection problem is a binary classification task where each frame in the video obtains an anomaly score for differentiating whether the frame belongs to a normal or abnormal class (e.g., see Figure~\ref{ArchDensenet121fc}).

No generic definition currently exists for abnormal events, which are usually dependent on the scene under consideration. For example, a car passing on the road is a normal activity but it becomes abnormal when it passes in the pedestrian lane. A typical approach for addressing such a scene's context dependency is to consider a scene's rare or unforeseen events as abnormal. 
Nevertheless, this may result in classifying unseen normal activities as abnormal. In general, it may not be possible to know all the normal and abnormal activities during training. It is only possible to have access to subsets of normal and abnormal activities. The lack of a generic definition and insufficiency in the data make it extremely hard for any learning algorithm to understand and capture the nature of an anomaly.
The results of the related literature report some success stories along with several convincing studies, which are mostly conducted in constrained conditions. Under uncontrolled scenarios, the task of crowd anomaly detection is still challenging for the research community. However, early research can be classified into the following two categories--- Deep learning paradigm  and Miscellaneous paradigm.

\subsection{Deep Learning Paradigm} 
Remarkable progress has been achieved in object-level information and deep learning models for tracking, classification, and clustering, which have been applied for recognizing abnormal events in video scenes. 
Currently, deep learning (e.g., CNN~\cite{LeCunEtAl1989}, AE~\cite{Vincent11}, GAN~\cite{GoodfellowPMXWOCB14}, LSTM~\cite{HochreiterS97}, U-Net~\cite{RonnebergerFB15}, YOLO~\cite{RedmonDGF16}, etc.)-based models are utilized as first-hand alternatives for crowd anomaly detection setups.
The primary strength of deep models is automatic salient feature extraction; however, they encounter
many challenges, namely application levels, such as when the anomalies have an inconsistent abnormal behavior and when the high cost of labeling data makes it difficult to obtain the large-scale labeled data required for training. Furthermore, crowd anomaly detection algorithms should work reliably and robustly under a wide range of changing circumstances, including varying levels of illumination, multiple view angles, and changing seasons and weather, and a single algorithm may not be the best option for all usage cases. 
Various methods have been proposed in the literature to improve efficiency, robustness against pixel occlusion, generalizability, computational complexity, and execution time~\cite{RezaeeEtAl2021}. Accordingly, a number of surveys (e.g., \cite{BorjaEtAl2018,AfiqZSNKFJGIF19,KhanAKQD20,SuarezEtAlcorr2020,BrahamEtAl2021,ElbishlawiAESM20,MohammadiEtAl2021corr,
MuSYW21ReviewPaper,SharmaEtAl2021,SanchezHTH20,YuanWZS22,RezaeeEtAl2021}) on crowd anomaly detection methods explain taxonomy, anomaly detection, crowd emotion, datasets, opportunities, and prospects.

\subsection{Miscellaneous Paradigm} 
In spite of the successes of deep learning models, some researchers predominantly focused on the use of dissimilar handcrafted spatiotemporal features and image processing procedures. For example, the Gaussian Mixture (GMM), Hidden Markov (HMM), and Gaussian Mixture Hidden Markov Models (GM-HMM), One-Class Support Vector Machines (OSVMs), and Bag-of-Words (BoW) techniques were commonly applied to model normal behavior patterns for detecting abnormal patterns~\cite{AfiqZSNKFJGIF19}. The Kanade--Lucas--Tomasi feature tracker (KLT), Multi-Observation HMM, k-means clustering, Principal Component Analysis Histogram of Oriented Gradient (PCA-HOG), and Histogram of Oriented Optical Flow~(HOFH) techniques were also applied to detect abnormal behaviors in crowded scenes~\cite{ZhouSZZ15,YibinEtAl2015,WangMLLZ16}. In the approaches based on spatiotemporal volume along with cuboids~\cite{ChenEtAl2016}, in addition to spatiotemporal textures (STT)~\cite{WangX16cviu}, trajectories~\cite{TranYF14}, interest points (STIP)~\cite{ChengCF15}, descriptors (STD)~\cite{LungJP15}, and texture maps (STTM)~\cite{LloydRMM17}, the features from the spatial and temporal dimensions were integrated for the purpose of crowd tracking and anomaly detection.
The GMM techniques were trained by the expectation--maximization (EM) algorithm. The EM fairly well computed the GMM parameters. However, it required a high computational cost. In the HMM approach, the computing process was based on the initial probability of hidden states and transition and observation matrices~\cite{AfiqZSNKFJGIF19}. 
The optical flow was considered to obtain the crowd features alongside important information regarding various behaviors or activities.  Both optical flow and spatiotemporal techniques played a major role in abnormal event detection frameworks.

With the advent of new deep learning models, crowd anomaly detection prototypes have shifted from traditional 2DCNN to 3DCNN, from standard deep learning to more intelligent attention-, transformer- or even quantum computing-based learning. One of the advantages of the existing surveys is that they represent widely recognized deep learning methods in such a manner that effectively highlights current issues. 
Although most of the surveys sufficiently discuss contemporary methods and materials with systematically deep discussion, they generally do not include the most recent advances in crowd anomaly detection.

To fill this gap, our paper aims to provide insight into the deep learning-based crowd anomaly detection methods published in mainstream English-language conferences and journals (but excluding Masters’ and doctoral theses, and unpublished) articles between 2020 and 2022.
We augment the knowledge of the anomaly detection research community. Our findings can also be adopted by newcomers to obtain an overall comprehension of recent progress in the field.

\section{Selection of Predictive Features} \label{VariousFeatureSelectionMethods}
Features are used for learning. They can also be the output of learning, derived from datasets.
Feature extraction is the task of removing appropriate information from raw data.
Extracting informative, discriminating, and independent features is a crucial element of effective anomaly detection algorithms.  The feature extraction quality directly influences the algorithm's detection accuracy~\cite{YuanWZS22}. Thus, feature extraction methods play a considerable role in crowd anomaly detection tasks.
The t-SNE (t-distributed Stochastic Neighbor Embedding) method is usually used as a dimensionality reduction tool for the visualization of high-dimensional feature distributions. It is widely applied in abnormal event detection tasks to visualize the distributions of learned features~\cite{ShinBC20,TsaiEtAlwacv2022}.

\subsection{Popular Crowd Features}
Crowd features are composed of a set of metrics that specify the dynamics, topological anatomy, and emotional state of the crowd. Those metrics can be maintained over time by computing at the individual and/or crowd levels.
\begin{itemize}
  \item \textbf{Motion Pattern}  $\Rightarrow$ The motion pattern is defined as the optical flow vector, which appeared as the brightness patterns in the images~\cite{AfiqZSNKFJGIF19}.
  \item \textbf{Trajectory} $\Rightarrow$ Trajectory features are very informative~\cite{ZhouM17}. The global motion pattern in a scene can be utilized to describe a trajectory~\cite{LinZXYXWL17}. In general, the KLT (Kanade--Lucas--Tomasi feature tracker) is utilized to obtain the trajectories~\cite{HariyonoJ17}.
  \item \textbf{Velocity} $\Rightarrow$ This measures the average speed at which individuals (when the approach is bottom-up) or crowds
  (in top-down approaches) are moving~\cite{ZhangZHGY18}.
  \item \textbf{Direction} $\Rightarrow$ At the macroscopic level, it determines the number of main directions of movement influencing the crowd~\cite{MusseJJB07}. The direction followed by each individual may also be extracted in microscopic approaches.
  \item \textbf{Density} $\Rightarrow$ This quantifies the proximity of individuals in the crowd, determining how dense the crowd is. At the macroscopic level, the objective is to perform density estimation rather than precise people counting, as clutter and severe occlusions make individual counting difficult in very dense crowds.
  \item \textbf{Collectiveness}  $\Rightarrow$ Individuals tend to follow the behaviors of others. This feature measures the degree that individuals act as a union in collective motions~\cite{ZhouTW13,ZhangKLWXY16,AfiqZSNKFJGIF19}. When part of a crowd,
      instead of behaving independently, individuals tend to follow the behaviors of others and move in the same direction
      as their neighbors~\cite{SanchezHTH20}.
  \item \textbf{Stability}  $\Rightarrow$  Crowd stability describes group tendencies, such as maintaining topological structures over time~\cite{AfiqZSNKFJGIF19}. 
  This is determined by the behavior of the group's members, which continues with the nearest neighbors at a consistent distance. 
  \item \textbf{Uniformity}  $\Rightarrow$ Group uniformity is calculated based on members' distances and evenly distributed locations in space. Furthermore, non-uniform groups tend to scatter in different directions~\cite{AfiqZSNKFJGIF19}.
  \item \textbf{Conflict} $\Rightarrow$ The conflict feature is used to illustrate the interaction or friction between people as they approach each other~\cite{AfiqZSNKFJGIF19}.
  \item \textbf{Valence}  $\Rightarrow$ This aims to measure the positive and negative effects of the crowd. According to the literature on Psychology, it is usually presented as a $[-1,~1]$ steady scale, ranging from unpleasantness to pleasantness~\cite{SanchezHTH20}.
  \item \textbf{Arousal}  $\Rightarrow$  This feature helps to monitor the tranquility and excitement of the crowd.  It is also presented in a $[-1,~1]$ steady scale, extending from passive to active~\cite{SanchezHTH20}.
\end{itemize}

\subsection{Manual Feature Extraction}
Conventional artificial feature extraction methods were repeatedly applied in the early development of human action recognition. Manual-feature-extraction-based methods can be categorized into several groups, including spatiotemporal interest points~\cite{Laptev05,ScovannerAS07}, template matching~\cite{YilmazS05,KlaserMS08}, trajectories~\cite{HuangEtAl2020}, and depth-sequences~\cite{YangZT12acm}. 
Nevertheless, due to changing illumination, viewing angles, and occlusion, conventional manual feature methods are no longer applicable in complex scenes~\cite{JaouediBB20,YuanWZS22}.

\subsection{Automatic Feature Extraction}
Deep learning can automatically extract feature vectors through iterative learning to conclude classification. 
For example, due to the grid-like nature of images, 
CNNs automatically extract salient features at different levels of abstraction.
CNN applies the feature extractor in the training process instead of manually implementing it. Usually, CNN consists of a convolution, sub-sampling (or pooling layer), and fully connected layers. The convolutional operation in the lower layers is utilized for analyzing the links between neighboring patches and learning the low-level local features~\cite{HeJSWLD17}. The sub-sampling layers are non-linear down-sampling operations that compute average or maximum values for each input image patch or feature map. 
Subsequently, they improve the robustness of translation but lessen the number of network parameters~\cite{AfiqZSNKFJGIF19}. Eventually, the outcomes of the fully connected layer include the learned feature, which can be applied for classification or detection tasks~\cite{LiCXC18}.

\subsection{Popular Feature Selection Methods}
Finding the most predictive features for a fixed deep learning model is important. Many methods exist to select features for supervised learning. The following methods can be adopted to select predictive features.
\begin{itemize}
\item \textbf{Filter Method} $\Rightarrow$ This category uses statistical tests that select features based on their distributions. Filter methods are much faster compared with wrapper methods because they do not involve training the models. Although these methods are computationally very fast, the p-values of statistical tests tend to be very small for big datasets. This indicates significant, yet tiny, differences in distributions. For this reason, they are not widely used in practice.
\item \textbf{Wrapper Method} $\Rightarrow$ These methods apply the concept of greedy algorithms that will try every possible feature combination based on a step forward, step backward, or exhaustive search. For each feature combination, these methods will train a machine learning model, usually with cross-validation, and determine its performance. Thus, wrapper methods are very computationally expensive, and often, impossible to carry out.
\item \textbf{Embedded Method} $\Rightarrow$ These methods train a single machine learning model and select features based on the feature importance returned by that model. They tend to work very well in practice and are faster to compute.
    On the downside, we cannot derive feature importance values from all machine learning models (e.g., nearest neighbors).
    Furthermore, decision tree-based algorithms may not perform well in very big feature spaces and, thus, the importance values might be unreliable. In brief, embedded methods are not suitable for every scenario or every machine learning model.
\item \textbf{Shuffling Method} $\Rightarrow$  The feature shuffling method assigns importance to a feature based on the decrease in a model performance score when the values of a single feature are randomly shuffled. It only trains one machine learning model, so it is quick and suitable for any supervised machine learning model. If two features are correlated, when one of the features is shuffled, the model will still have access to the information through its correlated variable. Feature shuffling is available in popular feature engines, including the Python open-source library.
\item \textbf{Thresholding Method} $\Rightarrow$  The thresholding method splits the data frame into a training and a testing set, and then it selects the features with performances above a threshold. It is fast because no machine learning model is trained, and it is also robust to outliers. It captures non-linear relationships between features and the target, which is model agnostic. Nevertheless, it requires the tuning of interval numbers for skewed variables. Rare categories will offer unreliable performance proxies or make the method impossible to compute.
\end{itemize}

\section{Performance Metrics for Crowd Anomaly Detection Models}\label{PopularPerformanceMetricsforCrowdAnomalyDetection}
Crowd anomaly detection functions fall into the category of an unbalanced binary classification task,
i.e., an AOS obtained by an anomaly detection model categorizes each frame in the video as either normal or abnormal.
In this section, we categorize the performance metrics for crowd anomaly detection and then discuss the
evaluation metrics, available from 2020 and 2022, in Table~\ref{SummaryOfLiteratureReview2020to2021}. 

\subsection{Classification of Performance Metrics}
The performance evaluation metrics for crowd anomaly detection can be roughly categorized as presented below.
\begin{itemize}
\item \textbf{Evaluation Metrics for Frame-level Detection} $\Rightarrow$ The motivation for crowd anomaly detection in the video is to come up with the AOS of each frame, which stipulates the probability that the frame holds abnormal events. The AOS has a range from 0 to 1~\cite{ChongT17}. It is used to explain the degree of anomaly. It is an absolute anomaly measured together with the temporal and spatial characteristics of a video~\cite{EsquivelG22}. A higher AOS value signifies a higher anomaly level~\cite{NayakPD21}. Commonly, the metrics of MSE~\cite{LuoLG17,GongLLSMVH19} and PSNR~\cite{MathieuCL15,LiuLLG18} are employed to calculate the AOS~\cite{ZhongCJR22}.
For example, the AOS of a frame $t$ (\emph{AOS$_t$}) can be computed by Equation~(\ref{EqAOS}) as~\cite{SunJSW21}:
     \begin{equation}\label{EqAOS}
            AOS_t = \frac{e_t - \min_t\,e_t}{\max_t\,e_t},
    \end{equation}
where $e_t$ belongs to the mean reconstruction error of all the pixel values in $t$, $\min_t\,e_t$ indicates the minimum reconstruction error among all frames in a video, and $\max_t\,e_t$ addresses the maximum frame-level reconstruction error in a video. Furthermore, AE~\cite{WangZLZZY18} or OCSVM~\cite{AzizBMZ21} models can be utilized to obtain an $AOS_t$. If the value of  $AOS_t$ is significant, the frame $t$ is categorized as an abnormal event. 
 While most existing anomaly detection algorithms calculate the AOS based on the current frame, a small amount of them use past, present, and future frames~\cite{ZhongCJR22}.
If any region of a frame is detected as an anomaly that is in line with the frame-level ground truth annotation, such detection is granted to be a correct hit regardless of the locality and the area of the region~\cite{ZhangYZZ20,YanSLZ20}.
The ACC, F-AUC, F1S, and F-EER  are well-known examples of this category.

\item \textbf{Evaluation Metrics for Pixel-Level Detection} $\Rightarrow$
Although frame-level evaluation methods have been adopted by scores of researchers, they cannot determine whether the anomaly location is accurate~\cite{ZhangMYHHS20}. To emphasize the correctness of the abnormal locality, pixel-level measurements can be used.
In this performance metric, the exact location of the abnormal event in the frame is marked as a mask. Anomaly detection algorithms are asked to create a comparable mask. An anomaly is said to be flawlessly recognized if the predicted mask is aligned with the ground truth. If over 40\% of the ground truth events are detected as anomalies in a frame, such detection is called a right detection~\cite{ZhangYZZ20,YanSLZ20,FanWLQLX20,MuSYS21,MuSYW21ReviewPaper,NayakPD21,WuLCH21}. This criterion can be employed to evaluate the anomaly localization capability~\cite{FanWLQLX20}. P-AUC and P-EER are common examples of this category.

\item \textbf{Evaluation Metrics for Event-Level Detection} $\Rightarrow$ If any position with a true anomaly is detected and localized as abnormal, such detection is granted a correct hit~\cite{ZhangYZZ20}. On the other hand, if any normal frame is detected as an anomaly, it is counted as a false alarm~\cite{KimG09,CongYL13}
    To reduce the noisy and incomprehensible local minima in the RGS, Hasan et al.~\cite{Hasan0003CNRD16} applied the Persistence1D~\cite{Persistence1D2020} algorithm to cluster local minima using a fixed temporal window of 50 frames. Plainly, local minima within 50 frames are in the same abnormal event~\cite{YanSLZ20}.
    To this end, in the CUHK-Avenue~\cite{LuSJ13} dataset, the spatial stream in~\cite{YanSLZ20} detected 36 abnormal events with 8 false alarms, while their temporal stream detected 32 abnormal events with 12 false alarms.

\item \textbf{Evaluation Metrics for Computational Complexity} $\Rightarrow$  It is often important to quantify the time and space complexities of a video anomaly detector. Here, RTM is commonly used. The number of operations executed by the model can be measured in FLOPS (floating-point operations per second)~\cite{Wu0S20}.
\end{itemize}

\subsection{Used Performance Metrics 2020-2022}
The following performance evaluation metrics were used between 2020 and 2022 for crows anomaly detection tasks.
\begin{itemize}
\item \textbf{ACC} $\Rightarrow$  It is a very familiar metric for binary classification problems. If $t_n$, $t_p$, $f_p$, and $f_n$ represent true negative, true positive, false positive, and  false negative, respectively, then  ACC can be computed using Equation~(\ref{accEq0}) as:
  \begin{equation}\label{accEq0}
    ACC = \frac{t_n+t_p}{t_p + t_n + f_p + f_n}.
  \end{equation}
  The major stumbling block of this metric is its paucity of unbalanced setups~\cite{SanchezHTH20}. Sometimes, the vocable ACC is applied interchangeably with the percent correct classification (PCC)~\cite{ShehuSSDTUKR21}.

  \item \textbf{PRS} $\Rightarrow$ It is defined as the number of correct positive results divided by the number of positive results predicted by the
  classifier. It can be computed using Equation~(\ref{PRSEq2}) as:
   \begin{equation}\label{PRSEq2}
     PRS = \frac{t_p}{t_p + f_p}.
   \end{equation}

  \item \textbf{RES} $\Rightarrow$ It computes the fraction of correctly classified elements that belong to the positive class. The recall is defined as the number of correct positive results divided by the number of all relevant samples.
      It can be computed using Equation~(\ref{RESEq3}) as:
   \begin{equation}\label{RESEq3}
     RES = \frac{t_p}{t_p+f_n}.
   \end{equation}

  \item \textbf{F1S} $\Rightarrow$  It is also called the F-score or F-measure. It is just the harmonic mean between precision and recall.
   It can be computed using Equation~(\ref{f1sEq4}) as:
   \begin{equation}\label{f1sEq4}
     F1S = \frac{2}{\frac{1}{PRS} + \frac{1}{RES}}= \frac{t_p}{t_p + \frac{f_p+f_n}{2}}.
   \end{equation}

  \item \textbf{AUC} $\Rightarrow$ True positive,  true negative, false positive, and false negative rates indicate that an anomalous frame is detected as anomalous, a normal frame is detected as normal, a normal frame is detected as anomalous, and an anomalous frame is detected as normal, respectively.
      The ROC curve is generated by plotting the true positive rate against the false positive rate at numerous threshold settings. The AUC is one of the most widely used metrics for evaluating flows and events in crowd videos~\cite{Sharif2017}. The AUC of a classifier equals the probability that the classifier ranks a randomly chosen positive sample higher than a randomly chosen negative sample~\cite{ShehuSSDTUKR21}. AUC = 0.00 indicates that the predictions of a model are 100\% wrong. Conversely, if the predictions are 100\% correct, then its AUC is just 1.
      AUC can be accomplished in both frame- and pixel-level evaluations.
      In a frame-level evaluation, an anomaly is measured at the frame level. A frame is counted as an anomaly even if an anomaly is detected for less than one pixel of the discrete frame~\cite{SabokrouEtAl2017}. 
      In a pixel-level evaluation, the evaluation scrutinizes each pixel independently~\cite{MuSYW21ReviewPaper}.
      However, many authors performed both F-AUC and P-AUC for evaluating anomaly detection systems~\cite{WuLCH21}.

  \item \textbf{EER} $\Rightarrow$  Originally, this performance metric was used in biometric systems. Currently, it is widely employed for crowd video anomaly detection. It is interpreted as the operating point at which the miss and false alarm rates are equal. It can be computed directly from the ROC curve as shown in Figure~\ref{EERcalulationFromROC}.
      \begin{figure}
       \centering
       \includegraphics[width=0.23\textwidth]{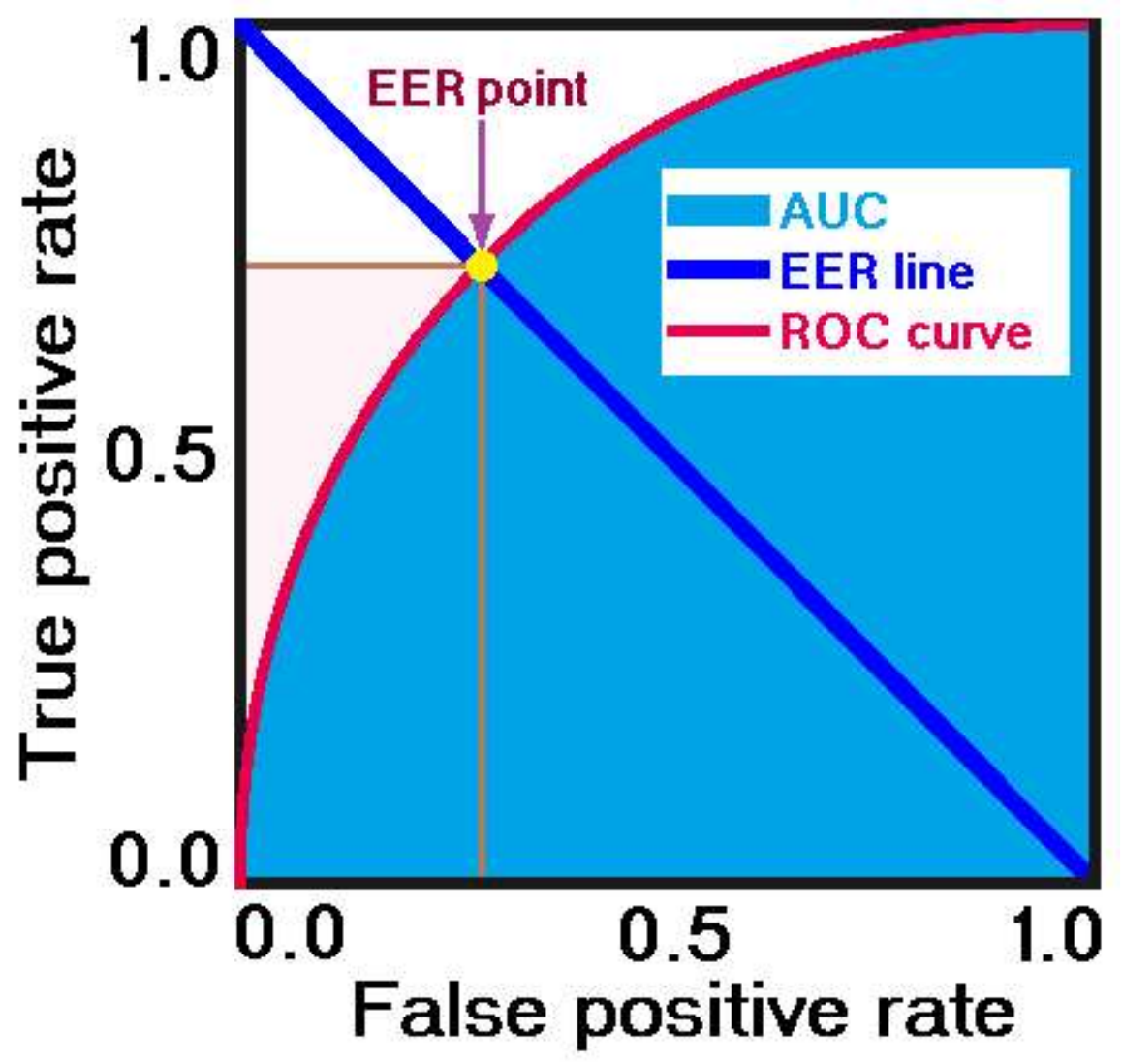}
        \caption{Estimation of EER from ROC curve.}
       \label{EERcalulationFromROC}
       \end{figure}
      An ERR is a point where false positive and true positive rates intersect.
      An EER can be accomplished at the frame or pixel level. Similar to AUC, many authors performed both F-EER and P-EER when evaluating anomaly detection systems~\cite{WuLCH21}.
      A detection algorithm with a lower EER is regarded as more accurate. EER is effective for the detection of video anomalies~\cite{Wu0S20,NayakPD21}. However,  EER can provide misleading results for anomaly detection~\cite{GiornoBH16,AsadYTCH21}.
      For example, Giorno et al.~\cite{GiornoBH16} identified every instance of an anomaly and also evaluated their fidelities compared with a human evaluation of anomalousness
      . They avoided metrics that scored anomaly detections in an event-detection manner, where labeling a single frame was considered a successful detection method of adjoining frames. They said~\cite{GiornoBH16}: ``\textit{Metrics like EER can be misleading in the anomaly detection setting. Consider the case when only 1\% of the video is anomalous: the EER on an algorithm that marks all frames normal would be 1\%, outperforming most modern algorithms. This extreme class imbalance is less prevalent in current standard datasets but will become an apparent problem as more realistic datasets become prevalent.}"

  \item \textbf{EDR} $\Rightarrow$  A higher EDR indicates a better performance. The EDR can be computed by Equation~(\ref{EqEDR}) as~\cite{LiMV14}:
     \begin{equation}\label{EqEDR}
      EDR = 1 - EER.
     \end{equation}

  \item \textbf{S4} $\Rightarrow$   The S4 limits between 0 and 1, and a higher score is always better, which can be computed by Equation~(\ref{EqS4}) as~\cite{DoshiY20a}:
     \begin{equation}\label{EqS4}
      S4 = (F1S)(1 -  \textit{Normalized root MSE}).
     \end{equation}

 \item \textbf{IoU} $\Rightarrow$ It is a measure of the magnitude of the intersection between two bounding boxes. It computes the size of the overlap between two objects divided by their combined total area. It can be computed via Equation~(\ref{IoUEq5}) as:
   \begin{equation}\label{IoUEq5}
     IoU = \frac{\emph{Area of overlap}}{\emph{Area of union}}.
   \end{equation}

 \item \textbf{APD} $\Rightarrow$ It is a  combination of the average detection delay (ADD) and alarm precision. The AUC metric summarizes the true positive versus the false positive rate, whereas APD measures the area under the precision versus normalized ADD curve~\cite{DoshiEtAlwacv2022a,DoshiEtAlwacv2022b}.

 \item \textbf{mAP} $\Rightarrow$ It is a popular metric used to measure the performance of models for document/information retrieval and object detection tasks. However, it can also be applied to anomaly explanation tasks~\cite{SzymanowiczEtAlwacv2022}.

 \item \textbf{DPLR} $\Rightarrow$ The PLR helps to evaluate the quality of a detection algorithm.
 The location of abnormal events is important for pixel-level evaluation. A frame with ground truth anomalies is considered as a true positive detection if at least 40\% of the ground truth anomalous pixels are detected. However, it suffers, because if 40\% of the abnormal ground truth events are overlapped, all falsely detected regions are ignored and then the system can give as many (false) detections as possible for covering the ground truth~\cite{WuLCH21}. 
 Nevertheless, a limitation can be addressed at pixel-level detection. 
     For example, a DPLR can be considered if a minimum of 5\% of the detected regions belong to the true anomalous pixels~\cite{WuLCH21}. If a substantial unconnected region is detected, it is not identified as a true positive by a DPLR~\cite{SabokrouFHK15,WuLCH21}.

\item \textbf{RBDR} $\Rightarrow$ Many anomaly detection studies reported the pixel-level AUC for some popular datasets (e.g., UCSD Ped2~\cite{ChanLV08}). However, Ramachandra et al.~\cite{RamachandraJ20} claimed that the pixel-level AUC is a flawed evaluation metric. Thus, they introduced RBDR and TBDR to replace the commonly used pixel- and frame-level AUC metrics. The region-based detection criterion estimates the RBDR over all frames in the test set versus the number of false positive regions per frame. A true positive takes place if a ground truth-annotated region has a minimum IoU of 0.1 with a detection region. The RBDR can be computed using Equation~(\ref{EqRBDR})~\cite{RamachandraJ20}:
    \begin{equation}\label{EqRBDR}
    \emph{RBDR} = \frac{\emph{Number of anomalous regions detected}}{\emph{Total number of anomalous regions}}.
   \end{equation}

\item \textbf{TBDR} $\Rightarrow$ The track-based detection criterion measures the TBDR versus the number of false positive regions per frame.
    A ground truth track is said to be detected if at least 10\%  of the ground truth regions in the track is detected.
    A ground truth region in a frame is said to be detected if the IOU between the ground truth region and a detected region is greater than or equal to 10\%.
    The total number of positives is the number of ground truth-annotated tracks in the testing dataset. The TBDR can be computed using Equation~(\ref{EqTBDR})~\cite{RamachandraJ20}:
   \begin{equation}\label{EqTBDR}
  \emph{TBDR} = \frac{\emph{Number of anomalous tracks detected}}{\emph{Total number of anomalous tracks}}.
  \end{equation}

\item \textbf{MISE} $\Rightarrow$  It estimates a smaller probability of classifying a high-level event to a low level. The MISE of a model determines if anomaly event classification can be analyzed quantitatively. A lower MISE implies a better anomaly detection performance~\cite{LinGWL21}.

\item \textbf{SGAP} $\Rightarrow$   The SGAP is computed by subtracting the average score of the normal from that of the anomaly~\cite{FengHZ21}. A larger SGAP hints that the anomaly detection model is more capable of distinguishing anomalies from normal events~\cite{LiuLLG18,ZhongCJR22}, while also enjoying stronger robustness to noises~\cite{ZhouZFDPX20}.

\item \textbf{RGS} $\Rightarrow$  The RGS can be deemed as the opposite of the AOS~\cite{LiuLLG18,NayakPD21}. Whether a frame contains anomalies can be judged by its RGS~\cite{HaoLWWG22}. In the testing phase, a detector touches on the RGS of an image. A lower RGS value signifies a higher anomaly level~\cite{NayakPD21}, where errors between a predicted frame based on the normal training data and the ground truth image are substantial~\cite{SaypadithO21}. The RGS curve drops steeply when an abnormal event occurs~\cite{HaoLWWG22}.

\item \textbf{FAR} $\Rightarrow$ In the video anomaly detection task, a higher AUC shows a better model performance, whereas a lower FAR on a normal video implies stronger anomaly-detection-method robustness~\cite{SultaniCS18,WanFXM20}.

\item \textbf{$d\prime$} $\Rightarrow$ Both decision and detection tasks involve some uncertainty. The $d\prime$ looks for squeezing out how inherently decidable the decision task is, or how detectable the signal is, regardless of the observer's error-avoidance preferences~\cite{Williams1996}.

\item \textbf{RTM} $\Rightarrow$    Basically, RTM depends on the input dimensions, the number of neurons and layers~\cite{NayakPD21}, the experimental setup, and the code organization~\cite{SharifBSH2008,Sharif14}. Preferably, RTM should be as low as possible for a desirable level of ACC.

\end{itemize}

\subsection{Performance Metrics Comparison}
The regression-problem-related tasks (e.g., crowd counting) mostly employ MSE and MAE. MSE addresses the robustness of the estimates, while MAE determines the accuracy of the estimates. Recently, Gao et al.~\cite{GaoEtAlcorr2020survey} and Fan et al.~\cite{FanZZLZW22} comprehensively reviewed the contemporary research advancements on crowd counting and density estimation and claimed that MAE and MSE are the most commonly used image-level measurements. On the other hand, classification problems use metrics of ACC, AUC, EER, F1S, and so on.

\begin{figure*}
\centering
\includegraphics[width=0.84\textwidth]{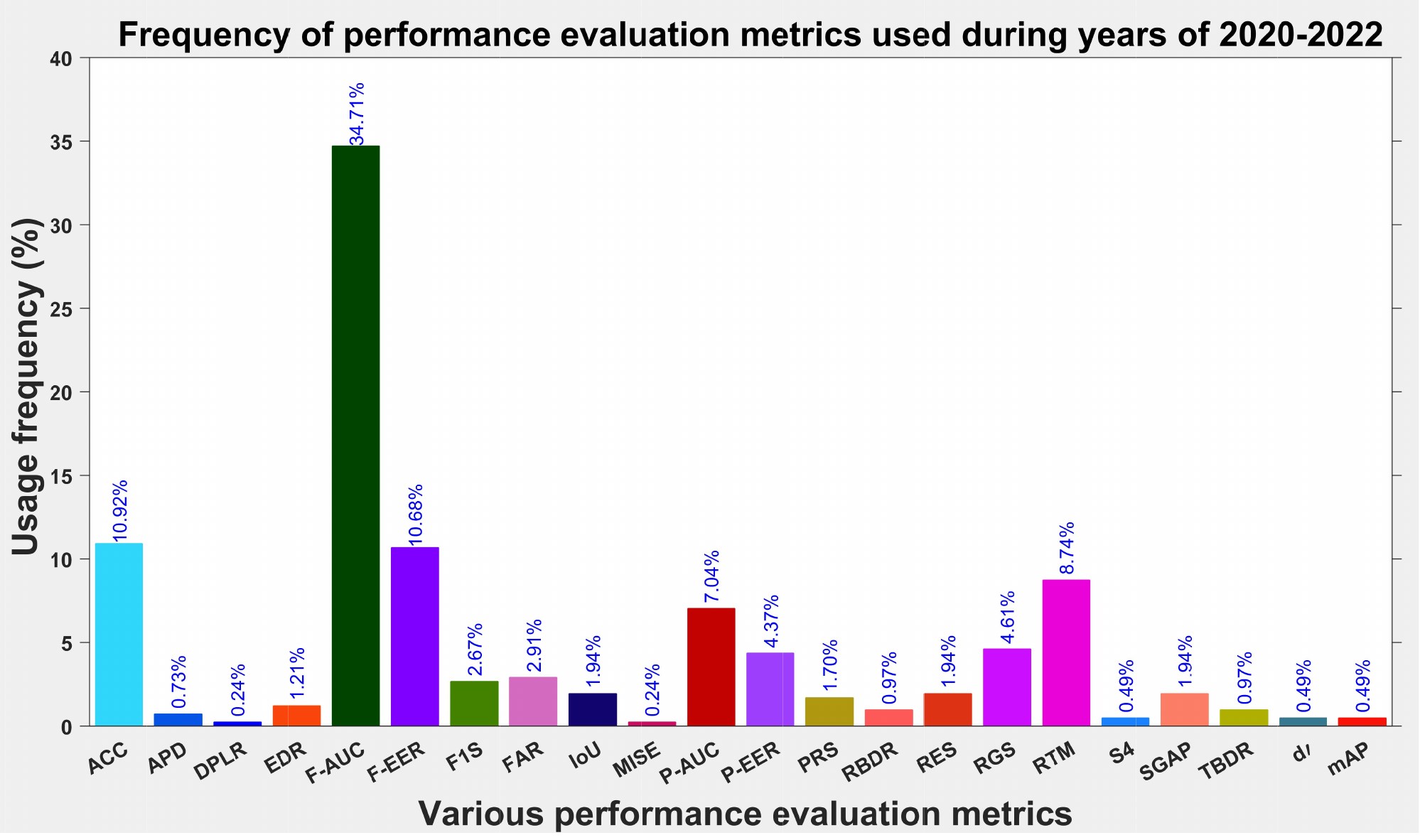}
\caption{Usage frequency of performance evaluation metrics between 2020 and 2022 considering Equation~(\ref{UsageFrequencyEq01}) and Table~\ref{SummaryOfLiteratureReview2020to2021}.}
\label{ACCAUCF1SEERploting}
\end{figure*}

We have calculated the usage frequency of an item (e.g., metric, dataset, and method) by using Equation~(\ref{UsageFrequencyEq01}) as:   
\begin{equation}\label{UsageFrequencyEq01}
\footnotesize
 \emph{Usage frequency (\%)} = \frac{(\emph{Total number of papers that used the item})(100)}{\emph{Sum of papers both that used and unused the item}}. \normalsize
\end{equation}  \normalsize
Figure~\ref{ACCAUCF1SEERploting} depicts the usage frequency of the performance evaluation metrics considering Equation~(\ref{UsageFrequencyEq01}) and Table~\ref{SummaryOfLiteratureReview2020to2021}.
Figure~\ref{ACCAUCF1SEERploting} shows that AUC, EER, ACC, and RTM became sequentially the most frequently applied performance indicators among a list of evaluation metrics for classification problems in deep learning. However, the usage frequency score of F-AUC hints that the F-AUC is an unparalleled performance indicator for anomalous frame localization.

\subsection{Our Observations}
In real-surveillance-video anomaly detection, we generally desire to rapidly detect the commencement and termination frames of the abnormal event. This is due to the fact that anomaly detection is a coarse-level perception and the clipped anomalous video segment can be sent for additional in-depth video analysis (e.g., activity detection, object detection, etc.)~\cite{NayakPD21}. Furthermore, recently, Cai et al.~\cite{CaiLGHL21} explained that in the actual abnormal frame the anomaly phenomenon is substantially distributed in a small patch of the image rather than across the full frame. If just that frame can be detected regardless, we neither need the pixel-level ground truth annotations nor the determination of the anomaly by computing either the MSE or PSNR of the complete frame. However, the standard F-AUC takes the paramount advantage in these respects. Consequently, the F-AUC became the most important and meaningful performance evaluation metric for video anomaly detection.

\section{Diversity of Crowd Datasets}\label{DatasetBriefDescriptionCrowdDatasets}  
Crowd activity monitoring in both private and public places is a very demanding endeavor.
The everyday activities of human beings create huge amounts of crowd trajectory data in both indoor and outdoor environments~\cite{ZhaoZCZH21}. The increase in crowd trajectory data availability offers new opportunities for reliability engineering analysis and assessment~\cite{LinZMTW21rel}. A dataset is a body of samples with shared attributes. 
Miscellaneous crowd datasets and benchmarks enable validating and comparing methods for developing smarter algorithms. However, accessing relevant images and datasets is one of the key challenges for image analysis researchers. Crowd datasets can be indoor, outdoor, or both. On the basis of both quantity and quality, they can be widely categorized as synthetic or real-world datasets.

\subsection{Synthetic Datasets: Quantity Unrestricted}
Synthetic datasets are algorithmically manufactured rather than recorded by real-world events. They are supposed to mimic real-world original data in such a way that both synthetic and real-world datasets cannot be differentiated from each other---not even by human domain specialists or sophisticated computer algorithms.

\subsubsection{Advantages of Synthetic Dataset}
The synthetic dataset might be a feasible solution to overcome some existing problems in real-world datasets. The quantity of synthetic data can be surpassed more easily than that of real-world data. For example, a billion synthetic frames can be generated with a powerful parallel machine (e.g., GPUs, and CPU clusters), whereas collecting such a number of real training samples might be impossible. One of the key advantages of a synthetic dataset is anonymity, as no personal information is available. The data cannot be traced back to the original owner. Thus, it does not need to consider possible copyright infringements
. Explicitly, synthetic data protect authentic data privacy and confidentiality. For instance, the US Census Bureau~\cite{CensusGov2021} utilized synthetic data without personal information that mirrored real data collected via household surveys for income and program participation. In addition to privacy protection in a synthetic dataset, it is fully annotated (i.e., there is no need for humans to spend time manually collecting and annotating training data), fully user-controlled, unrestricted to hardware devices for data collection, multi-spectral, easy to enrich with abnormal events, and cost-effective.

\subsubsection{Popular Synthetic Crowd Datasets}
Simulating pedestrian crowd movement in a virtual environment is not a novel task~\cite{SaeedRR22}.
We found many synthetic crowd datasets in the literature. For example, during 2020 and 2022 (see our review in Table~\ref{SummaryOfLiteratureReview2020to2021}), researchers used the following synthetic crowd datasets to detect abnormal events in crowd scenes.

\begin{enumerate}
\item \textbf{UMN}~\cite{UMNdataset2021} $\Rightarrow$ The UMN dataset is one of the standard crowd abnormal event testing datasets from the University of Minnesota. It is a synthetic dataset composed of three different scenes (e.g., one indoor and two outdoors)~\cite{LloydRMM17,SanchezHTH20}. Each footage had been recorded at a frame rate of 30 fps at a resolution of $480\times640$  using a static camera~\cite{LloydRMM17}. This dataset aims to accurately detect the changes in crowd movement. In each video scene, an unstructured crowd is walking, and the motion pattern is completely unstructured~\cite{SanchezHTH20}. An anomaly is marked when everyone suddenly starts running.

\item \textbf{SHADE}~\cite{LinGWL21} $\Rightarrow$  The SyntHetic Abnormality DatasEt (SHADE) was generated in the video game \textit{Grand Theft Auto V} (GTA5)~\cite{GTA5}. The videos in SHADE include those labeled as arrest, chase, fight, knockdown, run, shoot, scatter, normal type 1 (e.g., high-five and hug), and normal type 2 (e.g., people walk around normally).

\end{enumerate}

\subsubsection{Challenges of Synthetic Crowd Dataset}
Although a synthetic dataset possesses a long list of benefits, it comes with limitations. While synthetic data can mimic many properties of authentic data, it cannot copy the original content exactly. The quantity is not a problem for synthetic data. However, it is arduous to generate high-quality indoor and outdoor synthetic crowd datasets.
The main ultimatums for creating synthetic crowd datasets derive from the following facets:
\begin{itemize}
\item It is extraordinarily laborious to completely encode scenes with customized and vivid stories of real-world crowd information into synthetic crowd data;
\item The movements of each individual in the real world can be influenced by the strands of physical strength, current locality, crowd density, and subjective interest~\cite{ZhaoZCZH21}. Such strands limit the similar real-world movements of each individual in simulated data;
\item The crowd distribution varies in an indoor environment. Such variations in synthetic indoor data should satisfy the structures and functionalities of real-world indoor data.
\end{itemize}
The aforementioned factors might make the synthetic crowd datasets unpopular compared with the real-world version.

\subsection{Real-World Crowd Datasets: Quantity Restricted}
Real-world crowd data are observational data originally collected from a number of sources in real-world settings, usually by using sensors (e.g., cameras). Normally, the quality of a real-world crowd dataset is higher than that of a synthetic dataset.

\subsubsection{Advantages of Real-World Crowd Dataset}
While a synthetic dataset has many benefits and is highly useful in many circumstances, there is still a heavy reliance on human-annotated and real-world data. Compared with synthetic crowd datasets, real-world crowd datasets are basic, and they enrich the newly developed algorithms by providing more robust and confident results. Despite the different nature of synthetic datasets compared with real-world datasets, they can compromise the algorithmic output quality for critical decision-making. Even though synthetic crowd datasets are satisfactory, they are still inferior compared with certain properties of real-world crowd datasets 
.

\subsubsection{Popular Real-World Crowd Datasets} 
Many real-world crowd datasets have been proposed to detect anomalies in real-world video scenes. For example, during 2020 and 2022 (considering our succinct review in Table~\ref{SummaryOfLiteratureReview2020to2021}), researchers used the following real-world crowd datasets to detect abnormal events in crowd scenes.

\begin{itemize}
\item \textbf{LIVE}~\cite{SheikhSB06}  $\Rightarrow$ Basically a live dataset of real-world-scenario video sequences. The sequences were captured in daylight and at night time with different crowd levels. The sequences contain both normal and abnormal events with ground truths~\cite{RehmanTFJW21}. Both normal and abnormal events exit in a footpath, grocery shop, subway station, police booth, petrol pump, etc. The Video Quality Experts Group studied the video quality~\cite{VQEG2021}.

\item \textbf{Subway}~\cite{AdamRSR08} $\Rightarrow$ Two cameras recorded an underground train station. One camera pointed toward the entrance platform and the second camera observed the exit platform. Thus, this dataset contains two long videos of subway entrance and exit gate scenes.  Both video sequences are annotated at a frame level and have similar types of anomalies, including wrong-direction walking, loitering, and avoiding payment~\cite{CUHKdataset2021,LuSJ13,AsadYTCH21}.

\item \textbf{UCSD}~\cite{ChanLV08} $\Rightarrow$ The UCSD (University of California San Diego)  anomaly detection dataset was obtained from a stationary camera mounted at an elevation, which overlooked pedestrian walkways. The dataset is split into two subsets named Pedestrian 1 (Ped1) and Pedestrian 2 (Ped2). In Ped1, there is an acute angle between the camera view and the sidewalk, and the camera height is lower than in Ped2~\cite{HaoLWWG22}. The ground truth annotation includes a binary flag per frame, indicating whether an anomaly is present at that frame. Abnormal events mainly consist of two categories:  the movement of non-pedestrian entities and anomalous pedestrian motions~\cite{ZhouWLZT20}. Abnormal events include bikers, skaters, carts, wheelchairs, and people walking off the walkway.

\item \textbf{ImageNet}~\cite{DengDSLL009} $\Rightarrow$  It consists of over 15 million  high-resolution labeled images belonging to approximately 22000 categories.  It also includes variable-resolution images. The images were collected from the web and labeled by human labelers using Amazon's Mechanical Turk crowd-sourcing tool. The ImageNet project is a large visual database designed for use in visual-object-recognition software research. It has been running an annual software contest named ``ImageNet Large-Scale Visual Recognition Challenge (ILSVRC)" since 2010. The ILSVRC uses a subset of ImageNet. 

\item \textbf{PETS2009}~\cite{FerrymanEtal2009PETS2009} $\Rightarrow$ Its functions consist of: (i) crowd count and density estimation, (ii) the tracking of individual(s) within a crowd, and (iii) the detection of separate flows and specific crowd events in a real-world environment. The dataset scenarios were filmed from multiple cameras and involve multiple actors. It also contains crowd abnormal events~\cite{IlyasAQBH21}. Initially, there is a group of people normally walking around. As the anomaly begins, people start to disperse and run away. However, the lack of some other realistic scenarios, including fighting, fear, and abnormal objects, is a deficiency for this dataset~\cite{RabieeHMKNM16}.

\item  \textbf{UCF-Web}~\cite{MehranOS09} $\Rightarrow$ It was proposed by Mehran et al.~\cite{MehranOS09} as a harder version of the UMN~\cite{UMNdataset2021} dataset, with denser crowds~\cite{SanchezHTH20}. The anomalies are clash and escape panic scenarios~\cite{MuSYW21}.

\item \textbf{BEHAVE}~\cite{BlunsdenEtAl2010} $\Rightarrow$ In this dataset, various activities were simulated and captured at a rate of 25 fps.  An annotation file contains the starting and ending frame numbers and a class label that describes the event within the frames~\cite{DeepakSRC21}. Some of these events are grouping, walking together, chasing, and fighting. The anomalies are mainly caused by fighting.

\item \textbf{HockeyFight}~\cite{Nievas2011Hockey}  $\Rightarrow$  Nievas  et al.~\cite{Nievas2011Hockey}  collected 1000 clips of action from the hockey games of the National Hockey League  (NHL) in North America. Each clip is manually labeled as either ``fight" or ``non-fight". This dataset is used for fighting detection~\cite{AlmazroeyJ20}.

\item \textbf{UCF101}~\cite{SoomroEtAlcorr2012} $\Rightarrow$  This dataset consists of 13320 videos collected from youtube.com, including a variety of action forms. The format of the video is unified during the dataset's construction~\cite{Hu20}.

\item \textbf{Mall}~\cite{ChenLGX12} $\Rightarrow$ This dataset was collected from a publicly accessible webcam for crowd-counting and -profiling research.  Chen et al.~\cite{ChenLGX12} exhaustively annotated the data by labeling the head position of every pedestrian in all dataset frames.

\item \textbf{ViF}~\cite{HassnerIK12} $\Rightarrow$ The ViF (Violent Flows) is a dataset of real-world videos downloaded from the web consisting of crowd violence, along with standard benchmark protocols designed to test both violent/non-violent classification and violence outbreak detection. The average length of the video clips is 3.6 seconds.  The types of violent behaviors are related to the fighting events in the video clips~\cite{RabieeHMKNM16}.   

\item \textbf{CUHK-Avenue} ~\cite{LuSJ13} $\Rightarrow$ This dataset's videos were captured in the CUHK (Chinese University of Hong Kong) Campus Avenue~\cite{CUHKdataset2021}. The 16 training videos capture normal situations.
    The 21 testing videos include both normal and abnormal events. The latter are marked in rectangles. 
     The abnormal events are running, walking in opposite directions, throwing objects, and loitering~\cite{ZhouWLZT20,SunJSW21,HaoLWWG22}. The dataset contains a slight camera shake (e.g., in testing video 2 at frames 1051-1100) and a few outliers in the training data. Furthermore, normal patterns seldom appear in the training data~\cite{CUHKdataset2021}.

\item \textbf{CUHK-Crowd}~\cite{ShaoLW14} $\Rightarrow$  It includes crowd videos with various densities and perspective scales. Videos were collected from gettyimages.com and pond5.com, and included various environments, such as streets, shopping malls, airports, and parks.

\item \textbf{GBA2016}~\cite{MartinezEtAl2016} $\Rightarrow$  These videos were recorded on different days with a GoPro HERO4 camera at 50 fps at the Polytechnics School of the University of Alcala. The camera was located at a high height to reduce occlusions. The dataset includes various individuals performing  actions, e.g., walking, running, sitting down, and falling.

\item \textbf{ShanghaiTech}~\cite{ZhangZCGM16} $\Rightarrow$ It consists of two parts, namely PartA and PartB. The images of PartA were taken from the Internet, whereas the images of PartB were collected from a metropolitan street in the city of Shanghai. The training and evaluation sets are defined by the authors~\cite{ZhangZCGM16}. The training and testing phases are very biased in this dataset because the images are of various density levels and are not uniform~\cite{KhanAKQD20}.

\item \textbf{MED}~\cite{RabieeHMKNM16}  $\Rightarrow$ The videos of the MED (Motion Emotion Dataset) were recorded at 30 fps using a fixed video camera elevated at a height, overlooking individual walkways. The MED consists of ground truth emotion labels. It also comprises crowd behavior annotations. All videos start with normal behavior frames and end with abnormal ones.

\item \textbf{LV}~\cite{LeyvaSL17} $\Rightarrow$ The LV (Live Videos) dataset was captured by surveillance cameras from different view angles at various resolutions and frame rates~\cite{LeyvaSL17}.  Abnormal events include fighting, robbery, accidents, murder, kidnapping, and an illegal U-turn. Each abnormal event highly depends on the nature of its environment. The abnormal events are localized by specifying the regions of interest in a separate sequence.

\item \textbf{CUHK-Avenue17}~\cite{HinamiMS17} $\Rightarrow$ It is a subset of CUHK-Avenue~\cite{LuSJ13}. The training set is identical to CUHK-Avenue~\cite{LuSJ13}, however, the testing set is smaller. Hinami et al.~\cite{HinamiMS17} argued that the CUHK-Avenue~\cite{LuSJ13} testing set contains five videos (e.g., 1, 2, 8, 9, and 10) with static abnormal objects that are not properly annotated~\cite{Ionescu2018habil}.
    Henceforth, they evaluated their approach on a subset (called Avenue17) that excludes these five videos.

\item \textbf{ShanghaiTech Campus}~\cite{LuoLG17} $\Rightarrow$   The data collection was performed at ShanghaiTech University campus considering 13 different scenes with various lighting conditions and camera angles~\cite{AsadYTCH21}. This dataset is one of the biggest and most challenging datasets available for video anomaly detection~\cite{DoshiEtAlwacv2022a}. Anomalous events are produced by strange objects in the scenes, such as pedestrians moving at anomalous speeds (e.g., running and loitering) and in unexpected directions~\cite{SanchezHTH20}. The dataset has 130 abnormal events in 13 scenes~\cite{ShanghaiTechCampusDataset2021}. All abnormal videos are in the testing set because the dataset is proposed for unsupervised learning. To adapt to the weakly supervised setting, Zhong et al.~\cite{ZhongLKLLL19} reorganized the videos into 238 and 199 training and testing videos, respectively.

\item \textbf{UCF-QNRF}~\cite{IdreesTAZARS18} $\Rightarrow$ The UCF-QNRF is a large-scale crowd-counting dataset.
It contains a variety of scenarios and scenes with high resolutions and densities. Usually, it has extremely congested scenes, where the maximum count of an image can reach up to 12865~\cite{IdreesTAZARS18}.  It also includes  buildings, plants, the sky, and paths.  

\item \textbf{UCF-Crime}~\cite{SultaniCS18} $\Rightarrow$   It consists of 1900 long and untrimmed real-world surveillance videos. Usually, trimming refers to taking off either part of the beginning or end of a video clip. Anomalies include abuse, arrest, arson, assaults, road accidents, burglaries, explosions, fighting, robberies, shooting, stealing, shoplifting, and vandalism~\cite{UCF-CrimeDataset2021}.

\item \textbf{FDST}~\cite{FangZCGH19} $\Rightarrow$ Fang et al.~\cite{FangZCGH19} proposed a large-scale crowd-counting video dataset named Fudan-ShanghaiTech (FDST) with frame-wise ground truth annotation. The FDST contains many different scenes, including shopping malls, squares, and hospitals. It took more than 400 hours to annotate the FDST dataset~\cite{FangZCGH19}.

\item \textbf{IITB-Corridor}~\cite{RodriguesEtAlcorr2019} $\Rightarrow$ The videos in the IIT Bombay campus were captured with a single camera. The scene consists of a corridor with many normal and abnormal activities. Normal activities include walking and standing, whereas abnormal activities include protests, unattended baggage, cycling, sudden running, fighting, chasing, loitering, suspicious objects, hiding, and playing with a ball. The annotations for normal and abnormal video frames are provided at the frame level.

\item \textbf{MVTec}~\cite{BergmannEtAl2019mvtecad} $\Rightarrow$ All images of the MVTec anomaly detection dataset were captured by a $2048\times2048$-pixel high-resolution industrial RGB sensor. The training set contains non-anomalous objects, whereas the testing set includes various types of anomalies and non-anomalous samples. The dataset includes a detailed ground truth with pixel-wise mask annotations for each anomalous region. It comprises almost 1900 manually annotated regions.

\item \textbf{StreetScene}~\cite{RamachandraJ20}   $\Rightarrow$  This dataset consists of 46 training video sequences and 35 testing video sequences taken from a static USB camera looking down on a two-lane street with bike lanes and pedestrian sidewalks. Videos were collected from the camera during daylight at various times in two consecutive summers. The main recorded activities are cars driving, turning, stopping, and parking; pedestrians walking, jogging, and pushing strollers; and bikers riding in bike lanes.

\item \textbf{CitySCENE}~\cite{CitysceneDataset2020}  $\Rightarrow$  This dataset consists of a variety of real-world anomalies, which include carrying objects, crowds, graffiti, sweeping, smoking, and walking dogs. 
    The training and testing sets are trimmed and untrimmed, respectively. It can be used to compare algorithms for general and specific real-world anomaly detection~\cite{CitysceneDataset2020}.

\item \textbf{AI-CityChallenge20:T4}~\cite{Naphade20AIC20}  $\Rightarrow$  The 4th annual edition of the AI City Challenge has four challenging tracks. Track 4 (T4) addressed traffic anomaly detection~\cite{Naphade20AIC20}. The anomalous behaviors mainly consisted of vehicles driving off the road, stalled vehicles, and crashes. The viewing angles, weather, and lighting conditions of each video produced a unique and challenging dataset~\cite{DoshiY20a}. More than 25 hours of video data were captured on highways in Iowa, USA.

\item  \textbf{RWF2000}~\cite{ChengCL20} $\Rightarrow$ The RWF2000 (Real-World Fighting) dataset comprises 2000 real-life surveillance-video clips downloaded from youtube.com. Each video clip has a duration of 5 seconds with a frame rate of 30 fps.
    Because the clips were extracted from roughly 1000 unique videos, the authors~\cite{ChengCL20}  manually checked the training and testing sets to avoid data leakages. Anomalous behaviors mainly involved in-crowd, two-person, and multiple-person scenarios, making this a multiplex and arduous identification-modeling task~\cite{UllahUHMHSBA22}.

\item \textbf{JHU-CROWD++}~\cite{SindagiEtAl2020} $\Rightarrow$ Sindagi et al.~\cite{SindagiEtAl2020} introduced a large-scale unconstrained crowd-counting dataset named JHU-CROWD++, which was collected under a variety of diverse scenarios and environmental conditions, including  weather-based degradation and illumination variations. The JHU-CROWD++ consists of a set of annotations at both image and head levels The images were annotated with the help of Amazon Mechanical Turk workers~\cite{MTur2021}.

\item \textbf{AI-City Challenge21:T4}~\cite{Naphade21AIC21}  $\Rightarrow$  The fifth AI City Challenge has five tracks.
Track 4 consists of 100 training videos and 150 testing videos, each with an approximate length of 15 minutes. The videos were captured at a frame rate of 30 fps. The purpose of the challenge is to devise an algorithm that is capable of identifying all anomalies with minimum false alarms and detection delays. More than 62 hours of video data were captured on highways in Iowa, USA.

\item \textbf{NeuroAED}~\cite{ChenLLTHDCK21}  $\Rightarrow$  This dataset was acquired with a stationary neuromorphic vision sensor (DAVIS346) mounted on top of a retractable tripod with a maximum elongation of five meters. Pan-tilt was used to adjust the camera angle to ensure the coverage of the entire region of interest.  It consists of four sub-datasets, namely walking, campus, square, and stair. The abnormal events present are labeled as bikes or motorcycles.

\item \textbf{X-MAN}~\cite{SzymanowiczCC21}   $\Rightarrow$ The X-MAN (eXplanations of Multiple sources
of ANomalies)  dataset evaluates anomaly explanation methods. It consists of 22722 manually labeled frames in UCSD Ped2~\cite{ChanLV08} (1648), CUHK-Avenue~\cite{LuSJ13} (3712), and ShanghaiTech Campus~\cite{LuoLG17} (17362). Each frame holds between one and five explanation labels. Each label has a non-identical reason regarding why the frame should be anomalous.

\item \textbf{NOLA}~\cite{DoshiEtAlwacv2022a}  $\Rightarrow$  This dataset focused on a single-scene setup with 110 training video segments in 11 splits and 50 test segments captured during day and night, as well as on various days of the week, using a single moving camera from a famous street in New Orleans, USA.

\end{itemize}

\subsubsection{Challenges of Real-Word Crowd Datasets}
To analyze video sequences, ground truth video sequences are all-important. Woefully, to produce and annotate such ground truth video sequences at a rational level of detail is very time-consuming. For example, in the experience of Blunsden et al.~\cite{BlunsdenEtAl2010}, one hour of video (with more or less 90000 frames), took approximately six person-months of time for annotation at the level of individual bounding boxes and frame-by-frame behavior. 
 However, it is extremely hard to collect a sufficient number of videos for specific abnormal crowd behaviors. As the dataset increases in size, more challenges arise (e.g., the problem of manually labeled videos with different behaviors). In addition to confidentiality and time-sensitive issues, some common challenges in real-world crowd datasets arise from:
\begin{itemize}
  \item The limited number of training samples;
  \item The inclusion of only crowd images---a deep network can erroneously predict crowd even in scenes that do not contain crowd;
  \item Limited annotations;
  \item A lack of diversity regarding scenes and viewing angles~\cite{LuoLG17};
  \item Not including adverse weather conditions, such as haze, snow, and rain~\cite{SindagiEtAl2020};  
  \item Not including shifting illumination and man-made changes;
  \item Not including as many anomaly events as possible for anomaly detection cases;
  \item Neglecting to cover a wide range of geographical regions~\cite{JafarzadehEtAl2021corr} and camera devices.
\end{itemize}

\subsection{Comparison with Specifications of Crowd Datasets}
\begin{figure*}
\centering
\includegraphics[width=0.99\textwidth]{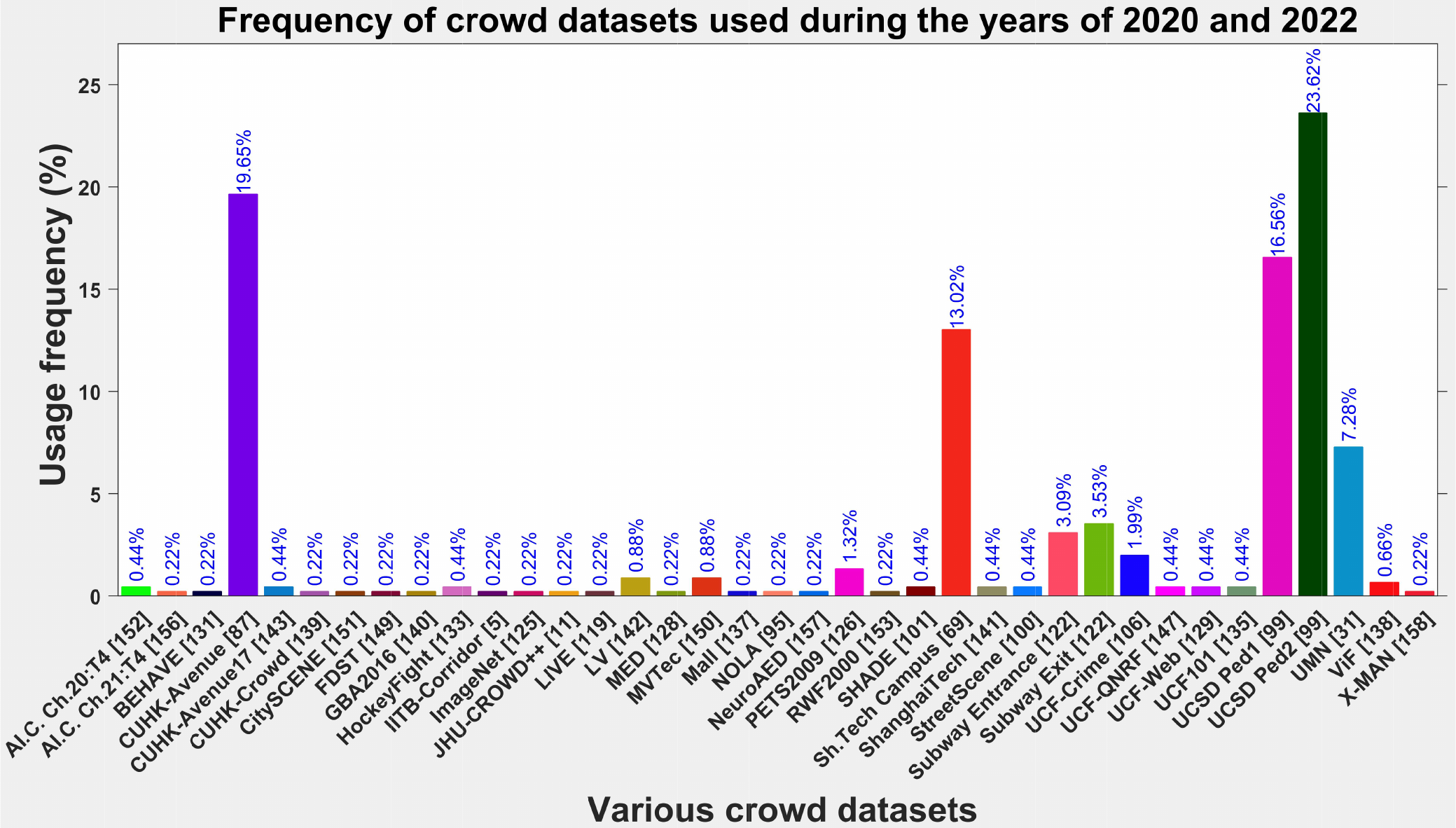}
\caption{Dataset usage frequency from 2020 to 2022 considering Table~\ref{SummaryOfLiteratureReview2020to2021}.}
\label{DatasetUsageFreq}
\end{figure*}

Most of the aforementioned crowd datasets provided considerable challenges. Sorting them by year, Table~\ref{DatasetsComparativeSummary} compares their numerous specifications.
Figure~\ref{DatasetUsageFreq} depicts the usage frequency of datasets considering Equation~(\ref{UsageFrequencyEq01}) and Table~\ref{SummaryOfLiteratureReview2020to2021}.

Figure~\ref{DatasetUsageFreq} shows that CUHK-Avenue~\cite{LuSJ13}, ShanghaiTech Campus~\cite{LuoLG17},  UCSD~\cite{ChanLV08}, and  UMN~\cite{UMNdataset2021} are the most popular datasets. Furthermore, it is noticeable that the real-world dataset UCSD~\cite{ChanLV08} has become the predominant standard benchmark for anomaly detection.
This dataset contains two shooting angles. The first shooting angle involves pedestrians approaching or moving away from the surveillance camera, whereas the second shooting angle shows pedestrians moving parallel to the camera. Only pedestrians appear in normal events, and abnormal events include bikes and trucks. The major challenge of this dataset is that the density of pedestrians on the road constantly changes from sparse to crowded~\cite{LiHDZCS22}. Nonetheless, the utilization of Ped2 in the literature is higher than Ped1 because of its resolution 
. For example, Hinami et al.~\cite{HinamiMS17}  selected Ped2 because Ped1 has a significantly lower frame resolution of $158\times240$, which would have made it difficult to capture objects in their model. Furthermore, in Ped1, there is an acute angle between the camera view and the sidewalk, and the height of the camera is lower than that of Ped2. Thus, the size of a person is swapped, which will also affect the easily perceived anomalies (e.g., skateboarders and bicycles)~\cite{HaoLWWG22}. 
Besides, some characteristics are gratuitously labelled as anomalies~\cite{CaiLGHL21}.
Notably, Chang et al.~\cite{ChangTXLZSY22} employed only Ped2 because some UCSD events ~\cite{ChanLV08} are labeled as normal in the training set but are considered anomalous in the testing set. Moreover, the camera location in Ped2 is higher. Therefore, the camera viewpoint is vertical to the sidewalk.  Thus, the size of a person is comparatively fixed in Ped2~\cite{HaoLWWG22}.
Anomalies in UCSD~\cite{ChanLV08} videos are assumed to be simple. Therefore, real-world anomalous events are not sufficiently reflected in video surveillance~\cite{MohammadiEtAl2021corr}. Despite being extensively used as a benchmark dataset, most anomalies in UCSD~\cite{ChanLV08} are crystal clear and can be effortlessly recognized from a single frame~\cite{DoshiEtAlwacv2022a}.

The real-word CUHK-Avenue dataset~\cite{LuSJ13} secured the second-best benchmark for a crowd dataset. The key reason could be that the authors would likely validate the performance of their models with the real-world challenges of this dataset. The challenges of the CUHK-Avenue~\cite{LuSJ13} dataset include: (1) a slight camera shake presents in frames 1051-1100 of test video 2~\cite{LiHDZCS22}; (2) the varying size and scale of people in the dataset because of the changing positions and view angles of the cameras~\cite{LiHDZCS22,ChenYCXJ21,HaoLWWG22,ChangTXLZSY22}; and (3) the inclusion of a few outliers in the training data, in addition to some normal patterns seldom appearing in the training data~\cite{LiHDZCS22}.

The real-world ShanghaiTech Campus dataset~\cite{LuoLG17}  became the third best frequently utilized benchmark.
It differs from the UCSD and CUHK Avenue datasets because it covers 13 different scenes (e.g., streets, squares, entrances of cafeterias, etc.) for both training and testing~\cite{FengSCCNC21,ChenYCXJ21,LiHDZCS22,HaoLWWG22}. Its abnormal events contain various situations. In addition to abnormal objects (e.g., vehicles and bicycles), abnormal behaviors (e.g., fighting and robbing) are also collected for testing~\cite{HaoLWWG22}. The ratios of each scene in the training and test sets can be varied~\cite{FengSCCNC21}. All these distinctive challenges made the ShanghaiTech Campus~\cite{LuoLG17} dataset unique in validating many algorithms.
However, the videos are captured from 13 distinct cameras, which put them together in a multi-scene articulation. Additionally, handling it as 13 individual datasets restricts the number of accessible training frames for each scene~\cite{DoshiEtAlwacv2022a}. Furthermore, a runtime error can occur if the dataset's size exceeds the computer memory.

Although the UCSD~\cite{ChanLV08}, CUHK-Avenue~\cite{LuSJ13}, and ShanghaiTech Campus~\cite{LuoLG17} datasets are very popular, they retain relatively clean data that were recorded at similar times during the day and in clear-weather setups. As a result, these datasets are not substitutes for real-world surveillance footage, where superficial factors including weather and the time of the day can affect the quality of the accumulated frames~\cite{LerouxEtAlwacv2022}. Additionally, Gudovskiy et al.~\cite{GudovskiyEtAlwacv2022} did not use the Subway~\cite{AdamRSR08} dataset because the dissimilar ground truth is annotated in different works~\cite{LuSJ13,Hasan0003CNRD16}, and non-identical ground truth is effective for the performance assessment of dissimilar  methods~\cite{GudovskiyEtAlwacv2022}.

\subsection{Our Observations}
\subsubsection{\textbf{Usage of Existing Datasets}} 
The performance of deep learning models can often be improved by feeding them more training data.  In essence, the UCSD~\cite{ChanLV08} dataset is simple. However, sometimes a deep model's poor temporal features extraction leads to low performance in UCSD~\cite{ChanLV08}~\cite{EsquivelG22}. Ped1 of UCSD~\cite{ChanLV08} holds additional anomalies that can at best be extracted by learning the normal temporal features. For example, skateboards in Ped1~\cite{ChanLV08} are a difficult shape for the human eye to recognize but they are noticed due to their speed.
Usually, a deep model can perform well in sparsely populated scenes (e.g., the UCSD~\cite{ChanLV08} dataset). However, the same model can lose its superiority in dense crowd scenes (e.g., the ShanghaiTech Campus~\cite{LuoLG17} dataset).
Thus, the probability of success when testing any new deep model with the UCSD~\cite{ChanLV08} dataset is higher than that of the ShanghaiTech Campus~\cite{LuoLG17} dataset.
Furthermore, both real-world (e.g., UCSD~\cite{ChanLV08}) and synthetic (e.g., UMN~\cite{UMNdataset2021}) datasets can help to show the robustness of the underlying models. 
To this end, a great number of researchers (see Table~\ref{SummaryOfLiteratureReview2020to2021}) validated their models using both UCSD~\cite{ChanLV08} and  UMN~\cite{UMNdataset2021} datasets. All these options not only made UCSD~\cite{ChanLV08} a top choice among existing datasets but also made UMN~\cite{UMNdataset2021} a more popular dataset compared with other available synthetic crowd datasets.

\subsubsection{\textbf{Own Problem of Datasets}}
Neural networks trained on popular datasets can suffer from overinterpretation. In overinterpretation, algorithms make confident predictions based on details that do not make sense to humans (e.g., random patterns and image borders). For example, neural models trained on the CIFAR-10~\cite{MScThesisKrizhevsky2009} dataset made confident predictions even when 95 percent of the input images were missing, with the remainder remaining senseless to humans~\cite{NewsFromAcademicGates2021}. Deep learning models can latch onto both meaningful and meaningless subtle signals. When deep model classifiers are trained on crowd datasets (e.g., ImageNet~\cite{DengDSLL009}), they can make seemingly reliable predictions based on sensible and senseless signals. Thus, the overinterpretation is solely a dataset problem. This problem cannot be diagnosed using typical evaluation methods based on the accuracy of the model. Regarding datasets, a common question can be asked: How would we adapt any crowd dataset in a way that may enable deep models to be trained to mimic closely how a human thinks about categorizing images from real-world scenarios?

\onecolumn

\captionsetup{width=1.0\textwidth}
\setlength{\tabcolsep}{0.093cm}
\scriptsize

\begin{longtable}{
|p{0.39cm}| p{0.75cm}  p{1.05cm}  p{0.50cm}   p{0.75cm} p{0.85cm} 
p{0.69cm} p{0.60cm} p{0.50cm}  p{1.05cm} 
p{0.50cm} p{0.50cm} p{0.50cm} p{0.70cm} p{0.70cm} p{0.70cm}   
p{2.15cm} p{1.35cm}|}

\caption{Comparison of various specifications of crowd datasets and their available web links. $H \Rightarrow$ height, $W\Rightarrow$ width.}

\label{DatasetsComparativeSummary}\\\hline

\multirow{2}{*}{Year} &\multirow{2}{*}{Dataset} &\multirow{2}{*}{Source} &Scene  &Video &Anomaly
&\multicolumn{3}{c}{Number of Videos} &Resolution   
&\multicolumn{3}{c|}{Annotation} &\multicolumn{3}{c}{Number of Frames}   
&\multirow{2}{*}{Remark on Anomaly}&\multirow{2}{*}{Dataset Link} \\\cline{7-9} \cline{11-12} \cline{13-15}
&       &    &Count &Length&Count&Training&Testing&Total &$H\times W$ &Using&Type&Count&Training&Testing&Total & & \\\hline

\multirow{6}{*}{\rotatebox{77}{\scriptsize  \textbf{2006}\scriptsize}}
&LIVE \cite{SheikhSB06}   
&Captured without planned object movements.   
&5      
&NA     
&NA     
&NA     
&NA     
&28     
&\tiny{$768\times$$1024$  $512\times$$768$}  
&Hu-man 
&Region level 
&\,\,\,\,NA    
&NA    
&NA    
&8400  
&Anomalies include vandalism, highway car accidents, vehicles in pedestrian areas, earthquakes, street crime, etc.  
&\url{https://live.ece.utexas.edu/research/quality}
\\\hline

\multirow{20}{*}{\rotatebox{77}{\scriptsize \textbf{2008} \scriptsize}}
&Subway Entrance \cite{AdamRSR08}  
&A real-life surveillance system.   
&5 
&96 minutes 
&66 
&NA   
&NA   
&NA  
&\tiny{$384\times$$512$}  
&Hu-man 
&Frame level 
&\,\,NA  
& 20000  
&116524 
&136524  
&Walking in the wrong direction, avoiding payment, loitering, irregular interactions between people, and miscellaneous. 
&NA  \\\cline{2-18}

&Subway Exit \cite{AdamRSR08}  
&A real-life surveillance system.   
&3 
&43 minutes 
&19 
&NA   
&NA   
&NA  
&\tiny{$384\times$$512$}  
&Hu-man 
&Frame level 
&\,\,NA  
&7500  
&24926 
&32426  
&Walking in the wrong direction, avoiding payment, loitering, irregular interactions between people, and miscellaneous. 
&\url{https://vision.eecs.yorku.ca/research/anomalous-behaviour-data/}
\\\cline{2-18}

&UCSD Ped1 \cite{ChanLV08}  
&Using 1st outdoor surveillance camera.
&5 
&5 minutes 
&40 
&34  
&36  
&70  
&\tiny{$158\times$$238$}  
&Hu-man 
&Pixel level 
&NA  
&6800  
&7200  
&14000  
&Circulation of non-pedestrian entities in the walkways and anomalous pedestrian motion patterns.
& \url{http://www.svcl.ucsd.edu/projects/anomaly/dataset.htm}
\\\cline{2-18}

&UCSD Ped2 \cite{ChanLV08}  
&Using 2nd outdoor surveillance camera.
&5 
&5 minutes 
&12 
&16  
&12  
&28  
&\tiny{$240\times$$360$}  
&Hu-man 
&Pixel level 
&NA  
&2550  
&2010  
&4560  
&Circulation of non-pedestrian entities in the walkways and anomalous pedestrian motion patterns.
& \url{http://www.svcl.ucsd.edu/projects/anomaly/dataset.htm}
\\\hline

\multirow{25}{*}{\rotatebox{77}{\scriptsize\textbf{2009}\scriptsize}} 
&Image Net \cite{DengDSLL009}  
&Web data, Amazon Mechanical Turk.
&5247 categories    
&NA     
&NA     
&NA     
&NA     
&NA     
&\tiny{Mean $350\times$$400$}  
&Hu-man 
&Image--Object levels 
&\,\,NA 
&2.88 million images   
&0.32 million images   
&3.2 million images  
&Its applications are object recognition, image classification, and automatic object clustering. 
&\url{https://image-net.org/update-mar-11-2021.php}
\\\cline{2-18}

&PETS 2009 \cite{FerrymanEtal2009PETS2009}  
&Multiple cameras, Reading University.
&8 views     
&NA     
&NA     
&3 sets    
&NA     
&NA     
&\tiny{$576\times$$768$  $576\times$$720$}  
&Hu-man 
&Frame Object levels 
&\,\,NA 
&NA   
&NA   
&NA  
&Detection of separate flows and specific crowd events in a real-world environment. 
&\url{http://cs.binghamton.edu/~mrldata/pets2009} 
\\\cline{2-18}

&UMN  \cite{UMNdataset2021}  
&Synthetic dataset \cite{LloydRMM17,SanchezHTH20}.
&3 
&4.3 minutes 
&11 
&NA   
&NA   
&11  
&\tiny{$480\times640$  $240\times$$320$}  
&Soft-ware 
&Temporal level 
&NA  
&NA  
&NA  
&7725  
&In each video crowd walks randomly. Everyone starts running suddenly, which is marked as an anomaly. 
&\url{http://mha.cs.umn.edu/proj_events.shtml#crowd}
\\\cline{2-18}

&UCF-Web \cite{MehranOS09}   
&Various media-hosting websites.
&5   
&NA  
&8   
&NA     
&NA     
&20    
&NA  
&Hu-man  
&Frame level 
&\,\,NA  
&NA  
&NA  
&NA  
&Abnormal events are classified as real-life escape panic, clashing, or fighting scenarios. 
&\url{https://www.crcv.ucf.edu/research/projects/abnormal-crowd-behavior-detection-using-social-force-model}
\\\hline

\multirow{6}{*}{\rotatebox{77}{\scriptsize\textbf{2010}\scriptsize}}
&BE-HAVE \cite{BlunsdenEtAl2010}  
&Trading tripod-mounted camcorder.
&5 
&51.2 minutes 
&NA   
&NA   
&NA   
&4    
&\tiny{$480\times$$640$}  
&Hu-man  
&Frame level 
&83545 bounding boxes  
&NA 
&NA  
&76800  
&Anomalies are mainly produced by fighting. Only the first sequence is fully annotated \cite{SanchezHTH20}. 
&\url{https://groups.inf.ed.ac.uk/vision/BEHAVEDATA/INTERACTIONS}  \\\hline

\multirow{6}{*}{\rotatebox{77}{\scriptsize \textbf{2011} \scriptsize}}
&Hockey Fight  \cite{Nievas2011Hockey}   
&Hockey games of the National Hockey League.
&NA     
&NA     
&500    
&NA     
&NA     
&1000   
&\tiny{$576\times$$720$}  
&Hu-man 
&Frame level 
&\,\,NA 
&NA  
&NA  
&50000 
&It is used for fighting detection in 500 fighting and 500 non-fighting short video clips. 
&\url{http://visilab.etsii.uclm.es/personas/oscar/FightDetection/HockeyFights.zip}  \\\hline

\multirow{10}{*}{\rotatebox{77}{\scriptsize \textbf{2012} \scriptsize}}
&UCF101 \cite{SoomroEtAlcorr2012}   
&Obtained from \url{www.youtube.com}.
&101  
&1620 minutes  
&NA   
&NA     
&NA     
&13320    
&\tiny{$240\times$$320$}  
&Hu-man  
&Spatio temporal level 
&\,\,24 groups  
&NA  
&NA  
&NA  
&This benchmark dataset is basically applied for action recognition. 
&\url{https://www.crcv.ucf.edu/data/UCF101.php}
\\\cline{2-18}

&Mall \cite{ChenLGX12}   
&At the shopping mall using a camera.
&8  
&33 minutes with 1 fps rate  
&NA     
&NA     
&NA     
&NA     
&\tiny{$240\times$$320$}  
&Hu-man  
&Pixel level 
&62325 pedestrians 
&\,\,800  
&\,\,1200  
&\,2000 
&It is mainly used for crowd-counting and -profiling research. 
&\url{http://personal.ie.cuhk.edu.hk/~ccloy/downloads_mall_dataset.html}
\\\hline

\multirow{6}{*}{\rotatebox{77}{\scriptsize \textbf{2012} \scriptsize}}
&ViF \cite{HassnerIK12}   
&Obtained from \url{www.youtube.com}.
&Many     
&Mean 14.8 minutes        
&NA     
&NA     
&NA     
&246   
&\tiny{$240\times$$320$}  
&Hu-man 
&Video level 
&\,\,NA 
&NA  
&NA  
&NA  
&This dataset consists of both violent and non-violent crowd behavior in real-world videos.  
&\url{https://talhassner.github.io/home/projects/violentflows/index.html}
\\\hline

\multirow{6}{*}{\rotatebox{77}{\scriptsize \textbf{2013} \scriptsize}}
&CUHK-Avenue \cite{LuSJ13}  
&Captured in CUHK campus avenue.
&5 
&30 minutes 
&47 
&16  
&21  
&37  
&\tiny{$360\times$$640$}  
&Hu-man  
&Frame--Pixel levels 
&\,\,NA  
&15328  
&15324 
&30652  
&Strange actions, wrong directions, unexpected objects.  The ground truth of anomalous objects
is marked by a rectangle. 
&\url{www.cse.cuhk.edu.hk/leojia/projects/detectabnormal/dataset.html}  \\\hline

\multirow{9}{*}{\rotatebox{77}{\scriptsize \textbf{2014} \scriptsize}}
&CUHK-Crowd  \cite{ShaoLW14}   
&Obtained from \url{www.gettyimages.com} \url{www.pond5.com}.
&215  
&120 minutes  
&8   
&NA     
&NA     
&474    
&\tiny{NA } 
&Hu-man 
&Frame level 
&\,\,927 groups  
&NA  
&NA  
&NA  
&It is mainly used for clustering pedestrians, crowd segmentation, group behavior analysis, and anomaly detection (e.g., crowd crossing in opposite directions). 
&\url{https://amandajshao.github.io/projects/CUHKcrowd_files/cuhk_crowd_dataset.htm}
\\\hline

\multirow{16}{*}{\rotatebox{77}{\scriptsize \textbf{2016} \scriptsize}}
&GBA 2016  \cite{MartinezEtAl2016}   
&GoPro HERO4 camera at U. of Alcala.
&4 
&NA 
&NA   
&NA     
&NA     
&3000    
&\tiny{$720\times$$1280$}  
&Hu-man 
&Frame level 
&\,\,NA  
&NA  
&NA  
&NA  
&Videos corresponding to people falling down are considered anomalous. 
&NA
\\\cline{2-18}

&Shan. Tech \cite{ZhangZCGM16}  
&Internet, some busy streets. 
&NA  
&NA 
&NA  
&NA  
&NA  
&NA  
&\tiny{Vary}  
&Hu-man 
&Pixel level 
&330165 heads  
&\,\,\,700 images 
&498 images 
&1198 images 
&Suitable for evaluating the crowd-count task. 
& \url{https://paperswithcode.com/dataset/shanghaitech}
\\\cline{2-18}

&MED  \cite{RabieeHMKNM16}   
&Video camera fixed at walkways.   
&5     
&NA     
&NA     
&NA    
&NA     
&31     
&\tiny{$235\times$$554$}  
&Hu-man 
&Video level 
&45000 video clips 
&NA   
&NA   
&NA  
&Motorcycle crossing a crowded scene, a suspicious backpack left by a person in a crowd, a motorcycle that is left in a crowd. 
&\url{https://www.dropbox.com/s/kgb6wqeuxud7r5t/Motion_Emotion_Dataset.zip}
\\\hline

\multirow{15}{*}{\rotatebox{77}{\scriptsize \textbf{2017} \scriptsize}}
&LV \cite{LeyvaSL17}   
&Captured by surveillance cameras.   
&30      
&236 minutes     
&34     
&NA     
&NA     
&30     
&\tiny{$144\times$$176$  $720\times$$1280$}  
&Hu-man 
&Region level 
&\,\,NA    
&127500    
&182440    
&309940    
&Abnormal events include fighting, robbery, accidents, murder, kidnapping, an illegal U-turn, etc.  
&\url{https://cvrleyva.wordpress.com}
\\\cline{2-18}

&CUHK-Ave-nue17 \cite{HinamiMS17}  
&Updated CUHK-Avenue \cite{LuSJ13}.  
&5 
&25 minutes 
&47 
&16  
&16  
&32  
&\tiny{$360\times$$640$}  
&Hu-man 
&Frame--Pixel levels 
&\,\,NA     
&15328  
&10622  
&25950  
&From testing, videos 1, 2, 8, 9, and 10 are excluded from CUHK-Avenue \cite{LuSJ13} dataset \cite{Ionescu2018habil}.
&\url{www.cse.cuhk.edu.hk/leojia/projects/detectabnormal/dataset.html}
\\\cline{2-18}

&Shan. Tech Campus \cite{LuoLG17}  
&University campus, surveillance cameras. 
&13 
&NA 
&130  
&238  
&199  
&437  
&\tiny{$2048\times$$2048$}  
&Hu-man 
&Pixel level 
&NA  
&274515 
& 42883 
&317398 
&Anomalies caused by sudden motion, including chasing and brawling. 
&\url{https://svip-lab.github.io/dataset/campus_dataset.html}
\\\hline

\multirow{12}{*}{\rotatebox{77}{\scriptsize \textbf{2018} \scriptsize}}
&UCF-QNRF  \cite{IdreesTAZARS18}  
&Flickr, Google, Hajj footage. 
&NA 
&NA 
&NA   
&NA   
&NA  
&NA   
&\tiny{$2013\times$$2902$}  
&Hu-man 
&Pixel level 
&1.25 million heads  
& 1201 images 
&  334 images 
& 1535 images 
&It is mainly dedicated to counting and density map estimation. 
&\url{https://www.crcv.ucf.edu/data/ucf-qnrf/}
\\\cline{2-18}

&UCF-Crime  \cite{SultaniCS18}  
&\url{www.youtube.com}
\url{www.liveleak.com}.   
&13 
&7680 minutes 
&13 
&1610   
&290   
&1900  
&\tiny{$240\times$$320$}  
&Hu-man 
&Video--Temporal levels 
&NA  
&11.67 million  
& 2.10 million 
&13.76 million  
&Anomalies include abuse, arrests, arson, assaults, road accidents, burglary, explosions, fighting, robbery, shooting, stealing, shoplifting, and vandalism. 
&\url{https://webpages.uncc.edu/cchen62/dataset.html}
\\\hline

\multirow{16}{*}{\rotatebox{77}{\scriptsize \textbf{2019} \scriptsize}}
&FDST   \cite{FangZCGH19}  
&Captured from 13 different scenes.   
&13 
&8.4 minutes 
&NA   
&60   
&40   
&100  
&\tiny{$1080\times$$1920$  $720\times$$1280$}  
&Hu-man 
&Frame  level 
&394081 annotated heads  
&\,\, 9000  
& 6000  
&15000  
&It is basically a  crowd-counting video dataset with frame-wise ground truth annotation. 
&\url{https://github.com/sweetyy83/Lstn_fdst_dataset}
\\\cline{2-18}

&IITB-Cor-ridor \cite{RodriguesEtAlcorr2019}  
&Videos captured on a college campus.   
&1 
&NA 
&108278 frames  
&NA   
&NA   
&NA   
&\tiny{$1080\times$$1920$}   
&Software \cite{NajibiSCD17} Hu-man 
&Frame  level 
&108278  
&\,\,\,301999  
&\,\,181567  
&\,483566  
&Anomalies involve a single person or a group of persons (e.g., chasing, fighting, and sudden running). 
&\url{https://rodrigues-royston.github.io/Multi-timescale_Trajectory_Prediction/}
\\\cline{2-18}

&MVTec   \cite{BergmannEtAl2019mvtecad}  
&Industrial RGB sensor.   
&15 categories 
&NA 
&70   
&NA   
&NA    
&NA   
&\tiny{$700\times$$700$  $1024\times$$1024$}  
&Hu-man 
&Pixel  level 
&\,\,\,1900  
& 3629  
& 1725  
& 5354 images  
&Anomalies are defects including scratches, dents, contamination, and structural changes. 
&\url{https://www.mvtec.com/company/research/datasets}
\\\hline

\multirow{45}{*}{\rotatebox{77}{\scriptsize \textbf{2020} \scriptsize}}
&Street Scene \cite{RamachandraJ20}  
&Static USB camera, bike--pedestrian ways. 
&17 
&226 minutes 
&205 
&46   
&35   
&81  
&\tiny{$720\times$$1280$ } 
&Hu-man 
&Frame--Pixel levels 
&NA  
&56847  
&146410 
&203257  
&Cars driving, turning, stopping, and parking; pedestrians walking, jogging, and pushing strollers; bikers riding in bike lanes. 
&\url{https://www.merl.com/demos/video-anomaly-detection}
\\\cline{2-18}

&City SCENE \cite{CitysceneDataset2020}   
&Real-world surveillance videos.   
&NA 
&468  minutes 
&12   
& 758     
&1507     
&2265    
&\tiny{NA}  
&Hu-man 
&Video level  
&NA  
&NA  
&NA  
&NA  
&Anomaly events are quarreling, fighting, robbery, vandalism, stealing, explosions, falling, smog, littering, smoking, walking a dog, etc. 
&\url{https://cityscene.github.io/#/}
\\\cline{2-18}

&AI-City Challenge 20:T4 \cite{Naphade20AIC20}  
&Provided by the Department of Transportation of Iowa.   
&NA 
&3000 minutes 
&100+   
&100    
&100    
&200    
&\tiny{$410\times$$800$}  
&Hu-man 
&Frame level 
&\,\,NA  
&3000 
&3000  
&6000 
&The anomalous behaviors mainly constitute vehicles driving off-road, stalled vehicles, and crashes \cite{DoshiY20a}. A total of 21 anomalies exist in the training dataset. 
&\url{https://www.aicitychallenge.org/2020-data-and-evaluation}
\\\cline{2-18}

&RWF 2000 \cite{ChengCL20}
&\url{www.youtube.com} video clips captured by surveillance cameras from various real-world scenes.   
&\,\,\,\,\,NA 
&166.67 minutes 
&NA   
&1600    
& 400    
&2000    
&Variable  
&Hu-man 
&Video level 
&\,\,30\%  
&240000 
& 60000 
&300000 
&It is a large-scale video database for violence detection. Each video clip is labeled as violent or non-violent behavior. Several videos would not have high imaging quality due to the dark environment, fast movement of objects, or lighting blur \cite{ChengCL20}. 
&\url{https://github.com/mchengny/RWF2000-Video-Database-for-Violence-Detection}
\\\cline{2-18}

&JHU CRO
WD++  \cite{SindagiEtAl2020}  
&Collected from various geographical spots.   
&Set of scenes 
&NA   
&NA   
&NA   
&NA   
&NA   
&Variable But on the average \tiny{$910\times$$1430$}   
&Hu-man 
&Image level 
&1.515 million  
&2722 images 
&1600 images  
&4372 images 
&It is basically a  crowd-counting video dataset with annotations including dots, approximate bounding boxes, and blur levels. 
&\url{http://www.crowd-counting.com/}
\\\hline

\multirow{28}{*}{\rotatebox{77}{\scriptsize \textbf{2021} \scriptsize}}
&AI-City Challenge 21:T4 \cite{Naphade21AIC21}   
&Multiple intersections and highways in Iowa.   
&NA 
&3720 minutes 
&100+   
&100    
&150    
&250    
&\tiny{$410\times$$800$}  
&Hu-man 
&Frame level 
&\,\,NA    
&3000  
&4500  
&7500  
&Traffic anomalies consist of single- or multiple-vehicle crashes and stalled vehicles.  A total of 18 anomalies exist in the training dataset. 
&\url{https://www.aicitychallenge.org/2021-data-and-evaluation}
\\\cline{2-18}

&Neuro AED  \cite{ChenLLTHDCK21}   
&Stationary neuromorphic vision sensor DAVIS346.   
&4 
&Mean 31 minutes 
&NA   
&82     
&70     
&152    
&\tiny{$260\times$$346$}  
&Hu-man 
&Pixel--Slice levels 
&NA  
&NA  
&NA  
&NA  
&People suddenly scattering in different directions, wrong direction and running events, non-pedestrian moving objects (e.g., bike and motorcycle). 
&\url{https://ispc-group.github.io}
\\\cline{2-18}

&SHADE \cite{LinGWL21}   
&Synthetic data generator.   
&9     
&245 minutes     
&NA     
&1701     
& 448     
&2149   
&\tiny{$1080\times$$1920$}  $240\times472$  
&Soft-ware \cite{GTA5} 
&Region level 
&\,\,\,NA 
&703946   
&175986   
&879932  
&Anomalies include an arrest, a chase, a fight, a knockdown, running, shooting, and scattering.
&NA
\\\cline{2-18}

&X-MAN \cite{SzymanowiczCC21}   
&Taken from UCSD Ped2 \cite{ChanLV08},  CUHK-Avenue~\cite{LuSJ13},
Shanghai Tech Campus ~\cite{LuoLG17}.   
&NA      
&NA      
&55      
&NA      
&NA      
&NA      
&NA     
&Hu-man 
&Region level 
&\,\,40618 
&\,\,NA   
&NA   
&22722  
&It aims to detect various anomalies in video clips, while automatically explaining reasons behind the response of the detector, e.g.,  some consecutive frames are anomalous due to people fighting \cite{SzymanowiczCC21}.
&\url{http://www.jjcvision.com/projects/explaining_anomalies.html#about}
\\\hline

\multirow{7}{*}{\rotatebox{77}{\scriptsize \textbf{2022} \scriptsize}}
&NOLA  \cite{DoshiEtAlwacv2022a}
&Videos from a single moving camera.   
&1     
&800 minutes     
&NA      
&110     
& 50     
&160   
&\tiny{$720\times$$1280$}  
&Hu-man 
&Region--Temporal levels 
&\,\,NA 
&990000   
&450000   
&1.44 million  
&Both spatial and temporal anomalies are labeled, e.g., loitering is considered nominal during the daytime;
yet, it is deemed anomalous in the course of the night. 
&\url{https://github.com/kevaldoshi17/NOLA}
\\\hline

\end{longtable}
\twocolumn
\normalsize

\begin{figure*}
\centering
{\small
\begin{forest}
    for tree={
      grow=east,
      draw=blue!80!darkgray,
      parent anchor=south east,
      child anchor=south west,
      anchor=south,
      align=center,
      l sep+=25.5pt,
      s sep+=-5pt,  
      inner sep=0.25pt, 
      outer sep=0.50pt,  
      edge path={
        \noexpand\path [draw, rounded corners=0pt, \forestoption{edge}] (!u.parent anchor) [out=0, in=180] to (.child anchor)\forestoption{edge label} -- (.south east);
      },
      for root={        ellipse,
        draw,
        parent anchor=east,
      },
    }
    [Deep\\Crowd\\Anomaly\\Detection\\Methods,root color={brown}
      [U-Net, parent color={lime}
        [Classical U-Net]
        [Multi-scale U-Net]
        [ST-U-Net]
        [VQ-U-Net]
        [Non-local U-Net]
      ]
      [CNNs, parent color={blue}
        [Pre-trained 2DCNN]
        [Pre-trained 3DCNN]
        [Fine-tuned CNN]
      ]
      [YOLO, parent color={cyan}
        [YOLOv1]
        [YOLOv3]
        [YOLOv4]
      ]
      [GANs, parent color={red}
        [Standard-GAN]
        [LeastSquare-GAN]
        [Relativistic-GAN]
        [NM-GAN]
        [DE-GAN]
        [PatchGAN]
        [P-GAN]
        [C-GAN]
        [BR-GAN]
      ]
      [AN, parent color={yellow}
        [Attention-Driven Loss]
        [Self-Attention]
      ]
      [AEs, parent color={green}
        [DAE]
        [VAE]
        [AEVB]
        [DpAE]
        [AAE]
        [2D-CAE]
        [3D-CAE]
        [TS-AE]
        [MoAE]
        [A3D-CAE]
        [CdAE]
        [RNN-AE]
      ]
      [LSTMs, parent color={pink}
        [Classical LSTM]
        [CNN-LSTM]
        [ConvLSTM]
        [SFA-ConvLSTM]
        [Social-LSTM]
      ]
      [Sundry, parent color={teal}
        [GCN]
        [RPN]
        [SCN]
        [HTM]
        [MLAD]
        [MatchNet]
        [TNN]
        [QCNN]
      ]
      [Hybrid, parent color={purple}
        [GAN-AE-3DCNN]
        [U-Net-AE-GAN]
        [ConvLSTM-GAN-AE]
      ]
    ]
  \end{forest}
  }
\caption{Taxonomy of deep crowd anomaly detection models based on principle part of each model in
Table~\ref{SummaryOfLiteratureReview2020to2021}.}
\label{DCADMtaxonomy}
\end{figure*}
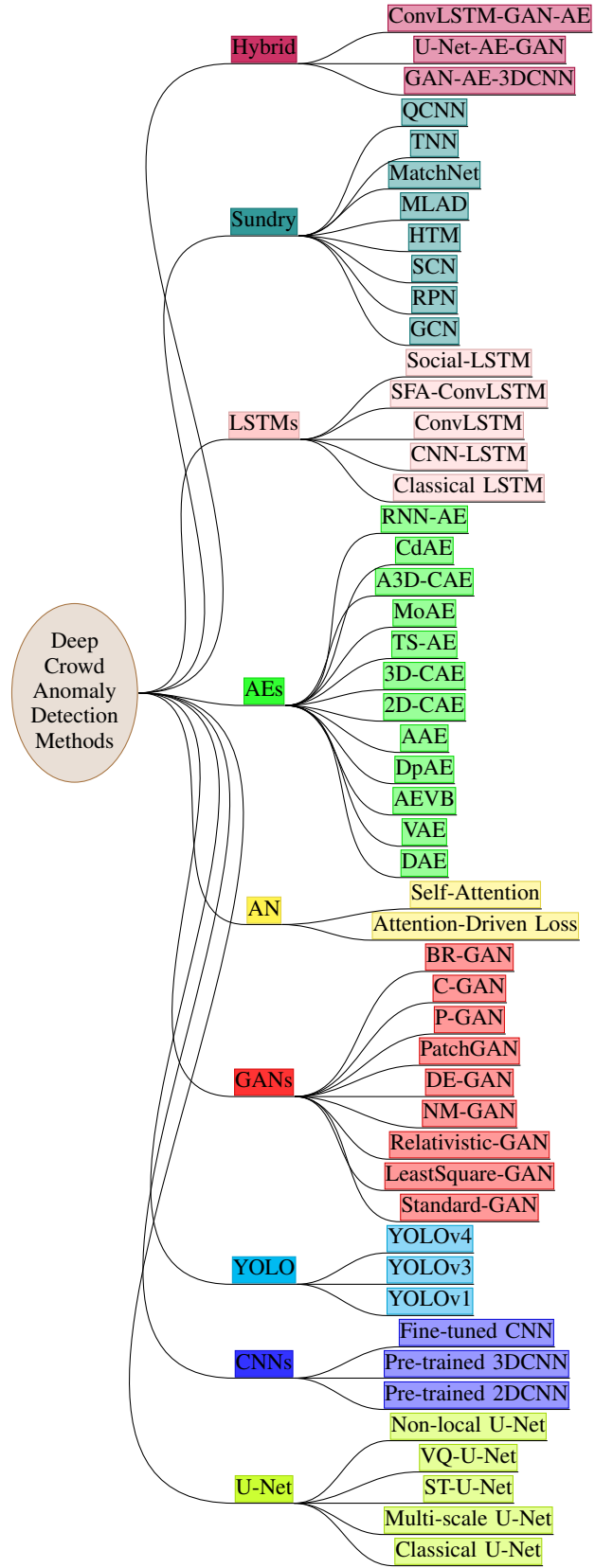

\section{Succinct Survey of Deep Crowd Anomaly Detection}\label{DeepMethodsForCrowdAnomalyDetection}
There are different types of deep learning algorithms but not all kinds of algorithms are employed in crowd anomaly detection.
Recently, a long list of deep learning algorithms that achieved actual and valuable results for crowd anomaly detection appeared in the literature. This section reviews the literature available from 2020 to 2022.
First of all, we have made a qualitative and quantitative summary in Table~\ref{SummaryOfLiteratureReview2020to2021} for deep learning-based crowd anomaly detection methods. We have classified the existing algorithms based on the information in Table~\ref{SummaryOfLiteratureReview2020to2021}. At the end, we have highlighted our observations.

\subsection{Taxonomy}
Sanchez et al.~\cite{SanchezHTH20} grouped deep anomaly detection models depending on their nature when performing anomaly detection tasks,  e.g., extracting deep features and then applying either OCSVM or Gaussian models, training a deep model capable of reconstructing the original image from its compressed representation and then detecting an anomaly from the reconstruction, making use of an anomaly score directly (i.e., end-to-end) without utilizing deep models as feature extractors, and etc.
While the taxonomy of Sanchez et al.~\cite{SanchezHTH20} entails a common view of crowd behavior analysis, our work uses it differently and comprehensively by only considering more recent studies.
Our taxonomy is based on the used key modules or techniques (e.g., CNN, LSTM, AE, etc.) of each model. According to the information shown in  Table~\ref{SummaryOfLiteratureReview2020to2021}, we have primarily classified the deep crowd anomaly detection algorithms into CNN-, LSTM-, AE-, GAN-, U-Net-, YOLO-, and AN-based models. Usually, a deep crowd anomaly detection model consists of more than one module or technique,  and sometimes the roles of modules are inseparable when sorting out their principal parts. This makes the categorization of deep models arduous. The taxonomy presented in Figure~\ref{DCADMtaxonomy} is based on the principal part of each model in Table~\ref{SummaryOfLiteratureReview2020to2021}. For example, Hasan et al.~\cite{Hasan0003CNRD16} applied an AE as a GAN for utilizing frame-level generation errors~\cite{WangYZ20a} to detect anomalies. In this case, we assume GAN as the principal module and thus the model belongs to the GAN category. When the roles of two or more principal modules of a model have similar weights, the model falls into a hybrid category. The sundry category contains many algorithms of different kinds.

\subsection{CNNs- or ConvNets-based Methods}
CNNs or ConvNets were introduced in the 1980s by Yann LeCun. The early version of ConvNet was called LeNet, named after LeCun, and could recognize handwritten digits~\cite{LeCunEtAl1989}.
Currently, a CNN is a popular model in computer vision, and it has the superiority of making good use of the correlation information of data.
It is mainly composed of convolution layers, most often followed by an activation function, pooling layers, and some fully connected layers. The convolutional layer is the core building block of a CNN, and it is represented as the backbone of many deep learning models. CNN adopts the convolution product as the main procedure for each layer. Convolutional layers convolve the input and pass its result to the next layer. The convolutional and pooling layers of CNN can function on the data with a Euclidean or grid-like structure (e.g., images).
Two accepted options exist in CNN while training images. The first option is to train the domain-specific problem statement from scratch. The second option is to use a pre-trained model, which is frequently referred to as transfer learning~\cite{TripathiEtAl2020}.
CNN has a number of parameters called \textit{hyperparameters}, which determine the network structure (e.g., the number of hidden units). Hyperparameters are variables including the number of hidden layers, the learning rate, the batch size, or the number of epochs.
The depth of a network is a crucial feature to consider when attempting to achieve optimal outcomes~\cite{MansourEGVL21}. The number of layers used in deep learning range from five to more than a thousand~\cite{Ferentinos18,SzegedyLJSRAEVR15,SimonyanZ14aVGG16}. The values of the hyperparameters are configured before training the network~\cite{Radhakrishnan2021}. The features picked up by pre-trained CNN models can be quickly employed for a number of different problems that they were not initially designed to resolve~\cite{GutoskiRHALL21}. Choosing a suitable CNN structure is important in the trained model~\cite{MansourEGVL21}. An Adam optimizer~\cite{KingmaB14} is frequently utilized for CNN as it performs significantly better than the Nesterov moment optimizer~\cite{HurEtAl2021corr}.

\subsubsection{Pre-trained 2DCNN} 
Instead of constructing a model from scratch to solve a similar problem, we can use a pre-trained model that is already built and trained to solve certain problems. For example, VGGNet16~\cite{SimonyanZ14aVGG16} is a pre-trained CNN model that can be employed to extract spatial features and for high-accuracy image recognition because of the depth of its network~\cite{VigneshYS17}. Ahmed et al.~\cite{AhmedEtAl2021} used VGGNet16~\cite{SimonyanZ14aVGG16} for crowd detection and an analysis of gender, as well as for analyzing ages.
As using VGGNet16~\cite{SimonyanZ14aVGG16} makes it difficult to represent the temporal relationship of the input video sequences accurately, Ye et al.~\cite{YeDYLD20} employed VGGNet16~\cite{SimonyanZ14aVGG16} to extract the spatial features, and then the obtained feature maps were fed into LSTM to further extract the temporal features of the input video clips.
Al-Dhamari et al.~\cite{Al-DhamariSM20} applied the pre-trained VGGNet19 model to extract descriptive features.
They showed that VGGNet19~\cite{SimonyanZ14aVGG16} had higher detection accuracy than other pre-trained networks using the UCSD Ped1~\cite{ChanLV08} and UMN~\cite{UMNdataset2021}  datasets, such as GoogleNet~\cite{SzegedyLJSRAEVR15}, ResNet50~\cite{HeZRS16}, AlexNet~\cite{KrizhevskySH12}, and VGGNet16~\cite{Al-DhamariSM20}.
Rezaei et al. ~\cite{RezaeiY21} combined the advantage of both the pre-trained VGGnet~\cite{SimonyanZ14aVGG16} and a multi-layer non-negative matrix factorization for anomaly detection in crowd video surveillance.
AlexNet~\cite{KrizhevskySH12} is considered a significant breakthrough in the computer vision area~\cite{KrizhevskySH17}. It reduced the error rate of classification from 26\% to 15\%, which is a major improvement~\cite{Al-DhamariSM20}. The knowledge transfer of the low-level features from AlexNet~\cite{KrizhevskySH12} is entirely feasible~\cite{FranzoniBM20}. To this end, Almazroey et al.~\cite{AlmazroeyJ20} employed the pre-trained AlexNet~\cite{KrizhevskySH12} to extract high-level features for abnormal events and behavior detection in crowd scenes.
Panget al.~\cite{PangYSH020} performed an initial detection to create pseudo anomalous and normal frame sets from a set of unlabeled videos, and then those sets were used to train a (pre-trained) ResNet50 model~\cite{HeZRS16} and a fully connected network in an end-to-end fashion.
Ilyas et al.~\cite{IlyasAQBH21} extracted spatiotemporal deep features from video frames using two ResNet101 models~\cite{HeZRS16} for crowd anomaly detection.
Upon performing panoptic segmentation~\cite{KirillovGHD19}, Wu et al.~\cite{WuSTH20} used pre-trained models (e.g., ResNet101~\cite{HeZRS16}) to generate high-level features from video streams.
Mansour et al.~\cite{MansourEGVL21} employed ResNet~\cite{HeZRS16} as the baseline network to act as an effective feature extractor for intelligent video anomaly detection and classification.
Doshi et al.~\cite{DoshiY21cvprwork} trained a Squeeze and Excitation Network~(SENet)~\cite{HuEtAlcorr2017} with a depth of 152 for anomaly detection in traffic videos.
Aljaloud et al.~\cite{AljaloudU21} compared the performance of their irregularity-aware semi-supervised deep learning model, the baseline CNN model, and the WideResNet model~\cite{ZagoruykoK16} for the detection of unusual events.
Tripathi et al.~\cite{TripathiEtAl2020} applied a variation of the Xception~\cite{Chollet17} model for crowd emotion analysis.
For the detection of anomalies in crowd videos, Mehmood~\cite{Mehmood21} used a two-stream CNN structure~\cite{SimonyanZ14} consisting of a spatial and temporal stream. Modified Xception~\cite{Chollet17} was used as the pre-trained 2D CNN in each of the two streams, and it outperformed Inception-v1~\cite{SzegedyLJSRAEVR15} and DenseNet~\cite{HuangLMW17}.
Bahrami et al. used a single-frame anomaly detection model~\cite{BahramiPVS21} that classified images by reconstructing
them using a pre-trained Xception~\cite{Chollet17} encoder and a decoder similar to Xception~\cite{Chollet17}. However, the order of the layers was partially reversed.
To detect video anomalies, Gutoski et al.~\cite{GutoskiRHALL21} performed a comparative study of transfer learning approaches using 12 different pre-trained 2DCNN models (e.g., GoogleNet~\cite{SzegedyLJSRAEVR15}, ResNet50~\cite{HeZRS16}, AlexNet~\cite{KrizhevskySH12},  VGGNet~\cite{SimonyanZ14aVGG16}, InceptionV3~\cite{SzegedyLJSRAEVR15}, DenseNet~\cite{HuangLMW17}, etc.)  and 7 benchmark datasets.
Chen et al.~\cite{ChenEtAlwacv2022} adopted a 2D convolutional backbone (e.g., ResNet101~\cite{HeZRS16}) for extracting the general-purpose feature vector. Gudovskiy et al.~\cite{GudovskiyEtAlwacv2022} applied ResNet~\cite{HeZRS16} and WideResNet~\cite{ZagoruykoK16} as examples of their CNN feature extractors. Using VGGnet16~\cite{SimonyanZ14aVGG16}, Tsai et al.~\cite{TsaiEtAlwacv2022} trained their model to
extract representative patch features from normal images.

\subsubsection{Pre-trained 3DCNN}
When applying 2DCNN to detect the anomalous events in video sequences of crowd scenes, thereby neglecting 
temporal-domain behavior characteristics. In 3D convolution, a 3D filter can move in all three directions (e.g., height, width, and channel of the image). The 3DCNNs are structured by changing the internal convolution layers from 2D to 3D operations~\cite{LeeKKL21}. In 3DCNN, convolution and pooling operations are performed spatiotemporally while in 2DCNNs they are done only spatially. The 2DCNN loses the temporal information of the input signal immediately after every convolution operation~\cite{TranBFTP15}. Although the 3D kernels tend to overfit because of a large number of parameters, the performance of 3DCNNs is greatly improved by using larger video databases~\cite{HaraKS17}. The 3DCNNs are more suitable for spatiotemporal feature learning compared with 2DCNNs. Thus, their performance is better than 2DCNNs. Explicitly, 3DCNN models can encapsulate information related to shapes and motions in video sequences better than 2DCNN-based models, thus boosting the anomaly detection accuracy in 3DCNN models.
Li et al.~\cite{LiLG20} proposed an anomaly detection algorithm based on 3D fully convolutional networks (e.g., 3DCNN~\cite{TranBFTP15}).
After background subtraction, Nasaruddin et al.~\cite{NasaruddinMAD20} fed the attention regions into a
pre-trained feature extractor (e.g., 3DCNN~\cite{TranBFTP15}) for deep anomaly detection.
Zaheer et al.~\cite{ZaheerMSL20} computed spatiotemporal features by employing a pre-trained feature extractor model (e.g., 3DCNN~\cite{TranBFTP15}) for crowd anomaly detection.
Zahid et al.~\cite{ZahidTDB20} segmented videos into 60-frame clips to localize anomalies temporally using Inception-v3~\cite{SzegedyLJSRAEVR15} and a pre-trained 3DCNN feature extractor~\cite{TranBFTP15}.
Hu et al.~\cite{HuZLS20} used a pre-trained 3D VGGNet16~\cite{SimonyanZ14aVGG16} model for anomaly detection and location in crowded scenes.
Sarker et al.~\cite{SarkerLRFL21} extracted spatiotemporal features for each video segment using a T-C3D feature extractor~\cite{LiuLGTM18}. The T-C3D model is pre-trained using the Kinetics dataset~\cite{SarkerLRFL21}.

The architecture of 3DCNNs is relatively shallow compared with many deep neural networks in 2D-based CNNs  (e.g., ResNets)~\cite{HaraKS17}. The 2D ResNets introduce shortcut connections that bypass a signal from one layer to
the next. Furthermore, Hara et al.~\cite{HaraKS17} simply extended from the 2D-based ResNets to the 3D ones to generate a standard pre-trained model for spatiotemporal recognition. The difference between 3D ResNet~\cite{HaraKS17} and the original 2D ResNet~\cite{HeZRS16} is the number of dimensions of convolutional kernels and pooling, i.e., 3D ResNets~\cite{HaraKS17} perform 3D convolution and pooling. The model trained on the Kinetics dataset~\cite{KayCSZHVVGBNSZ17} performs well without overfitting despite a large number of model parameters~\cite{HaraKS17}.
Despite the very high computational time required to train the 3D ResNets (e.g., three weeks for the Kinetics dataset~\cite{KayCSZHVVGBNSZ17})~\cite{HeZRS16}, some authors (e.g.,~\cite{GaoEtAlcorr2021,LinGWL21}) applied 3D ResNets for anomaly detection in crowds.
Hao et al.~\cite{HaoLWWG22} also employed 3D ResNet~\cite{HaraKS17} for video anomaly detection.
However, Degardin et al.~\cite{DegardinP21} described an iterative learning model using 3DCNN~\cite{TranBFTP15} for abnormal events detection.
Zheng et al.  \cite{ZhengZZLL22} adopted a two‐stream 3DCNN  as the backbone network for their video classification tasks.
Zhao et al.  \cite{ZhaoMSG22} and Zhang et al.  \cite{ZhangHLX22} also differently employed 3DCNN for video anomaly detection.

\subsubsection{Fine-tuned CNN}
Fine-tuning is an iterative process that reduces error rates. It takes a pre-trained model for a given task and then tweaks it for a second similar task.
Singh et al.~\cite{SinghRVTKW20} used an ensemble of different fine-tuned CNNs based on the hypothesis that dissimilar CNN architectures learn various levels of semantic representations from crowd videos and, hence, an ensemble of pre-trained CNNs (e.g., AlexNet~\cite{KrizhevskySH12}, GoogLeNet~\cite{SzegedyLJSRAEVR15}, and VGGNet~\cite{SimonyanZ14aVGG16} via SGD~\cite{RobbinsMonro1951} and AGD~\cite{DuchiHS11}) can enable enriched feature sets to be extracted.
Ullah et al. \cite{UllahUHMHSBA22} employed a self-pruned fine-tuned lightweight CNN for classifying normal or anomalous events.

\subsection{LSTMs-based Methods}
The LSTM~\cite{HochreiterS97} units are an advancement of the generic building blocks of the RNN~\cite{NawaratneASY20}.  An LSTM is designed to model temporal sequences. It keeps unique units called memory blocks in the recurrent hidden layer. The memory blocks hold memory cells with self-connections, which assist in storing the temporal state of the network along with gates (e.g., special multiplicative units) for managing the flow of information. Each memory block in the original architecture contained an input and output gate. The input gate manages the flow of input activations into the memory cell. The output gate supervises the output flow of cell activations into the rest of the network. The forget gate was attached to the memory block~\cite{GersSC00}. An LSTM is mapped to work more distinctively than a CNN because an LSTM is customarily used to process and make predictions from a given sequence of data. In many anomaly detection methods for videos, time modeling mostly adopted 3D convolution~\cite{ZhaoDSLLH17} and ConvLSTM structures~\cite{ShiCWYWW15,ChongT17,LuoLG17ICME}. In spite of that, methods using a 3D convolution need to calculate a large number of parameters and perform time-consuming training~\cite{CaiLGHL21}.  As a result, LSTM can be more suitable for temporal information modeling.

\subsubsection{Classical LSTM}
Xia et al.~\cite{XiaL21} used LSTM~\cite{HochreiterS97} to decode historical feature sequences with temporal attention for predicting the features. However, Xia et al.~\cite{XiaL21a} applied LSTM~\cite{HochreiterS97} to predict current-dimension-reduced HOG (histogram of oriented gradients) as the appearance feature and HOF (histogram of optical flow) as the movement feature. Moustafa et al.~\cite{MoustafaG20} proposed an LSTM-based approach for pathway and crowd anomaly detection, where the crowd scene was divided into a number of static-overlapped spatial regions.

\subsubsection{CNN-LSTM}
Deep CNNs are a kind of common deep neural network that is suitable for spatial relationship learning on raw input data.
CNNs, RNNs, and other deep learning models can learn better feature representations than hand-crafted feature models~\cite{YeDYLD20}. 
For example, Arifoglu et al.~\cite{ArifogluEtAl2019} combined 2D-CNN and LSTM to detect abnormal behavior in dementia sufferers. 
Among numerous CNN models, a CNN called VGGNet16~\cite{SimonyanZ14aVGG16} can be used to extract spatial features and for high-accuracy image recognition due to the depth of the network~\cite{VigneshYS17}.
However, it is difficult for VGGNet16~\cite{SimonyanZ14aVGG16} to accurately represent the temporal relationship of the input video sequences. To overcome this limitation,  Ye et al.~\cite{YeDYLD20} applied 2D-VGGNet16 and LSTM models to extract the spatiotemporal features of video frames and then constructed the feature expectation subgraph for each key frame of every video. Esan et al.~\cite{EsanEtAl2020} proposed a CNN-LSTM model for anomaly detection in crowded environments, where CNN was applied to extract the features from the image frames and LSTM was used as a mechanism for remembrance to make quick and accurate detections. Sabih et al.  \cite{SabihV22} combined CNN and LSTM to solve crowd anomaly detection problems.

\subsubsection{ConvLSTM}
In 2015, Shi et al.~\cite{ShiCWYWW15} proposed ConvLSTM for obtaining convolutional structures in both the input-to-state and state-to-state transitions by extending the fully connected LSTM~(FC-LSTM)
. The difference between ConvLSTM~\cite{ShiCWYWW15} and FC-LSTM~\cite{Graves13corr,SrivastavaMS15} is that the matrix operations of FC-LSTM are replaced with convolutions. This operation enables ConvLSTM~\cite{ShiCWYWW15} to perform better with images than FC-LSTM~\cite{ZhouWLZT20}.
As the LSTM encoder-–decoder generally fails to account for the global context of the learned representations with a fixed dimension
representation, Zhou et al.~\cite{ZhouWLZT20} proposed two composite LSTM encoder--decoder models with a ConvLSTM~\cite{ShiCWYWW15} unit to learn spatiotemporal features and detect abnormal events.
Nawaratne et al.~\cite{NawaratneASY20} employed a series of CNN layers to learn spatial representations, as well as
a series of ConvLSTM~\cite{ShiCWYWW15} layers to learn temporal representations for anomaly detection.
Li et al.~\cite{LiEtAL2020mlcorr} applied a motion model to efficiently detect anomalies in surveillance-video footage. They replaced the 3D convolutional layers of their motion model with ConvLSTM~\cite{LuoLG17ICME} to study the performance of their model. They observed that the model employing ConvLSTM~\cite{LuoLG17ICME} layers had delayed detection results and failed to detect the beginning of anomalous events. Furthermore, it also reported more false alarms after the anomalous events due to its slow response
. However, the 3D convolutional layers might fit the data better and thus extract more representative features.
As ConvLSTM is capable of forecasting behaviors and remembering the spatial data over time to avoid any crowd-related catastrophe, Varghese et al.~\cite{VargheseEtAl2020} devised a ConvLSTM-based  model for predicting nine
distinct crowd behaviors learned from their fuzzy computational model.
Liu et al. \cite{LiuEtAl2022} adopted a bidirectional ConvLSTM \cite{SongWZSL18} for handling temporal information when detecting video anomalies
.

\subsubsection{SFA-ConvLSTM}
SFA-ConvLSTM learns multi-scale motion information. It performs attention encoding to focus on movement regions in surveillance environments where events occur locally. The fields of spatiotemporal encoding can be controlled by adjusting the dilation rate of its convolutional kernels. The anomaly detection model developed by Lee et al.~\cite{LeeKR20} consisted of a stack of four SFA-ConvLSTMs, a bidirectional multi-scale encoder, a scale-selective aggregator (e.g., ConvLSTM~\cite{ShiCWYWW15}), and a spatial decoder.

\subsubsection{Social-LSTM}
Inspired by the application of RNNs in variegated sequence forecasting tasks~\cite{GravesMH13,VinyalsTBE15,CaoLYYWWHWHXRH15}, Alahi et al.~\cite{AlahiGRRLS16} introduced Social-LSTM for human trajectory forecasting in crowded spaces. Social-LSTM~\cite{AlahiGRRLS16} is an LSTM~\cite{HochreiterS97} with a novel social-pooling layer that captures the social interactions of nearby pedestrians. Kothari et al.~\cite{KothariEtAl2021} also employed a Social-LSTM-based model for human trajectory forecasting in crowds~\cite{AlahiGRRLS16}.

\subsection{AE-based Methods}
AEs were first introduced in the 1980s by Hinton and the PDP group~\cite{RumelhartEtAl1987} to describe the problem of back-propagation without a teacher by considering the input data as the teacher~\cite{Baldi2012}. Currently,  AEs are used to learn efficient codings of unlabeled data in deep architectures for transfer learning. An AE consists of two key parts, namely an encoder and a decoder. The encoder maps the input into the code, whereas the decoder maps the code to a restoration of the input. In unsupervised anomaly detection, the AE is trained on normal segments by minimizing their reconstruction errors~\cite{ShiXFZP21}, and then, a thresholded reconstruction error is utilized to detect anomalies. It is generally assumed that the reconstruction error will be lower for the normal segments because they are close to the training data, whereas the reconstruction error becomes higher for the abnormal segments~\cite{Hasan0003CNRD16,GongLLSMVH19,ChoKKCL22}.

\subsubsection{DAE}
DAE is a variant of the basic AE~\cite{Vincent11}.  DAE is trained by reconstructing a clean input by a corrupted input. A basic DAE is learned by minimizing the loss function. Deep DAE can be achieved by using multiple hidden layers that can learn the complicated distribution by given samples due to their multiple feature-representation spaces~\cite{LuTMH13}. The backpropagation algorithm~\cite{LeCunEtAl2015} is used to train DAE~\cite{WuSTH20}.
For anomaly detection, Wu et al.~\cite{WuEtAl2021cc} mainly focused on the behavior analysis of pedestrians by applying DAE. They applied pre-trained deep models to extract high-level concept and context features for training DAE that required a short training time (i.e., within 10 seconds on UCSD datasets~\cite{ChanLV08} running on a computer with the 64-bit Windows 10 OS and equipped with 16 GB DDR4 RAM and an Intel Core i7-9750H CPU at 2.60 GHz) while achieving comparable detection performance with alternative methods.

\subsubsection{VAE}
Sabokrou et al.~\cite{SabokrouEtAl2016} introduced a cascaded anomaly detection model, which detected abnormal events based on the reconstruction error of the standard autoencoder. However, such types of methods are based on deep reconstruction and treat samples that are different from normal samples as anomalies. It ignores the small probability of abnormal events. A large number of normal samples are often misjudged as abnormal, leading to false alarms~\cite{Ma21}.
VAE can provide an effective solution in this respect.
VAE is a directional probability graph model that describes the potential space state in a probabilistic way. Similar to the architecture of traditional autoencoders, VAE also includes two neural networks: a probabilistic and a generative decoder. VAE employs a backpropagation algorithm to train the model.
Ma~\cite{Ma21} assumed that the distribution of all normal samples complied with a Gaussian distribution, with the abnormal sample appearing with a lower probability in this Gaussian distribution. Therefore, Ma~\cite{Ma21} proposed an end-to-end deep learning framework based on VAE for abnormal event detection.
Shi et al.~\cite{ShiXFZP21} utilized GRU for the basic encoder and decoder unit to construct a VAE, as well as for the reconstruction probability of the anomaly score. Sharma et al.~\cite{SharmaEtAlwacv2022} proposed a generalized version of the VAE~\cite{KingmaEtAl2014corr} framework for abnormality detection.

\subsubsection{AEVB}
Kingma et al.~\cite{KingmaEtAl2014corr} introduced the AEVB algorithm for an independent and identically distributed dataset, which had continuous latent variables per data point. The AEVB is used to form a probability distribution of normal data by probability inference and reconstruction. Yan et al.~\cite{YanSLZ20} adopted two-stream recurrent AEVB, which provided a semi-supervised solution for abnormal event detection in videos.

\subsubsection{DpAE}
A simple AE contains one hidden layer between the input and output, whereas a DpAE can have multiple hidden layers.
Wang et al.~\cite{WangQZSS20} built a cascaded DpAE based on the deep autoencoder~\cite{WangEtAl2018fae}
for detecting abnormal video events.

\subsubsection{AAE}
Song et al.~\cite{SongSWCJ20} combined an attention-based AE along with a GAN~\cite{GoodfellowPMXWOCB14} model, called Ada-Net, to learn normal patterns for abnormal event detection in an unsupervised way.  The decoder of their used AE is treated as the generator in the GAN. Ada-Net is simple to implement and it produces high-quality models. It is also a TensorFlow-based lightweight framework for learning high-quality models automatically with minimum expert interaction. It can be trained in an end-to-end manner. It provides a comprehensive framework for learning not only neural network design but also how to aggregate models for obtaining even better results.
Le et al. \cite{LeEtAl2022attention} proposed an attention-based residual AE for video anomaly detection, which encoded both spatial and temporal information in a unified way.

\subsubsection{2D-CAE}
A CAE is a variant of CNNs that is developed for the unsupervised learning of convolution filters. It is a feed-forward multi-layer neural network in which the desired output is the input itself. It consists of three convolutional layers, two pooling layers in the encoder, three deconvolutional layers, and two unpooling layers in the decoder, with a symmetric structure~\cite{Yang00GL020}. The 2D-CAE reduces dimensionality and learned temporal regularity~\cite{KuenLL15}. Yang et al.~\cite{Yang00GL020} used 2D-CAE to extract the features of input video frames and to compute reconstruction errors. The reconstruction error is relatively small for the normal frames, while the error is higher for abnormal frames. The 2D-CAE was also utilized for crowd anomaly detection from surveillance videos~\cite{PawarEtAl2021}.
Fan et al.~\cite{FanWLQLX20} employed a fully convolutional network, which did not contain a fully connected layer, allowing the encoder--decoder structure to preserve relative spatial coordinates between the input image and the output feature map. They employed a two-stream network framework to combine the appearance and motion anomalies. They also employed RGB frames for the appearance, whereas they used dynamic flow images for motion. Wang et al.~\cite{WangY22} employed a recurrent type 2D-CAE for video anomaly detection.

\subsubsection{3D-CAE}
The 2D-CAE has a problem with temporal information. For example, Hasan et al.~\cite{Hasan0003CNRD16} employed
two 2D-CAEs and stacked multiple video frames in lieu of an image-channel dimension. They claimed it was more computationally efficient than the approaches that adopted sparse coding on large video datasets. However, after the first
2D convolution operation, the temporal information was completely lost as 2D convolutions were only effective in learning spatial features. To overcome this problem, Asad et al.~\cite{AsadYTCH21} used 3D-CAE for extracting spatiotemporal features. Moreover, Deepak et al.~\cite{DeepakCM21} applied a residual spatiotemporal autoencoder (R-STAE), which consisted of 3D convolution, deconvolution, and ConvLSTM layers~\cite{ShiCWYWW15}, to learn the patterns of normal activities from surveillance videos. R-STAE performed the unsupervised learning of the spatiotemporal representation of normal patterns and reconstructed them with low quantities of errors. Deepak et al.~\cite{DeepakSRC21} also applied a 3D spatiotemporal autoencoder (3D-STAE) that consisted of 3D convolutional and ConvLSTM layers to learn the intricate appearance and motion dynamics involved in training videos.
Hu et al.  \cite{HuLZGJC22} employed 3D-CAE for video anomaly detection.

\subsubsection{TS-AE}
The two-stream hypothesis argues that human beings possess two distinct visual systems~\cite{GoodaleMilner1992}.
This hypothesis is a widely accepted and influential model of vision neural processing~\cite{EyesenckKeane2010}. The two-stream convolutional network developed by Simonyan et al.~\cite{SimonyanZ14} decomposed videos into spatiotemporal components for action recognition. However, the typical two-stream architecture adopted supervised learning characteristics, which required marking incoming frames for abnormal detection. Therefore, it is inconsistent with current anomaly detection datasets~\cite{LiCZZY21}. The AE and two-stream networks can perform well. Nevertheless, the original RGB frames are the input for existing AE frameworks, and they vary in appearance, such as color or light, etc. 
In mitigation this challenge, Li et al.~\cite{LiCZZY21} combined a pre-trained spatial stream and a temporal stream to construct a deep spatiotemporal autoencoder (DSTAE) for anomaly detection. The spatial stream feeds stacked continuous RGB frames into DSTAE to extract the appearance characteristics, whereas the temporal stream feeds stacked continuous optical flow frames into DSTAE to extract the motion patterns.

Xie et al.~\cite{XieSHTM18} used Top-Heavy 3DCNN and Bottom-Heavy 3DCNN. The Top-Heavy model starts with 2DCNN convolutions and ends with 3DCNN, whereas the Bottom-Heavy model starts with 3DCNN convolutions and ends with 2DCNN.
The advantage of combining the dissimilar convolution dimensions is obtaining better results with less training data.
This is crucial for smaller video surveillance datasets. Moreover, 2D convolutional layers are faster than 3D convolutional layers and less prone to overfitting problems. For that reason, the Top-Heavy AE is easier to train and is faster and smaller than a classical 3D AE. Accordingly, Esquivel et al.~\cite{EsquivelG22} employed a Top-Heavy AE model for anomaly detection in video surveillance.
Liu et al. \cite{LiuLLZS22} applied a spatiotemporal-memories-augmented TS-AE framework for video anomaly detection.

\subsubsection{MoAE}
Most of the two-stream-based convolutional networks apply a warped optical flow as the source for motion modeling~\cite{Tu0001XQPVLY18}. Although the motion feature is very useful, the costly computation of optical flow estimation delays the method from being used in many real-time applications. Chang et al.~\cite{ChangTXLZSY22}, inspired by Wang et al.~\cite{Wang0002X00LTG19}, exploited motion representation to simulate the motion of the optical flow, which was directly obtained by the differences of the RGB values between video frames. They employed both U-Net~\cite{RonnebergerFB15} and 2DCNN as the backbone of their motion autoencoder to process consecutive video clips for anomaly detection. Differently, Feng et al. \cite{FengWZ22} designed a MoAE to exploit spatiotemporal features.

\subsubsection{A3D-CAE}
Sun et al.~\cite{SunJSW21} adopted denoising reconstruction errors to train a 3D-CAE.
Nevertheless, they omitted certain higher-order moments, which might produce extra errors.
As a result, they employed an adversarial learning strategy based on GANs~\cite{GoodfellowPMXWOCB14} to train their used autoencoder via an extra discriminator for learning more accurate data distributions. Their 3D-CAE encoder consisted of 3D convolutional layers. Specifically, their A3D-CAE model captured low-level appearance and motion information, which is simultaneously needed for accurate abnormal event detection in videos.

\subsubsection{CdAE}
Zhao et al.~\cite{ZhaoDSLLH17} cascaded the 3D convolution and ConvLSTM operation to a classical AE  for extracting the temporal information of video events. However, they only utilized a single-AE architecture for anomaly detection. Li et al.~\cite{LiCL21} built a better-cascaded classifier for video anomaly detection in crowded scenes, which consisted of a spatiotemporal adversarial autoencoder (STA-AE) and a spatiotemporal convolutional autoencoder (ST-CAE). The STA-AE was composed of a classical CAE and a discriminator~\cite{GoodfellowPMXWOCB14}. It was intended to make the latent representation of inputs match with an arbitrary prior. Both STA-AE and ST-CAE employed 3D convolution and deconvolution in the encoder and decoder of the convolutional autoencoder, respectively. Three-dimensional convolution and de-convolution enhanced the ability of CdAE for extracting effective patterns from temporal dimensions.

\subsubsection{RNN-AE}
The optimization of temporally-coherent sparse coding with the sequential iterative soft-thresholding algorithm~\cite{WisdomPPA16} is equivalent to a special type of stacked RNN (sRNN). Subsequently, Luo et al.~\cite{LuoLLTDPG21} proposed to learn an sRNN-AE for both spatial and temporal features to detect video anomalies. Wang et al.~\cite{WangEtAlcorr2020mpfp} applied a multi-path encoder--decoder architecture with RNNs and explicitly captured the temporal information in object and semantic motions via the multiple ConvGRUs~\cite{BallasYPC15} of different resolutions with non-local blocks~\cite{Wang0004GGH18} for unsupervised video anomaly detection.

\subsection{GANs-based Methods}
Unlike CNNs, GANs take part in a game-theoretic approach.
GANs are unsupervised learning algorithms in their ideal form. Henceforth, no labeled data are needed for their training.
In 2014, Goodfellow et al.~\cite{GoodfellowPMXWOCB14} introduced GANs. The main idea of GAN is that there are two adversaries, namely a Generator and a Discriminator, which are in a ceaseless battle to achieve more error-free predictions throughout their training process.
The generator takes noise as input and generates samples. The discriminator receives samples from the generator and training data. It must be able to distinguish the two data sources. In the training phase, the generator learns to produce a sample that is close to its ground truth. The discriminator learns how to distinguish the generated data from its ground truth. These two networks are trained at the same time, i.e., the generator must not be trained without updating the discriminator. Training a GAN network does not require any Monte Carlo approximations~\cite{HanWYWKDL20}. Adversarial learning~\cite{GoodfellowSS14ICLR} has shown success in improving the generation of image and video~\cite{ZhouZFDPX20}.
A back-propagation algorithm~\cite{HintonEtAl2012corr} is employed to obtain gradients. No inference is needed during learning, and a wide array of functions can be integrated into the GAN~\cite{GoodfellowPMXWOCB14}. These facts about GANs made them effective models for image generation and video prediction, especially in anomaly detection~\cite{XIA2022}.

\subsubsection{Standard-GAN}
As GAN can create data that do not exist in the dataset, the base model can be learned from a lot of data that are very similar to the actual data. Consequently, the problems produced by the short supply of labeled data are minimized. Wang et al.~\cite{WangYZ20a} applied the generation error of a generative neural network to detect anomalies. They first trained a GAN~\cite{GoodfellowPMXWOCB14} to generate normal samples, then judged the samples with large generation errors as anomalies.

\subsubsection{LeastSquare-GAN}
Usually, GANs are known to be successful in creating realistic images and videos. Nevertheless, standard GANs face the vanishing-gradient problem during learning as they hypothesize the discriminator as a classifier with the sigmoid cross-entropy loss function. To overcome this problem, Mao et al. \cite{MaoLXLWS17} used a modified version of GAN~\cite{GoodfellowPMXWOCB14} called LeastSquare GAN \cite{MaoLXLWS17}. Doshi et al. \cite{DoshiY21} employed the LeastSquare GAN \cite{MaoLXLWS17} for detecting online anomalies in surveillance videos.

\subsubsection{Relativistic-GAN}
Zhang et al.~\cite{ZhangWHWW21} utilized an autoencoder as a generator. To avoid the vanishing-gradient problem when combining with the self-attention mechanism, they adopted a discriminator based on a relativistic GAN~\cite{VaswaniSPUJGKP17}. In the training phase, the generator produced a future frame based on the historical clips of a video, and then the predicted future frame with its ground truth was fed into the discriminator. If the requirement of the discriminator was not met, training continued until 
the generator generated a frame that sufficiently confused the discriminator. In the testing phase, they compared the error between the future frame generated by the generator with its ground truth. If the error was greater than a known threshold, the generator failed to predict the development process of the event, which was marked as an abnormal event.

\subsubsection{NM-GAN}
Chen et al.~\cite{ChenYCXJ21} proposed an end-to-end pipeline called NM-GAN for video
anomaly detection. Their NM-GAN combined the reconstruction-based~\cite{RibeiroLL18} and GAN-based approaches~\cite{SabokrouEtAl2018}. Although the reconstruction-based approach is criticized for the uncertainty it brings to the unobserved samples, it is more conducive to obtain results in real-world applications~\cite{ChenYCXJ21}. 

\subsubsection{DE-GAN}
A double-encoder network enables the deep model to generate images of the underlying representation in the training phase. Han et al.~\cite{HanWYWKDL20} proposed a DE-GAN architecture to detect abnormal crowd events.
They removed fully connected hidden layers for deeper architectures and then used ReLU activation in a generator for all layers except for the output. In the discriminator, Tanh and Leaky ReLU activation was employed for all layers.

\subsubsection{PatchGAN}
PatchGAN~\cite{IsolaZZE17} is a type of discriminator for GANs, which only penalizes structures at the scale of local image patches. A GAN~\cite{GoodfellowPMXWOCB14} discriminator maps an input image to a single scalar output in the range of [0,1] for addressing the probability of the image that is either real or fake, while PatchGAN~\cite{IsolaZZE17} provides a matrix as the output with each element for signifying whether its corresponding patch is real or fake~\cite{TangZZGLY20}. For pixel-level tasks, Tang et al.~\cite{TangZZGLY20} followed the PatchGAN discriminator~\cite{IsolaZZE17} to predict the broad locations of abnormal events. Zhang et al.  \cite{ZhangFW22} also employed PatchGAN~\cite{IsolaZZE17} to generate high-quality frames.

\subsubsection{P-GAN}
Anomaly detection methods based on frame prediction (e.g.,~\cite{LuoLG17}) usually use a few previous frames to predict the target frame. Compared with the frame reconstruction methods (e.g.,\cite{SabokrouEtAl2018}), the frame prediction methods consider the anomaly not only in appearance and location but also in motion. Inspired by this advantage, rather than reconstructing training data for anomaly detection, Liu et al.~\cite{LiuLLG18} identified abnormal events by comparing them with their expectations and then introduced a future video-frame prediction-based anomaly detection model referred to as P-GAN~\cite{LiuLLG18}, which was also applied in~\cite{ChenWYZJ20}. P-GAN adopted the U-Net as the generator of GAN to predict future frames~\cite{LiuLLG18}. Park et al.~\cite{ParkNH20} applied four successive video frames to predict the fifth frame using P-GAN~\cite{LiuLLG18} for their unsupervised anomaly detection model. Zhong et al.~\cite{ZhongCJR22} also employed a kind of P-GAN~\cite{LiuLLG18} to detect anomalies in videos.

\subsubsection{C-GAN}
Mirza et al.~\cite{MirzaO14} introduced conditional versions of GANs. The data modes generated in GAN cannot be controlled~\cite{GoodfellowPMXWOCB14}, whereas C-GAN~\cite{IsolaZZE17} involves the conditional generation of images by a generator model
. Moreover, Vu et al.~\cite{VuBAT21} utilized four C-GANs to generate multi-type future-appearance and motion information for anomaly detection in surveillance videos. Cai et al.~\cite{CaiLGHL21} applied a C-GAN~\cite{IsolaZZE17} to optimize predicted-images generation. Compared with ConvLSTM~\cite{ShiCWYWW15}, ConvGRU~\cite{BallasYPC15} has fewer parameters and a similar structure and modeling effects. Therefore, the ConvGRU~\cite{BallasYPC15} module was chosen for time modeling. Ganokratanaa et al.~\cite{GanokratanaaAS22} proposed a variation C-GAN  to enhance the accuracy and quality of the synthesized image.

\subsubsection{BR-GAN}
Yang et al.~\cite{YangLW21} developed a BR-GAN for anomaly detection in videos. Their model consisted of a generator and two discriminators. 
As CAE models are trained separately, they are unable to learn the relation between different local information (e.g., appearance and gradient)~\cite{RoyEtAlcorr2020oca}. Considering this issue, Roy et al.~\cite{RoyEtAlcorr2020oca} proposed an unsupervised two-staged object-centric GAN~\cite{GoodfellowPMXWOCB14} for local anomaly detection in videos. The first stage of their method learns the normal local gradient appearance correspondences, and the second stage learns to classify events in an unsupervised manner.

\subsection{U-Net-based Methods}
A U-Net is a U-shaped structure transformed from a fully convolutional network~\cite{LongSD15}. The first half of the network is used for feature extraction, which is similar to the VGG network~\cite{RussakovskyDSKS15} in structure, and the second half is used for up-sampling. In 2015, Ronneberger et al.~\cite{RonnebergerFB15} proposed the first classical U-Net for biomedical image segmentation. In general, the frame prediction methods outperform the frame generation methods. U-Net~\cite{RonnebergerFB15} plays a vital role in frame prediction because the consecutive frames of one clip of surveillance video usually have the same background and a similar foreground~\cite{ChenWYZJ20}. Although U-Net~\cite{RonnebergerFB15} architecture can be widely employed for the tasks of reconstruction and frame prediction~\cite{LiuLLG18}, the skip connections in it may not be able to extract salient features from the video frames to a great extent for the reconstruction task~\cite{ParkNH20}. 

\subsubsection{Classical U-Net}
Considering the challenges posed by the insufficient utilization of motion patterns, which results in instability on different datasets, Chen et al.~\cite{ChenWYZJ20} proposed a U-Net-based bidirectional prediction model for anomaly detection~\cite{RonnebergerFB15}
. Many AE-based anomaly detection approaches assume that the AE will be unable to accurately reconstruct anomalous regions. However, in practice, neural networks generalize anomalies and reconstruct them efficiently, resulting in reduced detection capabilities. Accurate reconstruction is unlikely if the anomaly pixels were not visible to the AE~\cite{ZavrtanikKS21}. To address this issue, Zavrtanik et al.~\cite{ZavrtanikKS21} used a U-Net-based encoder--decoder network to reconstruct the removed regions~\cite{RonnebergerFB15}. Yu et al.~\cite{YuWCZXYK20} adopted U-Net~\cite{RonnebergerFB15} as the basic network architecture of generative DNNs, which were optimized by the default Adam optimizer~\cite{KingmaB14} in PyTorch~\cite{PaszkeGMLBCKLGA19}.
Zhang et al.~\cite{ZhangNH0Y21} designed a U-Net-based future frame prediction model with two branches~\cite{RonnebergerFB15}---one acted as the discriminator of a GAN~\cite{GoodfellowPMXWOCB14} and the other served as an encoder. During the testing phase, if a frame agreed with its prediction in both the image and latent spaces, there was a high probability it was a normal event. Otherwise, it was likely to be an abnormal event. Lu et al.~\cite{LuYR020} applied U-Net~\cite{RonnebergerFB15} to predict the future frame and passed the prediction to a ConvLSTM~\cite{ShiCWYWW15} to retain the information of the previous steps. To learn high-level features, Leroux et al.~\cite{LerouxEtAlwacv2022} applied a U-Net-type~\cite{RonnebergerFB15} AE that was trained to reconstruct individual input frames. Park et al.~\cite{ParkEtAlwacv2022} employed U-Net~\cite{RonnebergerFB15} to skip connections between the encoder and the decoder, which boosted the generation ability by preventing the vanishing-gradient problem, thereby achieving information symmetry.
Alafif et al. \cite{AlafifACAB022} adopted a U-Net~\cite{RonnebergerFB15} to detect abnormal behavior in Hajj like massive crowd videos.

\subsubsection{Multi-scale U-Net}
Saypadithet al.~\cite{SaypadithO21} employed inception modules and residual skip connections inside the framework for learning higher-level features, named multi-scale U-Net. A multi-scale U-Net was utilized as a GAN to extract a spatial feature. Inception modules were employed inside the multi-scale U-Net to make the network learn higher-level image features. Finally, the multi-scale U-Net reduced the training and testing parameter numbers while considerably improving the anomaly detection accuracy.

\subsubsection{ST-U-Net}
Wang et al.~\cite{WangEtAlcorr2021} were motivated by the prevalent cloze test for learning neural network architectures and ensembles for improved performance. Therefore, they proposed an approach named visual-cloze-completion for video anomaly detection. Explicitly, they designed a ST-U-Net to perform their proposed visual-cloze-completion model. As compared with the standard U-Net~\cite{RonnebergerFB15}, ST-U-Net synthesizes a recurrent network structure to accumulate temporal context information in ST-cubes and produce high-level feature maps to learn richer video semantics.

\subsubsection{VQ-U-Net}
The VQ-U-Net improves the vector quantization module~\cite{OordVK17}, where the encoder network outputs discrete rather than continuous information. 
To learn discrete representations of video by predicting future frames, Szymanowicz et al.~\cite{SzymanowiczEtAlwacv2022} mainly employed baseline U-Net~\cite{RonnebergerFB15}. They formed a VQ-U-Net at the output of the encoder. The output of the encoder was quantized,  which was then configured as the input to the decoder. VQ-U-Net can produce high-quality saliency maps.

\subsubsection{Non-local U-Net}
A non-local operation supports inputs of variable sizes and maintains the corresponding size in the output. Wang et al. \cite{Wang0004GGH18} wrapped the non-local operation into a non-local block. Without changing the initial behavior of any pre-trained model, a new non-local block can be inserted~\cite{Wang0004GGH18}. To this end, Zhang et al.~\cite{ZhangFW22} applied 3 non‑local  blocks in their U‑Net frame prediction model for surveillance-video anomaly detection.

\subsection{YOLO-based Methods}
Object detectors can be grouped into single- and two-stage detectors. Single-stage detectors (e.g., SSD~\cite{LiuAESRFB16} and YOLO~\cite{RedmonDGF16}) solve a simple regression problem, and directly output the bounding box coordinates, whereas two-stage detectors (e.g., Faster R-CNN~\cite{RenHG017}) employ an RPN first to create regions of interest and then perform object classification and bounding box regression. The YOLO only analyzes the image once to detect diverse objects, and it is also very fast.

\subsubsection{YOLOv1}
YOLO~\cite{RedmonDGF16} is a pre-trained object detection system~\cite{DoshiY20}. It is capable of processing higher fps on a GPU while providing the same or even better accuracy compared with ResNet~\cite{HeZRS16} or SSD~\cite{LiuAESRFB16}. As the whole detection pipeline is a single network, it can be directly optimized end-to-end on detection performance.

\subsubsection{YOLOv3}
The YOLOv3 detection rate is 30 fps~\cite{BaiHLWZSY19}, whereas the base YOLO model processes images in real-time at 45 fps~\cite{RedmonDGF16}. Both appearance and motion information, along with modeling latent distributions, are all-important for precisely detecting video anomalies.
Shine et al.~\cite{ShineAV20} developed a method for selecting anomaly candidates by scrutinizing 14 background frames per video using YOLOv3~\cite{RedmonEtAl2018corrYOLOv3}.
Doshi et al.~\cite{DoshiY20a} employed  YOLOv3~\cite{RedmonEtAl2018corrYOLOv3} for anomaly detection in traffic videos, as speed is  an important consideration  when detecting traffic-surveillance anomalies.
Doshi et al.~\cite{DoshiY20} also employed YOLOv3~\cite{RedmonEtAl2018corrYOLOv3} to obtain location and appearance features when detecting objects in real-time from surveillance videos.
Georgescu et al.~\cite{GeorgescuEtAlcorr2020} utilized YOLOv3~\cite{RedmonEtAl2018corrYOLOv3} to detect objects, and then trained a 3DCNN to produce discriminative anomaly-specific information by jointly learning multiple proxy tasks.
Ouyang et al.~\cite{OuyangEtAl2020corr} mainly performed image patch generation, encoding/decoding via two DAEs, density estimation, and anomaly inference to detect video anomalies. The YOLOv3~\cite{RedmonEtAl2018corrYOLOv3} detector was applied to extract patches from the current frame.

\subsubsection{YOLOv4} Doshi et al.~\cite{DoshiEtAlwacv2022b} used the YOLOv4~\cite{BochkovskiyEtAlcorr2020YOLOv4} pre-trained object detector to extract the bounding box (location) and class probabilities (appearance) for each object detected in a given frame. Explicitly, they adopted the extracted bounding boxes to construct a feature embedding to demonstrate the spatiotemporal activities observed in the frame~\cite{DoshiEtAlwacv2022a}.
Shortly after the release of YOLOv4, YOLOv5 was introduced using the Pytorch framework \cite{YOLOv52020}. YOLOv5 outperformed v3 and v4 in terms of ACC using the Microsoft COCO dataset~\cite{LinMBHPRDZ14}. YOLOv5 was used for miscellaneous computer vision tasks. For example, YOLOv5 \cite{YOLOv52020} showed considerable performance with impressive interpretability on drone-captured scenarios \cite{ZhuLWZ21}.

\subsection{AN-based Methods}
Attention mechanisms can quickly extract important features from small amounts of data~\cite{ZhangWHWW21}.
An attention-based model enables the neural network to dynamically shift or select attributes so that the overall decision-making is more reliable~\cite{ZhouZFDPX20}. Currently, attention-based approaches involve a variety of vision-based applications, including segmentation~\cite{ChenYWXY16} and image classification~\cite{WangJQYLZWT17,ZouWJZ022}.

\subsubsection{Attention-Driven Loss}
Zhou et al.~\cite{ZhouZFDPX20} built an attention map by combining a mask map and background for anomaly detection in video surveillance. They constructed a generative model using U-Net~\cite{RonnebergerFB15}, which avoided the vanishing-gradient problem along with information imbalance, and then they utilized a patch discriminator. In the discriminator, each output scalar of the discriminative network corresponded to a patch of an input image.
Chang et al.~\cite{ChangLSFZ22} adopted attention consistency loss for video anomaly detection.
Gong et al.~\cite{GongWDYXW22} applied dynamic multiple instances of learning loss to enlarge the inter-class distance between anomalous and normal instances.

\subsubsection{Self-Attention}
As a crowd size becomes bigger, traditional detection-based approaches perform poorly in occluded environments.
To address this limitation, Das et al.~\cite{DasEtAl2020} developed an attention-based
deep learning framework by adopting multi-column CNN architecture~\cite{ZhangZCGM16}.
First, they generated basic-detection-based sitting and standing density maps
to capture local information. Thereafter, they created a crowd-counting-based density map as a global counting feature.
Finally, they finished a cross-branch segregating refinement phase that separated the crowd density map into final sitting and standing density maps using an attention mechanism~\cite{HossainHCW19}.
Wang et al.~\cite{WangYY21} proposed a video anomaly detection algorithm based on future frame prediction using GAN and a self-attention mechanism~\cite{OktayEtAl2018corr}. A U-Net model was modified and added with an attention module for the generation model. A Markov GAN discrimination model with a self-attention mechanism was used as a discrimination model, which can affect the generator and improve the generation quality of the future video frame. The main difference between their model and the classical U-Net~\cite{RonnebergerFB15} is that in each layer they maintained the unchanged size of the feature map of two adjacent convolution operations.
Normally, direct implementations of the self-attention sliding-window strategy are costly. However, cyclic shifting is cheaper, although the system becomes more sophisticated~\cite{Goled2022analyticsindiamag}.

\subsection{Hybrid Networks}
A hybrid network is a combination of two or more networks that perform a task more effectively than when they work separately.

\subsubsection{GAN-AE-3DCNN}
As the GAN discriminator extracts the features to classify real and fake data, it cannot extract the features of the data successfully~\cite{KimBC18}. To this end, Shin et al.~\cite{ShinBC20} combined GAN~\cite{ShinC18}, AE~\cite{Vincent11}, and 3DCNN for robust anomaly detection in video surveillance. They converted the GAN generator~\cite{ShinC18} into an AE model based on 3DCNN and used the encoder of the AE as the base model.

\subsubsection{U-Net-AE-GAN}
Li et al.~\cite{LiHDZCS22} integrated the strengths of the U-Net~\cite{KohlRMFLMERR18}, conditional variational AE~\cite{SohnLY15}, and Wasserstein-GAN~\cite{ArjovskyCB17} models to make a probabilistic latent variable model for detecting variational abnormal behaviors.
Dong et al.~\cite{DongZN20} proposed a semi-supervised approach with a dual-discriminator-based GAN structure for video anomaly detection. They applied a generator implemented on U-Net~\cite{RonnebergerFB15} for predicting future frames, one discriminator for discriminating frames (appearance), and another discriminator for differentiating the motion between true and false. They implemented the discriminators with PatchGAN~\cite{IsolaZZE17}.

\subsubsection{ConvLSTM-GAN-AE}
Ji et al.~\cite{JiLZ20} considered both motion and appearance for video anomaly detection. In the motion generative branch, the corresponding optical flow map was generated by a ConvLSTM-based~\cite{ShiCWYWW15} CGAN~\cite{IsolaZZE17} from consecutive frames to learn normal motion patterns. To learn appearance patterns, consecutive frames are reconstructed by an AE~\cite{ShiXFZP21} in the reconstruction branch. In a different direction, Wang et al.  \cite{WangTZZS22} employed both ConvLSTM and VAE  for video anomaly detection.

\subsection{Sundry Networks}
Sundry networks consist of a variety of models (e.g., GCN, RPN, TNN, Q-CNN, HTM, MLAD, and MatchNet), which work separately to solve the crowd anomaly detection problem.

\subsubsection{GCN}
Although it is difficult to perform CNN on graphs due to the arbitrary sizes of the graphs, CNN can be generalized to GCN. A GCN consists of several convolutional and pooling layers for feature extraction, followed by the final fully-connected layers. Convolutions on graphs are defined through the graph Fourier transform.
The key aim of GCN is to take the weighted average of the features of all neighboring nodes, including the node itself, with lower-degree nodes for larger weights, and then to feed the resulting feature vectors through a neural network for training.
Several deep learning models~\cite{SunJHW20,DengxiongBK21,LuoLG21} have been successfully developed to extract and predict information on data graphs for detecting video anomalies.  
Sun et al.~\cite{SunJHW20} built a GCN-based~\cite{KipfW16} deep Gaussian mixture model~\cite{ZongSMCLCC18} for normal and abnormal scene clustering in videos.
Dengxiong et al.~\cite{DengxiongBK21} suggested a GCN-based method to localize abnormal videos.
Luo et al.~\cite{LuoLG21} applied a ResNet~\cite{HeZRS16} mechanism on spatiotemporal GCNs for video anomaly detection.
Cao et al.  \cite{CaoEtAlcorr2022} proposed a weakly supervised adaptive GCN to model the contextual relationships among video segments.
As the adjacency matrix of GCN is prior and hand-crafted, the baseline with GCN can achieve promising results~\cite{0030ZLW0LDL21}.

\subsubsection{RPN}
An RPN is a fully convolutional network that concurrently forecasts object bounds and objectness scores at each position. The RPN is trained end-to-end to create high-quality region proposals. Ren et al.~\cite{RenHG017} merged RPN and Fast Regional-CNN~\cite{Girshick15} into a single network by sharing their convolutional features for real-time object detection.
To analyze abnormal behavior, Hu et al.~\cite{HuDHYZCYZ20} detected and localized all objects (e.g., pedestrians and vehicles) from the crowded scene by using the RPN-based model created by Ren et al.~\cite{RenHG017}.

\subsubsection{SCN}
Sparse-coding-based anomaly detection aims to learn a dictionary to encode all normal events with small reconstruction errors~\cite{ZhaoLX11,LuSJ13}. Although it does not consider the temporal coherence between neighboring frames within normal or abnormal events, an abnormal event is associated with a large reconstruction error.
Wu et al.~\cite{WuLLSS20} proposed a two-stream neural network to extract spatiotemporal fusion features
in hidden layers. With these features, they employed a fast sparse coding network to build a normal dictionary for anomaly detection.

\subsubsection{HTM}
The HTM is a bio-inspired machine intelligence introduced by Hawkins et al.~\cite{HawkinsBlakeslee2005}. It aimed to lay out a theoretical framework for understanding the neocortex by capturing its structural and algorithmic properties.
It makes a prediction by matching new inputs with previously learned patterns to differentiate between normal and abnormal patterns~\cite{HawkinsEtAl2010}. A typical HTM network is a tree-shaped hierarchy of levels. Currently, HTM is predominately applied for anomaly detection in streaming data.
For example, Bamaqa et al.~\cite{BamaqaSBB20} introduced an HTM-based anomaly detection framework for detecting anomalies in crowds to avoid any suspicious behaviors (e.g., overcrowding or any potential accidents).

\subsubsection{MLAD}
Vu et al.~\cite{VuNLLP19} introduced MLAD using multi-level representations of both intensity and motion data.
Typical, MLAD is composed of two DAEs~\cite{VincentLBM08} for extracting the high representations of low-level data (e.g., pixel intensity and optical flow features) and two CGANs~\cite{IsolaZZE17} for detecting anomaly objects at each representation level. MLAD's training and testing codes~\cite{VuNLLP19} are publicly available~\cite{VuNLLP19}. It can produce pixel-wise detection, while many existing models only provide frame-wise detection. Wu et al.~\cite{WuLCH21} chose MLAD~\cite{VuNLLP19} as their baseline model for video anomaly detection. However, MLAD-based methods are not sufficient to analyze motion in videos, which may also create optical flow maps of abnormal events~\cite{JiLZ20}. 

\subsubsection{MatchNet}
MatchNet~\cite{HanLJSB15} is a unified model that combines learning feature representations and comparison functions.
It consists of a deep convolutional network that extracts features from patches and a network of three fully connected layers that computes the similarity between the extracted features. Considering it similar in the spirit of MatchNet~\cite{HanLJSB15}, Ramachandra et al.~\cite{RamachandraJV21} used a source set of labeled video anomaly detection datasets to generate similar and dissimilar video patch pairs. The labeled datasets, used to generate training examples, should be disjointed from the target video anomaly detection dataset on which testing would eventually be performed.

\subsubsection{TNN}
Vaswani et al.~\cite{VaswaniSPUJGKP17} proposed TNN to resolve sequence-to-sequence tasks while handling long-range dependencies with simplicity. TNNs utilize attention layers as their core building blocks. Since the advent of Transformers in Natural Language Processing (NLP), the computer vision community has been focusing very conscientiously on how to employ Transformers in vision-based applications. The Vision Transformer (ViT) instigated by Dosovitskiy  et al.~\cite{DosovitskiyEtAlCorr2020} was an architecture directly inherited from NLP~\cite{VaswaniSPUJGKP17}. However, it was instead employed for image classification with raw image patches as input \cite{TouvronCDMSJ21}. Recently, the purview of Transformers' applications has heightened in the computer vision domain.

Existing anomaly deletion approaches seldom explicitly consider local consistency at a frame level in addition to the global coherence of temporal dynamics in video sequences. To this end, Feng et al.~\cite{FengSCCNC21}  presented a convolutional transformer to perform future frame prediction based on past frames for video anomaly detection. It consisted of three key components, i.e., a convolutional encoder to capture the spatial information of the input video clips, a temporal self-attention module to encode the temporal dynamics, and a convolutional decoder to integrate spatiotemporal features and predict the future frame.
Yuan et al.~\cite{YuanCZWC21} modified the ViT~\cite{DosovitskiyEtAlCorr2020} to make it capable of video prediction. Then, they combined U-Net~\cite{RonnebergerFB15} and ViT~\cite{DosovitskiyEtAlCorr2020} to capture richer temporal information and more global contexts for video anomaly detection.

\subsubsection{Q-CNN}
Quantum computing is a new computational paradigm in quantum deep learning. The benefits and applications of quantum computing by using artificial intelligence tools and algorithms offer much easier handling, enabling the computing of a huge volume of data.
CNN is challenging to learn efficiently if a given dimension of data or the model becomes too large. Q-CNN extends the key features and structures of existing CNN to quantum systems~\cite{OhCK20}. Quantum computing remains a challenging problem, as it is hard to implement non-linearities with quantum unitaries~\cite{SchuldSP14}. Tang~\cite{Tang2018corr} presented quantum-inspired classical algorithms for principal component analysis and supervised clustering. Recently, Blekos et al.~\cite{BlekosK21} developed a quantum counterpart of 3D-CNN for quantum video classification.
It was based on efficiently quantum calculating the difference between successive video frames and then training a quantum convolutional neural network by replacing the convolution operation with a quantum inner-product estimation~\cite{KerenidisLP20}.

\subsection{Our Observations} 

\subsubsection{\textbf{CNN is the best performing model in 2020-2022}}
\begin{figure*}
\centering
\subfloat[]{\includegraphics[width=0.4915\textwidth]{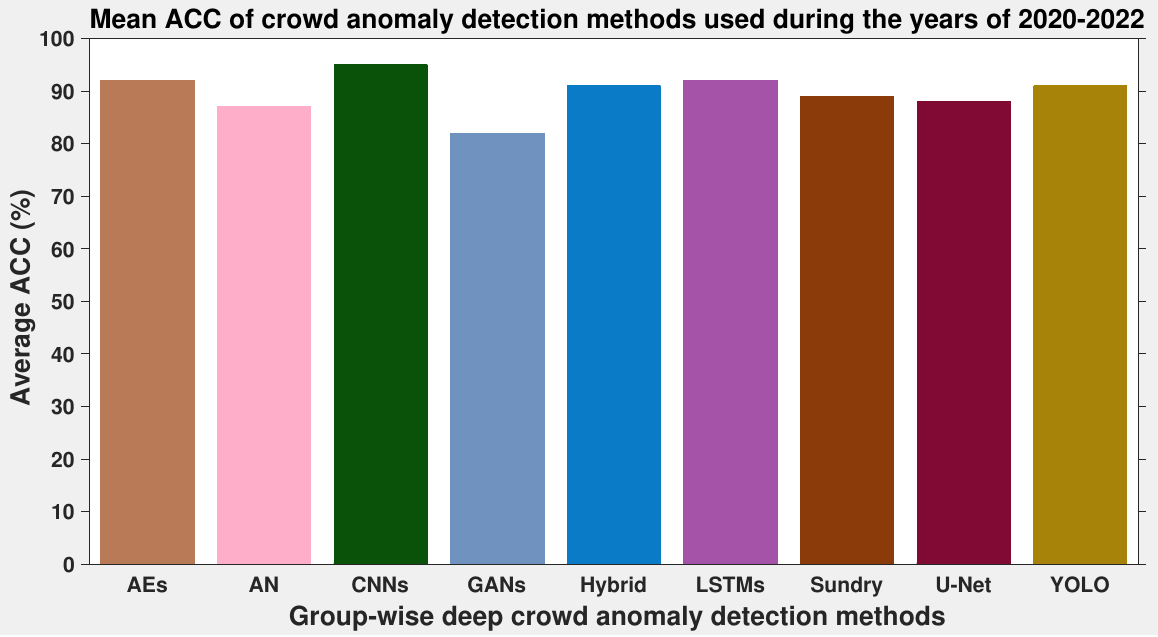}%
\label{GroupPlottingMeanACC}}
\hfil
\subfloat[]{\includegraphics[width=0.4915\textwidth]{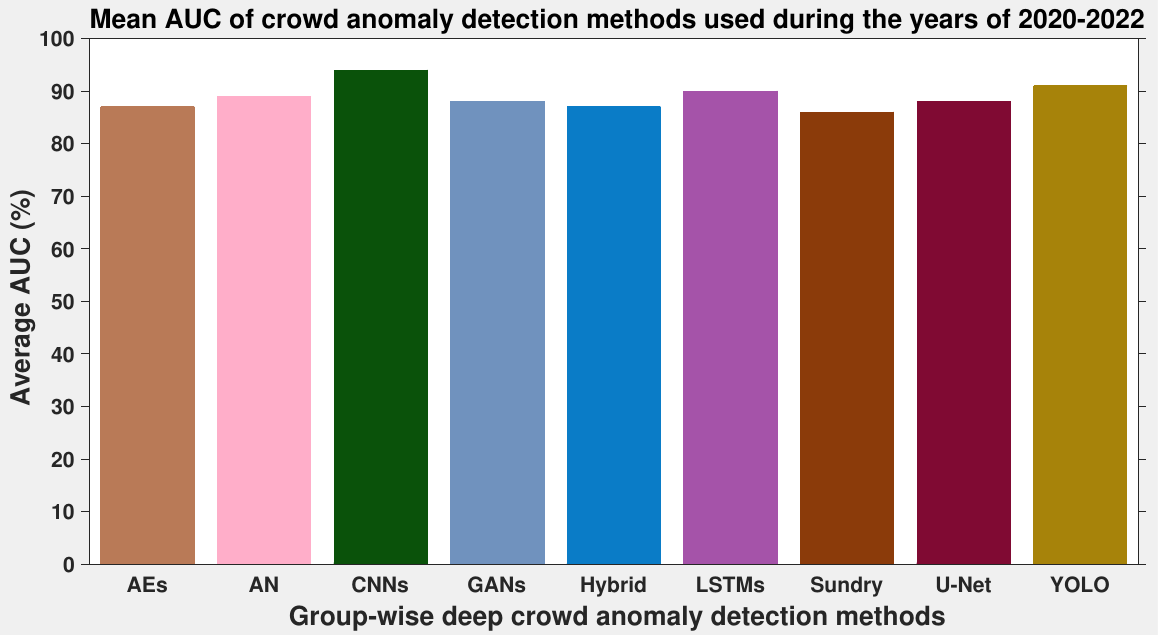}%
\label{GroupPlottingMeanAUC}}
\caption{Group-wise  comparison  of average ACC and AUC considering Figure~\ref{DCADMtaxonomy} and Table~\ref{SummaryOfLiteratureReview2020to2021}. }
\label{GroupWisePlottingsACCAUC}
\end{figure*}

Here, we show which models (listed in Figure~\ref{DCADMtaxonomy}) produced the highest average performance (i.e., ACC and AUC scores) in recent years (e.g., during the years 2020-2022) for crowd anomaly detection.
Figure~\ref{GroupWisePlottingsACCAUC} plots the average ACC and AUC values by considering the taxonomy in Figure~\ref{DCADMtaxonomy} and the values of ACC and AUC from Table~\ref{SummaryOfLiteratureReview2020to2021}. Figure~\ref{GroupPlottingMeanACC} and Figure~\ref{GroupPlottingMeanAUC} show that, on average, the CNN-based models have the highest average ACC and AUC among all models in Figure~\ref{IndivPlottingDCADMtaxonomy}.

CNNs, constructed from layers of artificial neurons, compute the weighted sum of the inputs to offer an output in the form of activation values. Various methods and techniques, including quantization and pruning, have been developed to address the issue of CNN complexity. While training a model considering CNN, the central challenges include overfitting, exploding gradients, and class imbalances. In spite of that, CNNs are well-suited for computer vision and image classification problems. The possible reasons include that:
\begin{itemize}
  \item It is easy to understand and fast to implement;
  \item It accepts data of any dimensionality~\cite{Casalegno2021};
  \item It is good for extracting local and position-invariant features. A well-trained CNN can detect an object from an image even if it is smaller, larger, rotated, or translated from the original image; 
  \item It is a translation equivariant architecture with shared kernel parameters~\cite{GudovskiyEtAlwacv2022};
  \item It automatically detects the important features, thus decreasing the human effort required to develop its functionalities;
  \item It can be used to reduce the number of needed parameters to train it without sacrificing performance;
  \item It accepts pixel values to output various visual features;
  \item As CNN has feature-parameter sharing and dimensionality reductions, the number of parameters is reduced. Consequently, it is computationally efficient;
  \item Convolutional layers of CNN take advantage of the inherent image properties.
\end{itemize}

\subsubsection{\textbf{CNN is the most popular model in 2020-2022}}
\begin{figure*}
\centering
\includegraphics[width=0.99\textwidth]{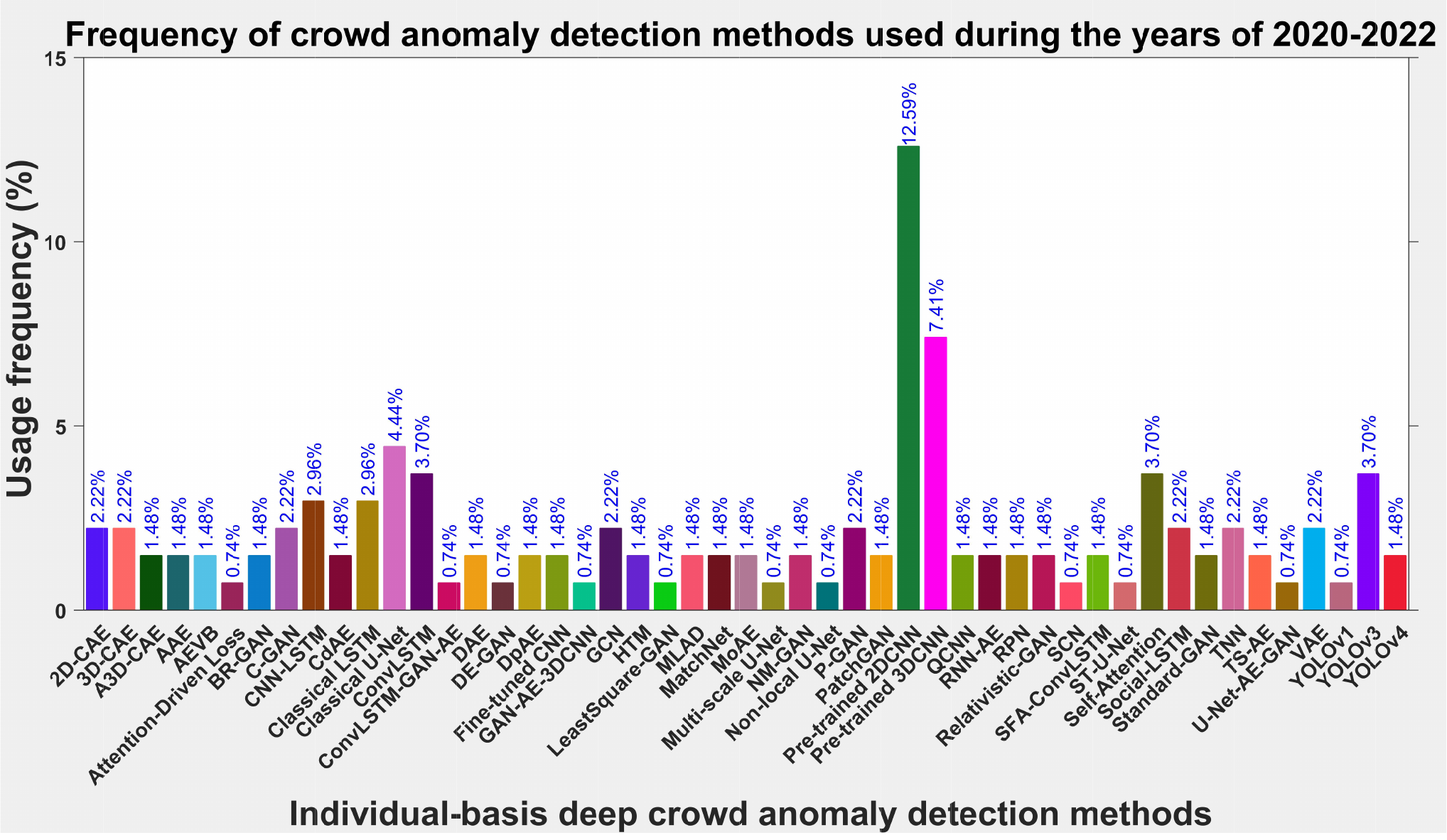}
\caption{Individual basis comparison of usage frequency.}
\label{IndivPlottingDCADMtaxonomy}
\end{figure*}

\begin{figure}
\centering
\includegraphics[width=0.476\textwidth]{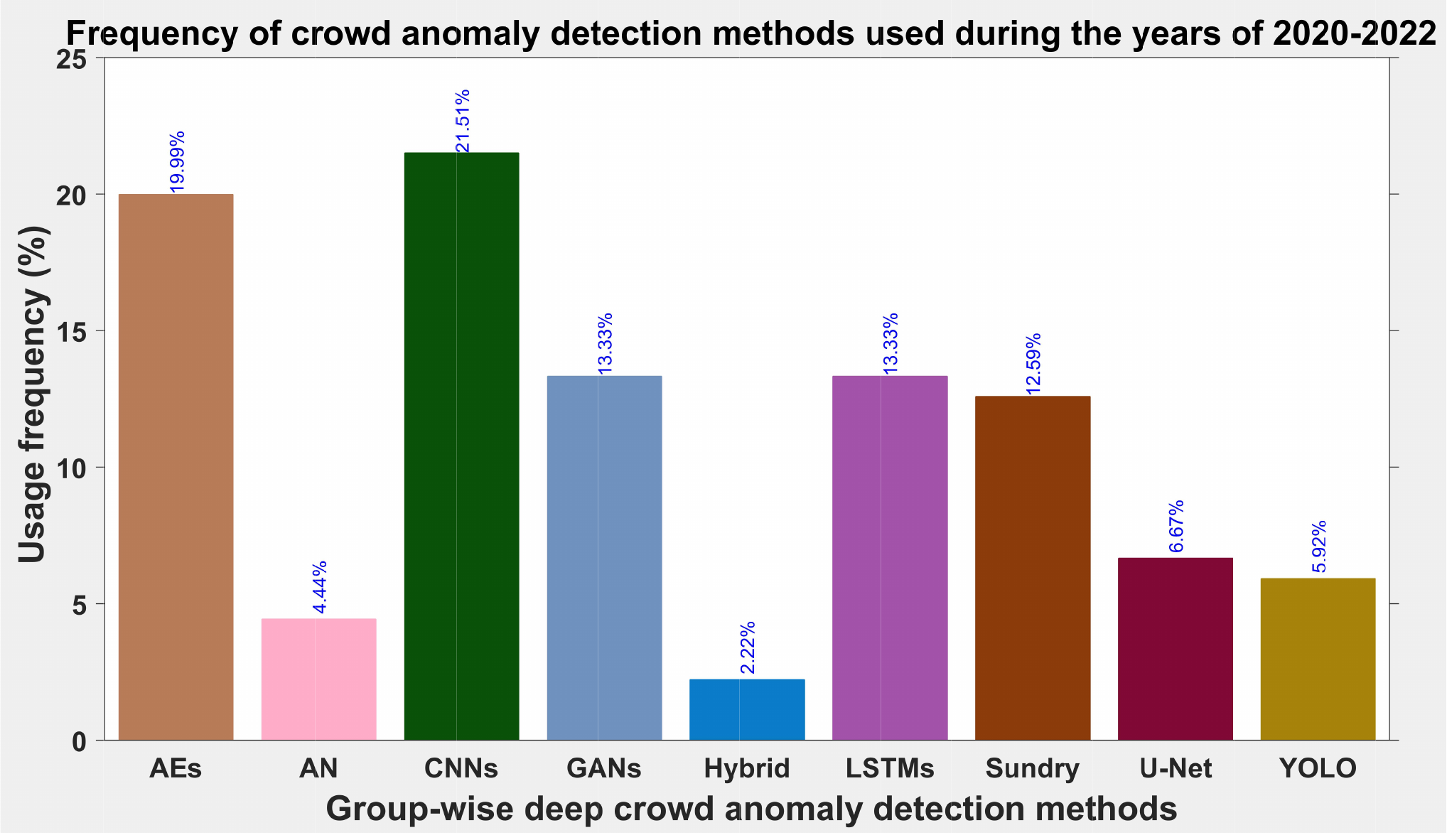}
\caption{Group-wise comparison of usage frequency.}
\label{GroupPlottingDCADMtaxonomy}
\end{figure}

The taxonomies in Figure~\ref{DCADMtaxonomy} and Table~\ref{SummaryOfLiteratureReview2020to2021} show the most frequently used models from 2020 to 2022 for crowd anomaly detection. 
Considering  Equation~(\ref{UsageFrequencyEq01}), Figure~\ref{DCADMtaxonomy}, as well as Table~\ref{SummaryOfLiteratureReview2020to2021};
Figures~\ref{IndivPlottingDCADMtaxonomy} and~\ref{GroupPlottingDCADMtaxonomy} demonstrate the usage frequency of individual and group-wise deep crowd anomaly detection methods, respectively. It is noticeable that the pre-trained 2DCNN-based models have become the most frequently used models. In a group-wise view, CNNs-based models grew as the most frequently used models, and AEs took second place.

CNNs are supervised models, whereas AEs are unsupervised models. AEs can be trained on unlabeled data. If the data are labeled, then the CNN-based model can be employed for better performance.

\subsubsection{\textbf{Can CNNs overcome biased datasets?}}
Deep learning models could be used to accomplish tasks rapidly, however, this does not mean they always do so reasonably.
What would be the consequence if the datasets used to train deep learning models contained biased data?
The deep models would likely exhibit the same bias when they make decisions in practice.
For instance, if a dataset contains mostly images of white men, then a facial-recognition model trained with these data may be less accurate for women or people with different skin tones~\cite{MITnews21Feb2022}.
In 2021, a team of researchers~\cite{MadanEtAlcorr2021} from MIT, Harvard University, and Fujitsu Ltd sought to understand when and how a deep model is capable of overcoming this kind of dataset bias.
Basically, they posed the following question~\cite{MadanEtAlcorr2021}: ``\emph{Can a network shown only the Ford Thunderbird from the front and the Mitsubishi Lancer from side generalize to classify the category and viewpoint for a Thunderbird seen from the side?}". Explicitly, if each Ford Thunderbird in the training dataset is exhibited from the forepart and
if the trained model is given an image of a Ford Thunderbird from the side view, then the detection result may be misclassified, even if the model was trained on hundreds of thousands of car images.
The researchers  \cite{MadanEtAlcorr2021}  used an approach from neuroscience to study how training data affects whether an artificial neural network can learn to recognize objects it has not seen before.
Their results showed that the generalizability of a deep model is influenced by both the diversity of the data and the way the model is trained. Can CNNs perform well on growing training data diversity?
They demonstrated that CNNs generalize better to out-of-distribution category-viewpoint combinations (i.e., combinations not seen during training) as the training data diversity grows \cite{MadanEtAlcorr2021}. 

\subsubsection{\textbf{Can Transformers outperform CNNs?}}
Trockman et al.  \cite{TrockmanEtALcorr2022} scrutinized the ViT~\cite{DosovitskiyEtAlCorr2020}. They noticed that the Transformer-based models (e.g., ViT~\cite{DosovitskiyEtAlCorr2020}) might exceed the performance of CNNs in some settings. Nevertheless, due to the quadratic runtime of the self-attention layers in Transformers, ViTs demand the use of patch embeddings. To this end, they posed the following question \cite{TrockmanEtALcorr2022}: ``\textit{Is the performance of ViTs due to the inherently-more-powerful Transformer architecture, or is it at least partly due to using patches as the input representation?}"
They showed some evidence for the latter. Specifically, they introduced an extremely straightforward model that is similar in spirit to the ViT~\cite{DosovitskiyEtAlCorr2020} and the even-more-basic MLP-Mixer~\cite{TolstikhinEtALcorr2021}, called ConvMixer. For their ConvMixer, they used only standard convolutions to carry out the mixing steps. They claimed that despite the simplicity of their ConvMixer, it outperformed both the ViT~\cite{DosovitskiyEtAlCorr2020} and MLP-Mixer~\cite{TolstikhinEtALcorr2021}. They further stated that ConvMixer was competitive compared with ResNets~\cite{HeZRS16}, DeiTs (Data-efficient image Transformers)~\cite{TouvronCDMSJ21} and ResMLPs (Residual Multi-Layer Perceptrons) ~\cite{TouvronEtAlcorr2021}.

\subsubsection{\textbf{Are CNNs and Transformers complementary technologies?}}
In 2022, a team of researchers~\cite{LiuEtAlcorr2022} from Facebook AI Research (FAIR) and UC Berkeley studied the differences between CNNs and Transformers to discover the confounding variables while comparing their network performances.
As per the team~\cite{LiuEtAlcorr2022}, the objective of their study was to ``\textit{bridge the gap between the pre-ViT and post-ViT eras for ConvNets, as well as to test the limits of what a pure ConvNet can achieve.}"
The team~\cite{LiuEtAlcorr2022} proposed a family of pure CNNs called \textit{ConvNeXt}. The team~\cite{LiuEtAlcorr2022} found that ConvNeXts could compete well with Transformers in terms of accuracy, robustness, and scalability. In addition, the ConvNeXt possesses the efficiency standard of CNNs. Due to the fully-convolutional nature of training and testing, its implementation is easy. As computer vision applications are very diverse, the ConvNeXt may be more suited for some tasks in computer vision, while Transformers may be more flexible for other tasks. Specifically, Transformers may be more flexible when employed for tasks demanding discretized, sparse, or structured outputs~\cite{LiuEtAlcorr2022}.

\subsubsection{\textbf{Are we shifting from CNN to Transformer technologies?}}
Cautiously, Transformers have begun to catch up with CNNs for computer vision tasks.
One of the key reasons why people are choosing hierarchical Transformers over CNNs is due to the CNN's poor scalability, with multi-head attention being the key component~\cite{Goled2022analyticsindiamag}.
Carion et al.~\cite{CarionMSUKZ20} proposed the DEtection TRansformer (DETR), which mainly consists of a set-based global loss that can make unique predictions via bipartite matching as well as a Transformer encoder--decoder architecture.
In spite of Yann LeCun's own innovation of LeNet (i.e., the early version of CNN), he preferred the DETR architecture~\cite{Goled2022analyticsindiamag}. Are we on the verge of shifting from CNN to Transformer technologies? The debate between CNNs and Transformers is ongoing.


\subsubsection{\textbf{Will the quantum computing be the next paradigm?}} 
Quantum computing is one of the fastest-growing technologies. Currently, the computing capacities of traditional computers restrict the computational capabilities of deep learning algorithms. Quantum computing can process vast datasets at much faster speeds and feed data to artificial intelligence technologies, which can look over data at an ultra-fine level to find diverse patterns and anomalies. A qubit is the basic unit of quantum information. It is the complement in quantum computing to the bit (binary digit) of classical computing. In 2021, IBM and QuEra Computing were among the first businesses to build quantum computers with more than 100 qubits~\cite{IBMQuantum2022}. One of the key considerations when applying quantum computers is that we can perform more sophisticated analyses and build deep learning models. Compared with traditional computers, quantum computers use more data more efficiently. Accordingly, researchers can have a better understanding of their working data and models.
In the literature, there is a common consensus that quantum computers will assist in the solving of previously impossible issues, notably in the areas of data science and artificial intelligence. 

\onecolumn

\scriptsize

\captionsetup{width=1.0\textwidth}

\setlength{\tabcolsep}{0.13cm}

\begin{longtable}{
|p{0.367cm}| p{0.75cm}  p{1.05cm}   p{1.75cm}  p{1.11cm} p{0.65cm} p{0.63cm} p{0.35cm} p{0.35cm} p{0.20cm} p{0.57cm} p{0.59cm}  p{3.25cm} p{0.34cm}  p{0.35cm} p{0.32cm} p{0.32cm}|}

\caption{Summary of deep crowd anomaly detection methods mostly published in mainstream journals and conferences between 2020 and 2022. Symbol $\clubsuit$ indicates the code of models is publicly released. $e^{-n}$ belongs to $10^{-n}$.}

\label{SummaryOfLiteratureReview2020to2021}\\\hline

\multirow{2}{*}{Year}&\multirow{2}{*}{Ref} &\multirow{2}{*}{Aim} &\multirow{2}{*}{Technique}  &\multirow{2}{*}{Dataset}
&Sample &Hard- &\multicolumn{3}{c|}{Hyperparameters\,\,\,\,} &\multicolumn{2}{c}{Mean ratio of} &\multirow{2}{*}{Analysis of Experimental Result}  &\multicolumn{4}{c|}{Mean Performance} \\\cline{8-10} \cline{10-12} \cline{14-17}
&&&&&Size&ware&Batch&\,\,Epoch&$\,\,\eta\,$&Training&\,\,Testing&&ACC&AUC&F1S&EER\\\hline

\multirow{77}{*}{\rotatebox{77}{\normalsize\textbf{2020}\scriptsize}}
&Park et al. \cite{ParkNH20}$\clubsuit$
&To detect video anomaly. 
&P-GAN \cite{LiuLLG18}, MAML algorithm \cite{FinnAL17}, Adam \cite{KingmaB14}, PyTorch \cite{Paszke2017automatic}. 
&UCSD Ped2 \cite{ChanLV08}, C.Av. \cite{LuSJ13}, S.T. Camp. \cite{LuoLG17}. 
&309727 frames 
&Nvidia GTX TITAN Xp  
&4  
&10  60 
&2$e^{\emph{-5}}$   2$e^{\emph{-4}}$   
&\,\,90\%  
&\,\,10\%  
&Frame-level AUC of 0.97,  0.885, and 0.705 obtained from UCSD Ped2 \cite{ChanLV08},
C.Ave. \cite{LuSJ13}, and S.T. Cam. \cite{LuoLG17}, respectively. AOS and PLR were considered.   
&NA  
&0.853  
&NA  
&NA 
\\
\hline

\end{longtable}

\normalsize

\twocolumn

\section{Architectural Impacts of 2DCNN Models}\label{ArchImpactsOf2DCNNOnCrowdAnomalyDetection}
Because 2DCNN-based models were the most popular anomaly detection approaches from 2020 to 2022 (see Figure~\ref{IndivPlottingDCADMtaxonomy}), we studied their architectural impacts on various crowd datasets for anomaly detection. The architectural influence of pre-trained CNN models on video anomaly detection is not a new research area \cite{PangYSH020,Al-DhamariSM20,GutoskiRHALL21}. For example, Pang et al. \cite{PangYSH020}  employed ResNet50 \cite{HeZRS16} to exclusively examine appearance-based anomalies . Without counting RTM, they compared the performance of ResNet50 \cite{HeZRS16}, VGGNet \cite{SimonyanZ14aVGG16}, and 3DCNN \cite{TranBFTP15} for Ped1-Ped2 in the UCSD \cite{ChanLV08}, Entrance-Exit of Subway \cite{AdamRSR08}, and UMN \cite{UMNdataset2021} datasets.
In a different vein, using only UCSD Ped1 \cite{ChanLV08} and UMN \cite{UMNdataset2021}  datasets, and without counting RTM,  Al-Dhamari et al.  \cite{Al-DhamariSM20} argued that VGGNet19 \cite{SimonyanZ14aVGG16} had the highest detection ACC among the GoogleNet \cite{SzegedyLJSRAEVR15}, ResNet50 \cite{HeZRS16}, AlexNet \cite{KrizhevskySH12}, and VGGNet16  \cite{SimonyanZ14aVGG16} models. Gutoski et al. \cite{GutoskiRHALL21} supported this position by taking 12 pre-trained CNN models on ImageNet~\cite{DengDSLL009} as feature extractors and then employing the obtained features to seven video anomaly detection benchmark datasets. Without examining RTM, they performed a simple statistical analysis of their results. Nevertheless, they reached a promising conclusion that the architectural differences are negligible when choosing a pre-trained model to detect video anomalies.

In this section, we performed a similar experiment to Gutoski et al. \cite{GutoskiRHALL21} by additionally considering  the RTM of six pre-trained 2DCNN models (see Table \ref{SummaryOfLiteratureReview2020to2021}) and rigorous statistical analyses. 
We have taken RTM into account because it's software performance depends on the features of the computing environment, including RTM \cite{SCHMIDT2013159}. 

\subsection{Essential Techniques}
The pre-trained 2DCNN models in Table \ref{SummaryOfLiteratureReview2020to2021} cannot provide necessarily sufficient performance scores. As of today,  numerous pre-trained 2DCNN models exist, which can be used to transfer their learning from the ImageNet dataset to other models
. In CNN, each layer has two types of parameters, namely weights and biases. Table \ref{2DCNNparameters} represents the essential information of our six used pre-trained 2DCNN models \cite{KerasApps2021}.
In a deep network model, parameters (e.g., batch size, learning rates, etc.) and functions (e.g., activation, optimization, and loss functions) should be appropriately chosen \cite{WuLCH21}. Consequently, we compared the performance of our used pre-trained 2DCNNs, employed them as feature extractors, and then trained  OCSVM \cite{ScholkopfWSSP99} over the extracted features to learn the normal patterns. SVMs learned the smallest region of the feature space as normal, and during testing, new samples located outside the region were classified as anomalies. We took the videos, extracted their frames, and categorized the frames into two folders for training and testing. The training limits only positive samples (i.e., normal images), while the testing contains both abnormal and random normal samples.

\begin{table}
\begin{center}
\caption{Brief of six pre-trained models.}
\begin{tabular}{|c|c|c|c|c|c|c|c|c|c|c|c|c|c|c|c|c|c|c|c|c|c|}\hline
Model&Size &Parameters&Depth \\\hline
DenseNet121 \cite{HuangLMW17}          &033MB  &008062504    &121  \\
VGGNet16 \cite{SimonyanZ14aVGG16}         &528MB  &138357544    &023  \\
VGGNet19 \cite{SimonyanZ14aVGG16}         &549MB  &143667240    &026  \\
InceptionV3 \cite{SzegedyLJSRAEVR15}   &092MB  &023851784    &159  \\
MobileNet  \cite{HowardZCKWWAA17corr}  &016MB  &004253864    &088  \\
ResNet50 \cite{HeZRS16}                &099MB  &025636712    &168 \\\hline
\end{tabular}
\label{2DCNNparameters}
\end{center}
\end{table}

\subsection{Hardware and Software Specifications}
The hardware specification involved a Tesla P100 16GB VRAM as GPU along with 13GB RAM + 2-Core of 2GHz Intel Xeon as CPU.
The software specification encompassed the Kaggle platform \cite{Kaggle2021} with Python 3.9 and machine learning packages, including Pandas \cite{Beazley12e}, Numpy \cite{WaltCV11}, and  Scikit-Learn \cite{PedregosaVGMTGBPWDVPCBPD11}.  Pandas \cite{Beazley12e}, Numpy \cite{WaltCV11}, and  Scikit-Learn \cite{PedregosaVGMTGBPWDVPCBPD11} work cooperatively. For example, Pandas \cite{Beazley12e}  can help to load, clean, or manipulate a data frame. The Numpy \cite{WaltCV11} array can allow the translated Pandas data frame. Scikit-Learn \cite{PedregosaVGMTGBPWDVPCBPD11} functions can return the Numpy array.

\subsection{Experimental Setup}
We have considered UCSD Ped1  \cite{ChanLV08}, UCSD Ped2  \cite{ChanLV08}, UMN  \cite{UMNdataset2021}, CUHK-Avenue \cite{LuSJ13}, ShanghaiTech Campus \cite{LuoLG17}, and UCF-Crime \cite{SultaniCS18} \cite{SultaniCS18} datasets.
A dataset such as UMN~\cite{UMNdataset2021} contains slightly more than 7000 images. Furthermore,  all images have to be converted to a Numpy array, which currently used memory configurations cannot accommodate.
As a result, we placed 4000 and 2550 sample images in the training and testing folders, respectively.
We used a total of 6550 sample images from each dataset, which is a fair distribution of the datasets required for the statistical test.
All 4000 images in the train folder were normal, whereas the 2550 images in the testing folder consisted of both anomalous and randomly chosen images. The classification was based on learning from the $\mathcal{X}$ class and predicting anything that was negative in the $\mathcal{X}$ class
.
In our pre-processing, we fetched the images and re-sized each of them into a (224,224,3) array and augmented them by re-scaling, then we converted them all into a Numpy array from a list. The array of images was divided into training and testing sets. We employed these data to then extract the features from the image arrays using a pre-trained 2DCNN.

For DenseNet121~\cite{HuangLMW17}, as an example, we achieved a (1,7,7,1024) feature image into a list array, which we converted into a Numpy array for efficiency. Furthermore, the format was [samples, rows, columns, channels]. At the end of the deep feature extraction for both the training and testing data, we obtained (4000,7,7,1024) and (4000,7,7,1024) feature matrices, respectively.
We divided the data into \textit{x\_train} and \textit{x\_outliers} because it was one classification
. Henceforth, the \textit{x\_train} contained only positive, whereas the \textit{x\_outliers} belonged to negatives that were randomly selected from the test folder.
We flattened the data from (2000,7,7,1024) into (2000,50176) for the 2D input of our OCSVM classifier. We applied a standard scaler to the data for some statistical scaling and used principal component analysis to reduce the feature space dimensionality of the data. We considered the parameters of OCSVM as (gamma=`scale', kernel=`rbf', nu=0.01).
The predictions returned by OCSVM were of the form \{-1, 1\}, where -1 and 1  were referred to as \textit{anomaly}  and  \textit{normal} cases, respectively.

\subsection{Experimental Results}
\begin{figure*}
\centering
\subfloat[]{\includegraphics[width=0.46\textwidth]{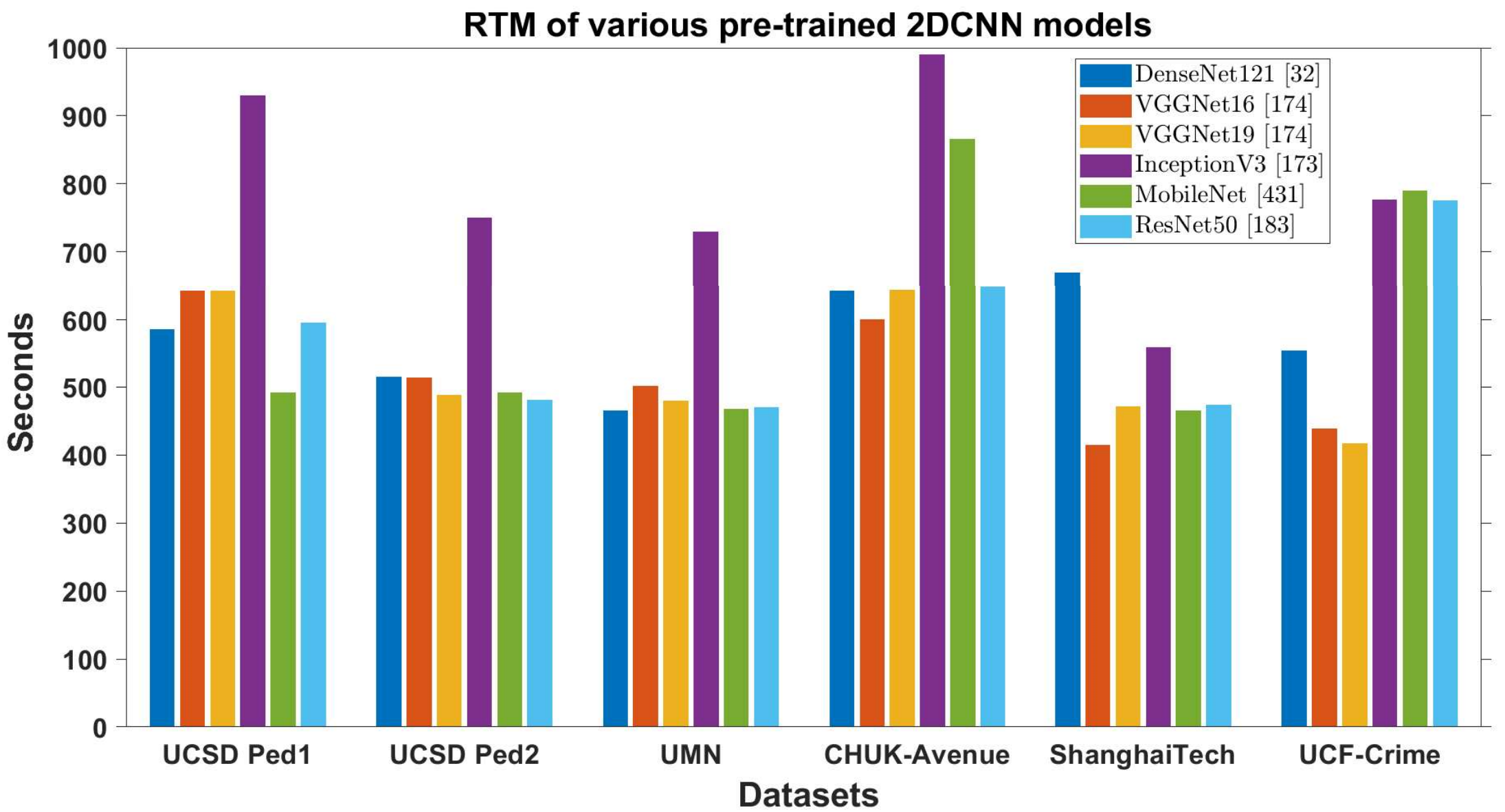}%
\label{GroupPlottingRTM}}\hfil
\subfloat[]{\includegraphics[width=0.46\textwidth]{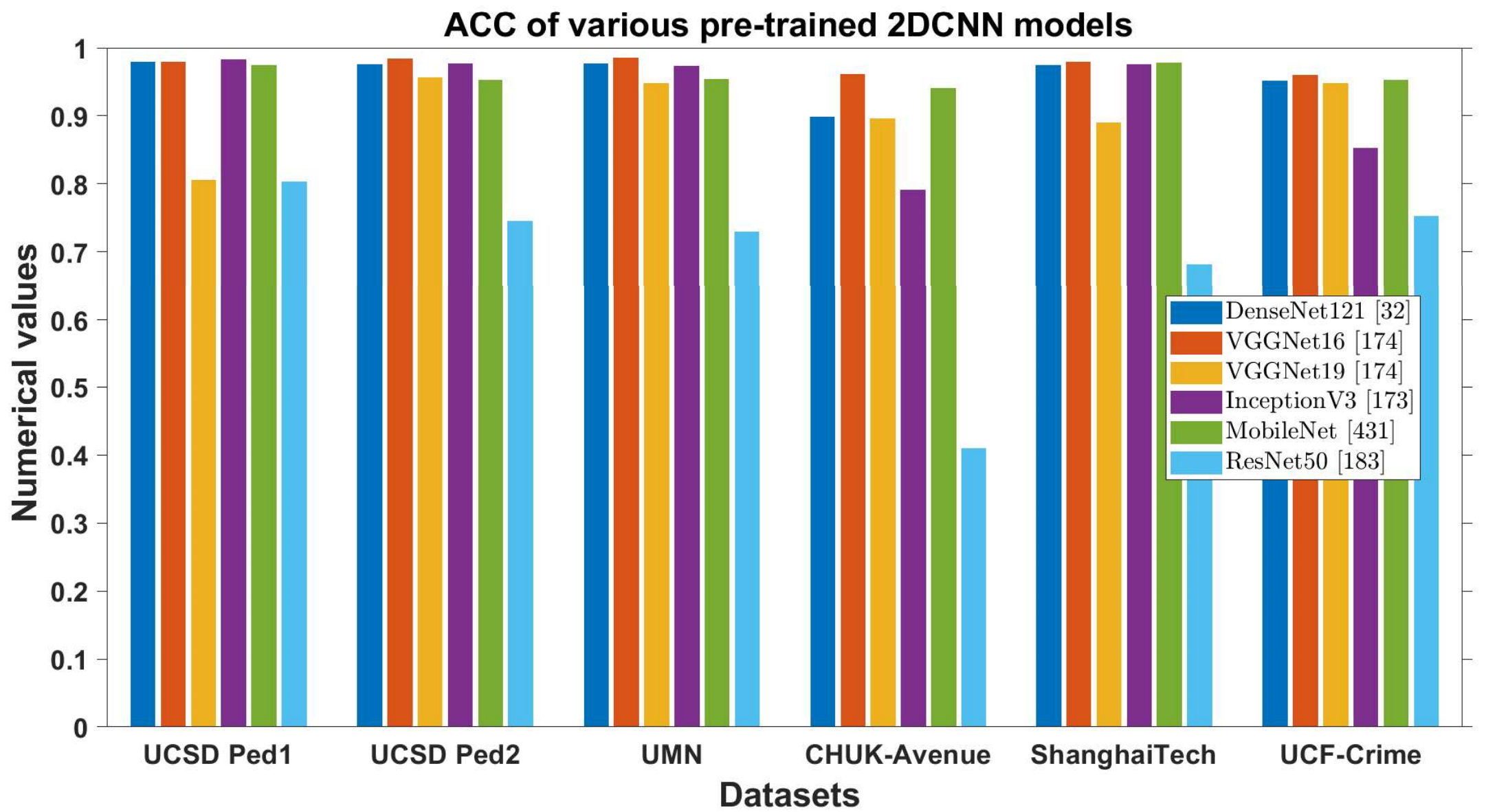}%
\label{GroupPlottingACC}} \hfil
\subfloat[]{\includegraphics[width=0.4585\textwidth]{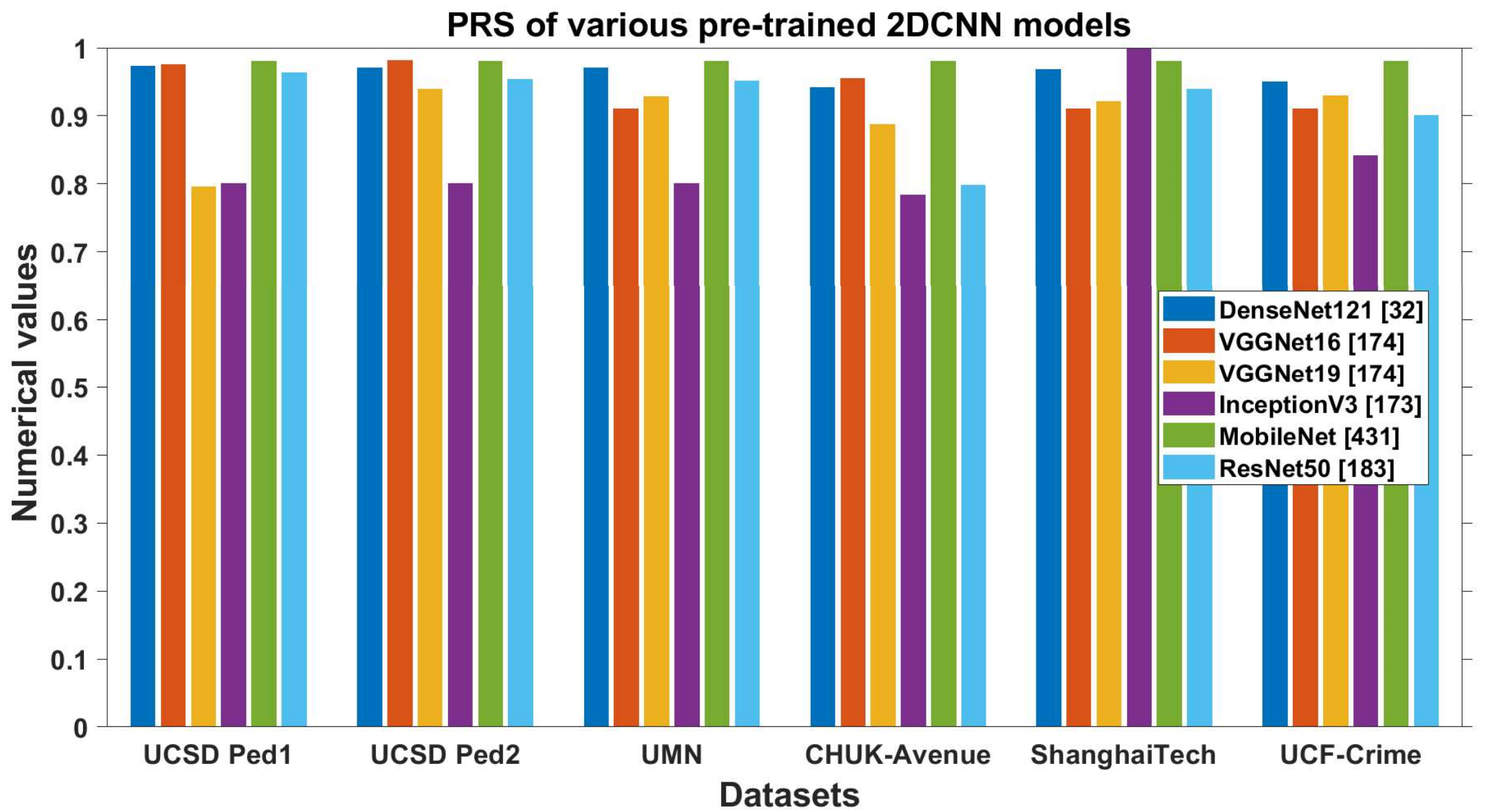}%
\label{GroupPlottingPRS}}\hfil
\subfloat[]{\includegraphics[width=0.46\textwidth]{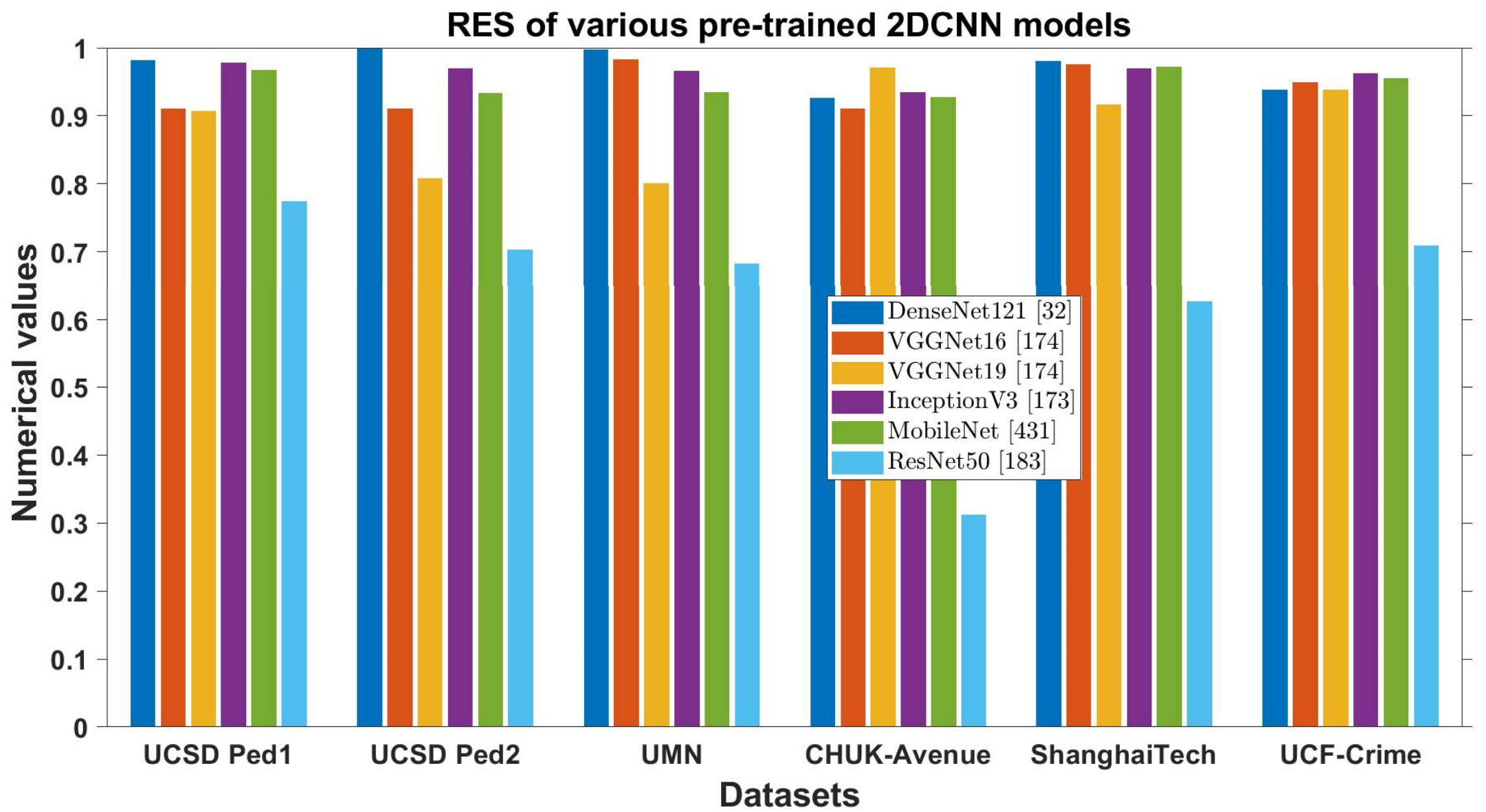}%
\label{GroupPlottingRES}} \hfil
\subfloat[]{\includegraphics[width=0.46\textwidth]{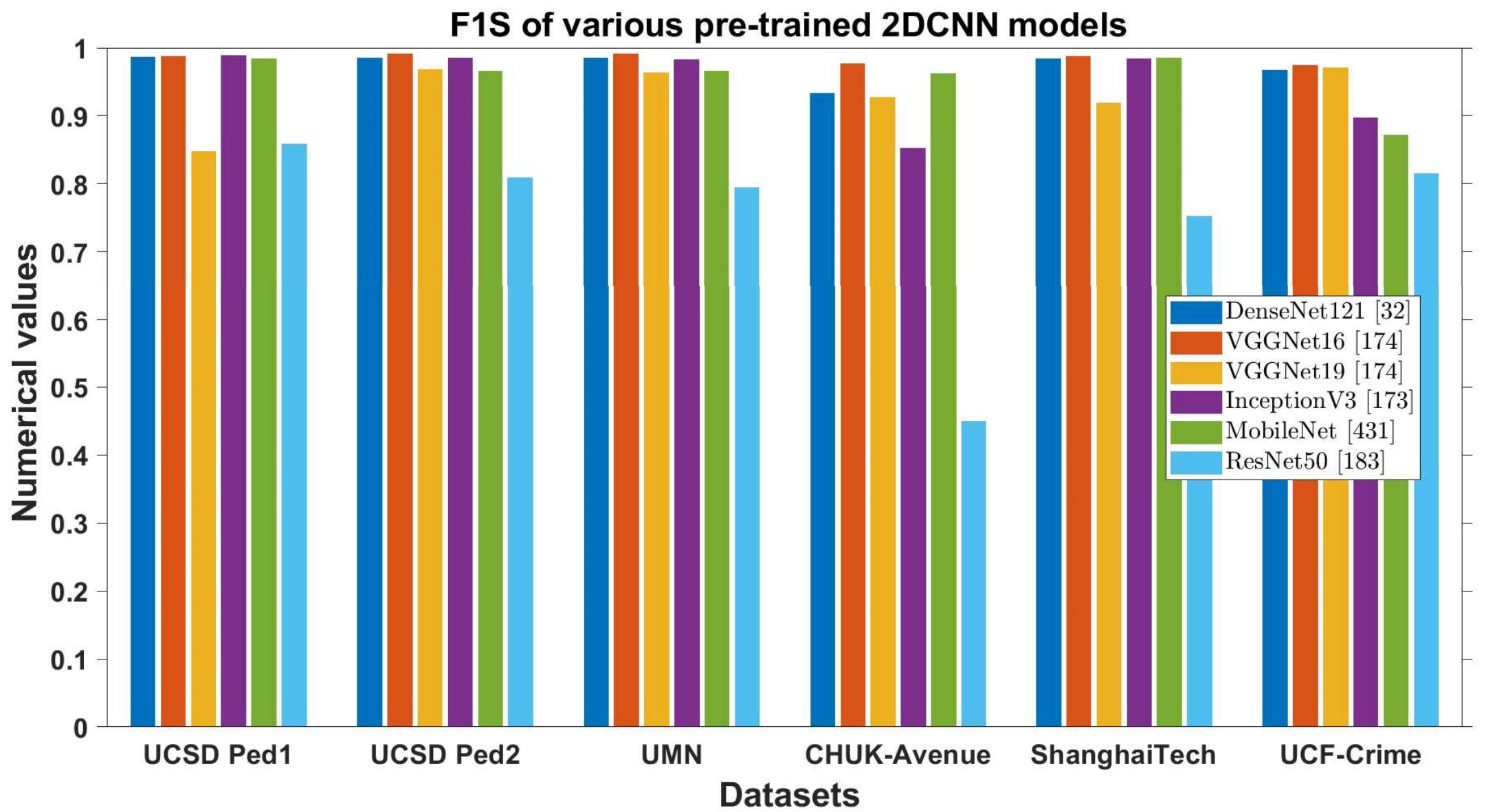}%
\label{GroupPlottingF1S}}\hfil
\subfloat[]{\includegraphics[width=0.46\textwidth]{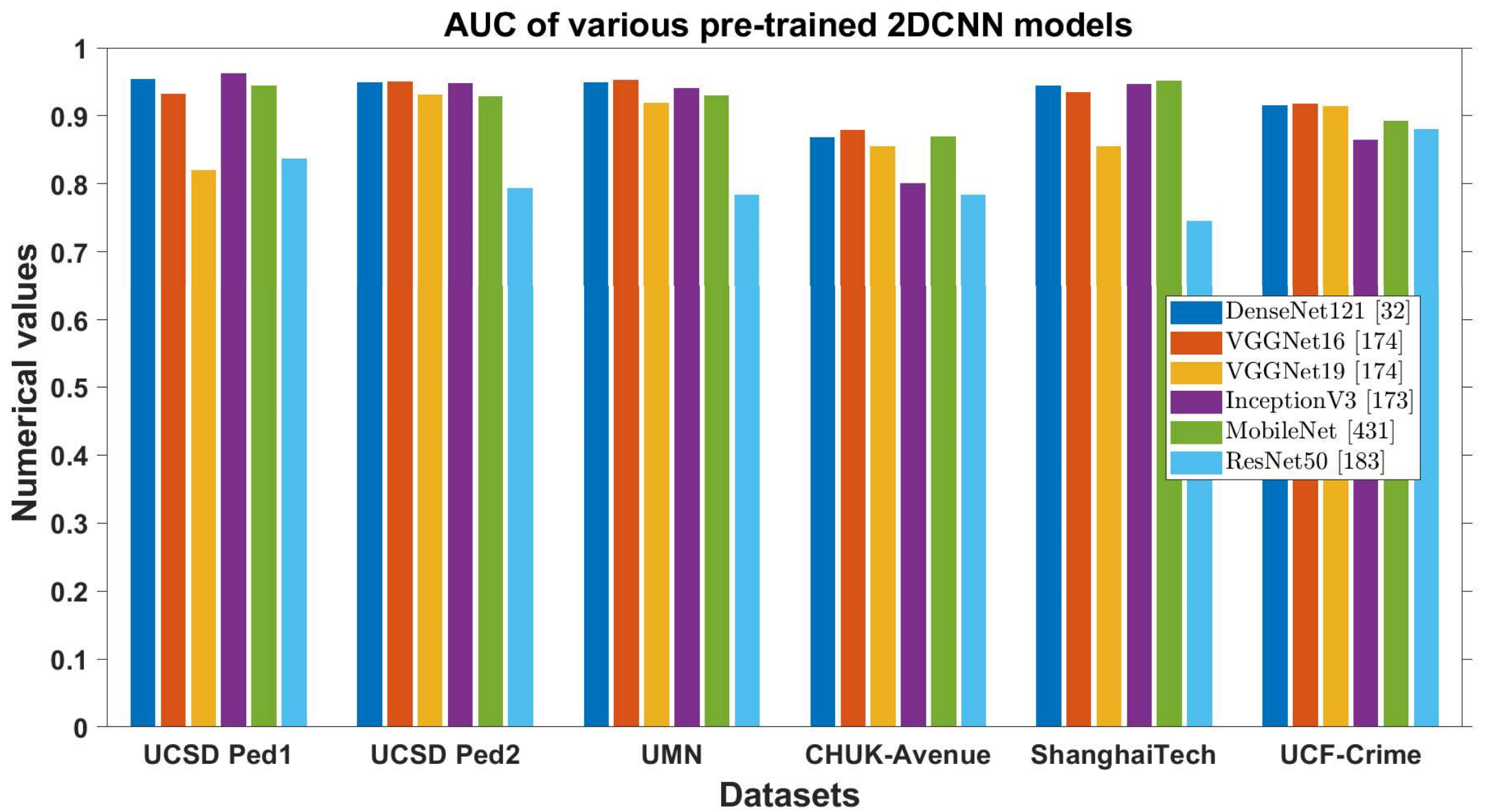}%
\label{GroupPlottingAUC}}\hfil
\caption{Plotting of RTM, ACC, PRS, RES, F1S, and AUC using data from Table \ref{ExperimentalResults2DCNN}.}
\label{plottingExperimentalResults2DCNN}
\end{figure*}

Table~\ref{ExperimentalResults2DCNN} records the experimental performance scores of ACC, PRS, RES, F1S, AUC, and TRM along with their ineffectiveness scores. 
 Figs. \ref{GroupPlottingRTM}, \ref{GroupPlottingACC}, \ref{GroupPlottingPRS}, \ref{GroupPlottingRES}, \ref{GroupPlottingF1S}, and \ref{GroupPlottingAUC}  visualize the scores of RTM, ACC, PRS, RES, F1S, and AUC, respectively, using data in Table \ref{ExperimentalResults2DCNN}.
From Figure \ref{plottingExperimentalResults2DCNN}, it is easy to observe that the effectiveness scores vary based on the underlying datasets and models. As a result, it is extremely hard to determine which model is superior to its alternative. Even so, non-parametric statistical tests can provide an agreeable solution.

\begin{table*}
\begin{center}
\caption{Experimental results and analysis.}
\begin{tabular}{|c|c|c|c|c|c|c|c|c|c|c|c|c|c|c|c|c|c|c|c|c|c|}\hline
\multirow{2}{*}{Datasets}&\multirow{2}{*}{Model} &\multicolumn{5}{c|}{Scores of Effectiveness} &\multicolumn{5}{c|}{Scores of Ineffectualness}&Runtime \\\cline{3-12}
&&ACC &PRS  &RES &F1S &AUC &1-ACC &1-PRS  &1-RES &1-F1S &1-AUC &Seconds \\\hline

\multirow{6}{*}{\rotatebox{90}{\textbf{U. Ped1 \cite{ChanLV08}}}}
&DenseNet121 \cite{HuangLMW17}            &0.9785 &0.9728 &0.9809 &0.9862  &0.9533  &0.0215    &0.0272    &0.0191   &0.0138   &0.0467       &586 \\
&VGGNet16 \cite{SimonyanZ14aVGG16}                 &0.9756 &0.9701 &0.9989 &0.9844  &0.9482  &0.0244    &0.0299    &0.0011    &0.0156    &0.0518       &516 \\
&VGGNet19 \cite{SimonyanZ14aVGG16}                 &0.9762 &0.9702 &0.9987 &0.9847  &0.9486  &0.0238    &0.0298    &0.0013    &0.0153    &0.0514       &465 \\
&InceptionV3 \cite{SzegedyLJSRAEVR15}         &0.8983 &0.9419 &0.9245 &0.9333  &0.8674  &0.1017    &0.0581    &0.0755    &0.0667    &0.1326       &642 \\
&MobileNet  \cite{HowardZCKWWAA17corr}            &0.9742 &0.9676 &0.9802 &0.9835  &0.9442  &0.0258    &0.0324    &0.0198    &0.0165    &0.0558      &669 \\
&ResNet50 \cite{HeZRS16}                &0.9516 &0.9500 &0.9367 &0.9672  &0.9153  &0.0484    &0.0500    &0.0633    &0.0328    &0.0847       &554 \\\hline

\multirow{6}{*}{\rotatebox{90}{\textbf{U. Ped2 \cite{ChanLV08}}}}
&DenseNet121 \cite{HuangLMW17}           &0.9783 &0.9749 &0.9101  &0.9872  &0.9320  &0.0217    &0.0251    &0.0899    &0.0128    &0.0680      &642 \\
&VGGNet16 \cite{SimonyanZ14aVGG16}                &0.9841 &0.9814 &0.9101  &0.9906  &0.9503  &0.0159    &0.0186    &0.0899    &0.0094    &0.0497      &514 \\
&VGGNet19 \cite{SimonyanZ14aVGG16}                &0.9847 &0.9100 &0.9822  &0.9910  &0.9523  &0.0153     &0.0900     &0.0178     &0.0090    &0.0477     &501 \\
&InceptionV3 \cite{SzegedyLJSRAEVR15}        &0.9603 &0.9549 &0.9100  &0.9769  &0.8788  &0.0397    &0.0451    &0.0900    &0.0231    &0.1212      &600 \\
&MobileNet  \cite{HowardZCKWWAA17corr}           &0.9791 &0.9100 &0.9757  &0.9877  &0.9346  &0.0209    &0.0900    &0.0243    &0.0123    &0.0654      &414 \\
&ResNet50 \cite{HeZRS16}               &0.9590 &0.9100 &0.9483  &0.9735  &0.9172  &0.0410    &0.0900    &0.0517    &0.0265    &0.0828     &439 \\\hline

\multirow{6}{*}{\rotatebox{90}{\textbf{UMN  \cite{UMNdataset2021}}}}
&DenseNet121 \cite{HuangLMW17}           &0.8053 &0.7953 &0.9068 &0.8474  &0.8200  &0.1947    &0.2047    &0.0932    &0.1526    &0.1800       &563 \\
&VGGNet16 \cite{SimonyanZ14aVGG16}                &0.9557 &0.9387 &0.8080 &0.9684  &0.9308  &0.0443    &0.0613    &0.1920    &0.0316    &0.0692      &488 \\
&VGGNet19 \cite{SimonyanZ14aVGG16}                &0.9477 &0.9285 &0.8000 &0.9629  &0.9183  &0.0523    &0.0715    &0.2000    &0.0371    &0.0817       &479 \\
&InceptionV3 \cite{SzegedyLJSRAEVR15}        &0.8961 &0.8875 &0.9701 &0.9270  &0.8547  &0.1039    &0.1125    &0.0299    &0.0730    &0.1453       &644 \\
&MobileNet  \cite{HowardZCKWWAA17corr}           &0.8895 &0.9206 &0.9163 &0.9185  &0.8744  &0.1105    &0.0794    &0.0837    &0.0815    &0.1256     &471 \\
&ResNet50 \cite{HeZRS16}               &0.9470 &0.9300 &0.9379 &0.9701  &0.9142  &0.0530    &0.0700    &0.0621    &0.0299    &0.0858       &417 \\\hline

\multirow{6}{*}{\rotatebox{90}{\textbf{C.Avenue \cite{LuSJ13}}}}
&DenseNet121 \cite{HuangLMW17}           &0.9827 &0.8000 &0.9780 &0.9889  &0.9625  &0.0173    &0.2000    &0.0220    &0.0111    &0.0375       &930 \\
&VGGNet16 \cite{SimonyanZ14aVGG16}                &0.9760 &0.8000 &0.9697 &0.9846  &0.9479  &0.0240    &0.2000    &0.0303    &0.0154    &0.0521      &750 \\
&VGGNet19 \cite{SimonyanZ14aVGG16}                &0.9727 &0.8000 &0.9657 &0.9826  &0.9408  &0.0273    &0.2000    &0.0343    &0.0174    &0.0592      &729 \\
&InceptionV3  \cite{SzegedyLJSRAEVR15}       &0.7913 &0.7840 &0.9342 &0.8525  &0.7999  &0.2087    &0.2160    &0.0658    &0.1475    &0.2001       &990 \\
&MobileNet  \cite{HowardZCKWWAA17corr}           &0.9750 &0.9993 &0.9692 &0.9840  &0.9467  &0.0250    &0.0007    &0.0308    &0.0160    &0.0533       &559 \\
&ResNet50 \cite{HeZRS16}               &0.8519 &0.8410 &0.9616 &0.8973  &0.8646  &0.1481    &0.1590    &0.0384    &0.1027    &0.1354       &776 \\\hline

\multirow{6}{*}{\rotatebox{90}{\textbf{S.T.Camp. \cite{LuoLG17}}}}
&DenseNet121 \cite{HuangLMW17}           &0.9740 &0.9800 &0.9674 &0.9834  &0.9434  &0.0260    &0.0200    &0.0326    &0.0166    &0.0566       &491 \\
&VGGNet16 \cite{SimonyanZ14aVGG16}                &0.9522 &0.9800 &0.9331 &0.9654  &0.9283  &0.0478    &0.0200    &0.0669    &0.0346    &0.0717      &492 \\
&VGGNet19 \cite{SimonyanZ14aVGG16}                &0.9530 &0.9800 &0.9341 &0.9660  &0.9295  &0.0470    &0.0200    &0.0659    &0.0340    &0.0705      &467 \\
&InceptionV3 \cite{SzegedyLJSRAEVR15}      &0.9398 &0.9800 &0.9274 &0.9623  &0.8696  &0.0602    &0.0200    &0.0726    &0.0377    &0.1304       &866 \\
&MobileNet  \cite{HowardZCKWWAA17corr}           &0.9773 &0.9800 &0.9713 &0.9855  &0.9508  &0.0227    &0.0200    &0.0287    &0.0145    &0.0492       &465 \\
&ResNet50 \cite{HeZRS16}               &0.9523 &0.9800 &0.9553 &0.8719  &0.8922  &0.0477    &0.0200    &0.0447    &0.1281    &0.1078       &789 \\\hline

\multirow{6}{*}{\rotatebox{90}{\textbf{U.Crime \cite{SultaniCS18}}}}
&DenseNet121 \cite{HuangLMW17}           &0.8033 &0.9627 &0.7743 &0.8583  &0.8371  &0.1967    &0.0373    &0.2257    &0.1417    &0.1629      &595 \\
&VGGNet16 \cite{SimonyanZ14aVGG16}                &0.7444 &0.9532 &0.7022 &0.8087  &0.7936  &0.2556    &0.0468    &0.2978    &0.1913    &0.2064       &481 \\
&VGGNet19 \cite{SimonyanZ14aVGG16}                &0.7290 &0.9512 &0.6827 &0.7949  &0.7830  &0.2710    &0.0488    &0.3173    &0.2051    &0.2170     &470 \\
&InceptionV3  \cite{SzegedyLJSRAEVR15}       &0.4101 &0.7981 &0.3123 &0.4488  &0.5244  &0.5899    &0.2019    &0.6877    &0.5512    &0.4756      &648 \\
&MobileNet  \cite{HowardZCKWWAA17corr}           &0.6815 &0.9389 &0.6268 &0.7517  &0.7454  &0.3185    &0.0611    &0.3732    &0.2483    &0.2546      &473 \\
&ResNet50 \cite{HeZRS16}               &0.7523 &0.9000 &0.7087 &0.8145  &0.8799  &0.2477    &0.1000    &0.2913    &0.1855    &0.1201       &775 \\\hline

\end{tabular}
\label{ExperimentalResults2DCNN}
\end{center}
\end{table*}

\subsection{Average Ranking of 2DCNN Models}
The Friedman test \cite{Friedman1937} and its derivatives (e.g., Iman--Davenport test \cite{Iman1980}) are usually referred to as the most well-known non-parametric tests for multiple comparisons. Consequently, we performed the  Friedman test for our research\cite{Friedman1937}. The Friedman test \cite{Friedman1937} characteristically takes measures in preparation for ranking the performance of a set of algorithms in descending order. Notwithstanding, it can solely inform us about the appearance of differences among all samples of results under comparison. Its alternatives, e.g., Friedman's aligned rank test \cite{Hodges1962} and the Quade test \cite{Quade1979}, can provide us with further information, and they express opposition through rankings. They also provide better results based on the features of a given experimental study.

To achieve the non-parametric statistical test results from our experimental results, we applied Friedman \cite{Friedman1937}, Friedman's aligned rank \cite{Hodges1962}, and Quade \cite{Quade1979}  tests to our obtained results in Table \ref{ExperimentalResults2DCNN}. The aim of applying the Friedman~\cite{Friedman1937}, Friedman's aligned rank \cite{Hodges1962}, and Quade \cite{Quade1979} non-parametric tests is to determine whether there are significant differences among various models considering the data in Table \ref{ExperimentalResults2DCNN}. These tests provide rankings for the models regarding each individual dataset, i.e., the best-performing model receives the highest rank of 1, the second-best algorithm obtains the rank of 2, and so on. The mathematical equations and further explanation of the non-parametric procedures of the Friedman~\cite{Friedman1937}, Friedman's aligned rank~\cite{Hodges1962}, and Quade~\cite{Quade1979} tests can be found in Quade~\cite{Quade1979} and  Westfall et al.~\cite{Westfall2004}.

Table \ref{AverageRanking2DCNN} shows the average ranking computed by using the Friedman \cite{Friedman1937}, Friedman's aligned rank \cite{Hodges1962}, and Quade~\cite{Quade1979} non-parametric statistical tests.
\begin{table*}
\begin{center}
\caption{Average ranking of each model using non-parametric statistical tests.}
\begin{tabular}{|c|c|c|c|c|c|c|c|c|c|c|c|c|c|c|c|c|c|c|c|c|c|}\hline
\multirow{3}{*}{Datasets/Statistics}&\multirow{3}{*}{Model} &\multicolumn{9}{c|}{Multiple Comparison Tests}  \\\cline{3-11}
&&\multicolumn{3}{c|}{Friedman Ranking \cite{Friedman1937}} &\multicolumn{3}{c|}{Aligned Friedman Ranking \cite{Hodges1962}} &\multicolumn{3}{c|}{Quade Ranking \cite{Quade1979}}  \\\cline{3-11}
&&Score &Statistics&p-value&Score &Statistics&p-value &Score &Statistics&p-value\\\hline

\multirow{6}{*}{\rotatebox{90}{\textbf{U. Ped1 \cite{ChanLV08}}}}
&DenseNet121 \cite{HuangLMW17}           &1.8333
&\multirow{6}{*}{\rotatebox{90}{24.095238}} \multirow{6}{*}{\rotatebox{90}{$\chi^2$ with}}
\multirow{6}{*}{\rotatebox{90}{5 DOF}}
&\multirow{6}{*}{\rotatebox{90}{0.000208}} &16.0000
&\multirow{6}{*}{\rotatebox{90}{16.017387}} \multirow{6}{*}{\rotatebox{90}{$\chi^2$ with}}
\multirow{6}{*}{\rotatebox{90}{5 DOF}}
&\multirow{6}{*}{\rotatebox{90}{0.006795}} &2.1429
&\multirow{6}{*}{\rotatebox{90}{10.322425}}
\multirow{6}{*}{\rotatebox{90}{F-distribution}}
\multirow{6}{*}{\rotatebox{90}{with 5, 25 DOF}}
&\multirow{6}{*}{\rotatebox{90}{0.000019}}  \\
&VGGNet16 \cite{SimonyanZ14aVGG16}                 &2.5000 && &10.6667  && &2.4286   &&  \\
&VGGNet19 \cite{SimonyanZ14aVGG16}                 &1.8333 && &10.0000  && &1.7143   &&  \\
&InceptionV3 \cite{SzegedyLJSRAEVR15}           &5.8333 && &31.5000  && &5.7143   &&  \\
&MobileNet  \cite{HowardZCKWWAA17corr}          &4.3333 && &20.5000  && &4.5714   &&  \\
&ResNet50 \cite{HeZRS16}                        &4.6667 && &22.3333  && &4.4286   &&  \\\hline

\multirow{6}{*}{\rotatebox{90}{\textbf{U. Ped2 \cite{ChanLV08}}}}
&DenseNet121 \cite{HuangLMW17}           &4.0833 &\multirow{6}{*}{\rotatebox{90}{12.214286}}
\multirow{6}{*}{\rotatebox{90}{$\chi^2$ with}} \multirow{6}{*}{\rotatebox{90}{5 DOF}} &\multirow{6}{*}{\rotatebox{90}{0.031967}} &22.2500 &\multirow{6}{*}{\rotatebox{90}{7.435555}}
\multirow{6}{*}{\rotatebox{90}{$\chi^2$ with}} \multirow{6}{*}{\rotatebox{90}{5 DOF}}  &\multirow{6}{*}{\rotatebox{90}{0.190210}} &4.3810
&\multirow{6}{*}{\rotatebox{90}{3.099033}}
\multirow{6}{*}{\rotatebox{90}{F-distribution}}
\multirow{6}{*}{\rotatebox{90}{with 5, 25 DOF}}
&\multirow{6}{*}{\rotatebox{90}{0.025980}}  \\
&VGGNet16 \cite{SimonyanZ14aVGG16}                 &2.5833 && &13.0833  && &2.9048   &&  \\
&VGGNet19 \cite{SimonyanZ14aVGG16}                 &2.0000 && &12.8333  && &2.1429   &&  \\
&InceptionV3 \cite{SzegedyLJSRAEVR15}           &5.0000 && &26.5000  && &5.1429   &&  \\
&MobileNet  \cite{HowardZCKWWAA17corr}          &2.8333 && &15.6667  && &2.5238   &&  \\
&ResNet50 \cite{HeZRS16}                        &4.5000 && &20.6667  && &3.9048   &&  \\\hline

\multirow{6}{*}{\rotatebox{90}{\textbf{UMN  \cite{UMNdataset2021}}}}
&DenseNet121 \cite{HuangLMW17}           &5.5000 &\multirow{6}{*}{\rotatebox{90}{14.190476}}
\multirow{6}{*}{\rotatebox{90}{$\chi^2$ with}} \multirow{6}{*}{\rotatebox{90}{5 DOF}}
&\multirow{6}{*}{\rotatebox{90}{0.014444}} &30.5000 &\multirow{6}{*}{\rotatebox{90}{13.917021}}
\multirow{6}{*}{\rotatebox{90}{$\chi^2$ with}} \multirow{6}{*}{\rotatebox{90}{5 DOF}}
&\multirow{6}{*}{\rotatebox{90}{0.016145}} &5.2381
&\multirow{6}{*}{\rotatebox{90}{2.856555}}
\multirow{6}{*}{\rotatebox{90}{F-distribution}}
\multirow{6}{*}{\rotatebox{90}{with 5, 25 DOF}}
&\multirow{6}{*}{\rotatebox{90}{0.035686}}  \\
&VGGNet16 \cite{SimonyanZ14aVGG16}                 &2.3333 && &12.0000  && &2.9048   &&  \\
&VGGNet19 \cite{SimonyanZ14aVGG16}                 &3.1667 && &15.1667  && &3.4762   &&  \\
&InceptionV3 \cite{SzegedyLJSRAEVR15}           &4.1667 && &23.1667  && &4.0476   &&  \\
&MobileNet  \cite{HowardZCKWWAA17corr}          &3.8333 && &19.6667  && &3.4762   &&  \\
&ResNet50 \cite{HeZRS16}                        &2.0000 && &10.5000  && &1.8571   &&  \\\hline

\multirow{6}{*}{\rotatebox{90}{\textbf{C.Avenue \cite{LuSJ13}}}}
&DenseNet121 \cite{HuangLMW17}           &2.1667 &\multirow{6}{*}{\rotatebox{90}{19.047619}}
\multirow{6}{*}{\rotatebox{90}{$\chi^2$ with}} \multirow{6}{*}{\rotatebox{90}{5 DOF}}  &\multirow{6}{*}{\rotatebox{90}{0.001883}} &17.1667 &\multirow{6}{*}{\rotatebox{90}{14.324476}}
\multirow{6}{*}{\rotatebox{90}{$\chi^2$ with}} \multirow{6}{*}{\rotatebox{90}{5 DOF}}  &\multirow{6}{*}{\rotatebox{90}{0.013675}} &2.8571
&\multirow{6}{*}{\rotatebox{90}{5.512877}}
\multirow{6}{*}{\rotatebox{90}{F-distribution}}
\multirow{6}{*}{\rotatebox{90}{with 5, 25 DOF}}
&\multirow{6}{*}{\rotatebox{90}{0.001485}}  \\
&VGGNet16 \cite{SimonyanZ14aVGG16}                 &2.5000 && &13.6667  && &2.7619   &&  \\
&VGGNet19 \cite{SimonyanZ14aVGG16}                 &3.6667 && &15.5000  && &3.4286   &&  \\
&InceptionV3 \cite{SzegedyLJSRAEVR15}           &6.0000 && &31.5000  && &6.0000   &&  \\
&MobileNet  \cite{HowardZCKWWAA17corr}          &2.3333 && &10.5000  && &1.9524   &&  \\
&ResNet50 \cite{HeZRS16}                        &4.3333 && &22.6667  && &4.0000   &&  \\\hline

\multirow{6}{*}{\rotatebox{90}{\textbf{S.T.Camp. \cite{LuoLG17}}}}
&DenseNet121 \cite{HuangLMW17}           &2.4167 &\multirow{6}{*}{\rotatebox{90}{18.452381}}
\multirow{6}{*}{\rotatebox{90}{$\chi^2$ with}} \multirow{6}{*}{\rotatebox{90}{5 DOF}}
&\multirow{6}{*}{\rotatebox{90}{0.002430}} &30.5000 &\multirow{6}{*}{\rotatebox{90}{13.917021}}
\multirow{6}{*}{\rotatebox{90}{$\chi^2$ with}} \multirow{6}{*}{\rotatebox{90}{5 DOF}}
&\multirow{6}{*}{\rotatebox{90}{0.016145}} &2.3571
&\multirow{6}{*}{\rotatebox{90}{12.911119}}
\multirow{6}{*}{\rotatebox{90}{F-distribution}}
\multirow{6}{*}{\rotatebox{90}{with 5, 25 DOF}}
&\multirow{6}{*}{\rotatebox{90}{0.000003}}  \\
&VGGNet16 \cite{SimonyanZ14aVGG16}                 &4.2500 && &12.0000  && &4.2143   &&  \\
&VGGNet19 \cite{SimonyanZ14aVGG16}                 &3.0833 && &15.1667  && &2.8810   &&  \\
&InceptionV3 \cite{SzegedyLJSRAEVR15}           &5.4167 && &23.1667  && &5.6429   &&  \\
&MobileNet  \cite{HowardZCKWWAA17corr}          &1.4167 && &19.6667  && &1.1190   &&  \\
&ResNet50 \cite{HeZRS16}                        &4.4167 && &10.5000  && &4.7857   &&  \\\hline

\multirow{6}{*}{\rotatebox{90}{\textbf{U.Crime \cite{SultaniCS18}}}}
&DenseNet121 \cite{HuangLMW17}           &1.6667 &\multirow{6}{*}{\rotatebox{90}{17.52381}}
\multirow{6}{*}{\rotatebox{90}{$\chi^2$ with}} \multirow{6}{*}{\rotatebox{90}{5 DOF}}
&\multirow{6}{*}{\rotatebox{90}{0.003606}} &12.6667 &\multirow{6}{*}{\rotatebox{90}{12.186329}}
\multirow{6}{*}{\rotatebox{90}{$\chi^2$ with}} \multirow{6}{*}{\rotatebox{90}{5 DOF}}
&\multirow{6}{*}{\rotatebox{90}{0.032322}} &1.9524 &\multirow{6}{*}{\rotatebox{90}{4.064339}}
\multirow{6}{*}{\rotatebox{90}{F-distribution}}
\multirow{6}{*}{\rotatebox{90}{with 5, 25 DOF}}
&\multirow{6}{*}{\rotatebox{90}{0.007750}}  \\
&VGGNet16 \cite{SimonyanZ14aVGG16}                 &2.8333 && &13.5000  && &2.9524   &&  \\
&VGGNet19 \cite{SimonyanZ14aVGG16}                 &3.3333 && &15.3333  && &3.0952   &&  \\
&InceptionV3 \cite{SzegedyLJSRAEVR15}           &5.8333 && &31.6667  && &5.7143   &&  \\
&MobileNet  \cite{HowardZCKWWAA17corr}          &4.3333 && &21.1667  && &4.0952   &&  \\
&ResNet50 \cite{HeZRS16}                        &3.0000 && &16.6667  && &3.1905   &&  \\\hline

\multirow{6}{*}{\rotatebox{0}{\textbf{$\sigma$}}}
&DenseNet121 \cite{HuangLMW17}      &1.5217 &\multirow{6}{*}{\rotatebox{90}{0}}
&\multirow{6}{*}{\rotatebox{90}{0}} &7.6107 &\multirow{6}{*}{\rotatebox{90}{0}}
&\multirow{6}{*}{\rotatebox{90}{0}} &1.3445 &\multirow{6}{*}{\rotatebox{90}{0}}
&\multirow{6}{*}{\rotatebox{90}{0}}  \\
&VGGNet16 \cite{SimonyanZ14aVGG16}                 &0.7130 && &1.1455  && &0.6121   &&  \\
&VGGNet19 \cite{SimonyanZ14aVGG16}                 &0.7499 && &2.1960  && &0.7150   &&  \\
&InceptionV3 \cite{SzegedyLJSRAEVR15}           &0.6947 && &4.1683  && &0.7083   &&  \\
&MobileNet  \cite{HowardZCKWWAA17corr}          &1.1860 && &4.0844  && &1.3226   &&  \\
&ResNet50 \cite{HeZRS16}                        &1.0755 && &5.6270  && &1.0485   &&  \\\hline

\multirow{6}{*}{\rotatebox{0}{\textbf{Mean}}}
&DenseNet121 \cite{HuangLMW17}           &2.9444 &\multirow{6}{*}{\rotatebox{90}{17.5873}}
&\multirow{6}{*}{\rotatebox{90}{0.0091}} &21.5139 &\multirow{6}{*}{\rotatebox{90}{12.9663}}
&\multirow{6}{*}{\rotatebox{90}{0.0459}} &3.1548 &\multirow{6}{*}{\rotatebox{90}{6.46110}}
&\multirow{6}{*}{\rotatebox{90}{0.0118}}  \\
&VGGNet16 \cite{SimonyanZ14aVGG16}                 &2.8333 && &12.4861  && &3.0278   &&  \\
&VGGNet19 \cite{SimonyanZ14aVGG16}                 &2.8472 && &14.0000  && &2.7897   &&  \\
&InceptionV3 \cite{SzegedyLJSRAEVR15}           &5.3750 && &27.9167  && &5.3770   &&  \\
&MobileNet  \cite{HowardZCKWWAA17corr}          &3.1805 && &17.8611  && &2.9563   &&  \\
&ResNet50 \cite{HeZRS16}                        &3.8194 && &17.2222  && &3.6944   &&  \\\hline

\multirow{6}{*}{\rotatebox{0}{\textbf{$\sigma$ of Mean}}}
&DenseNet121 \cite{HuangLMW17}      &\multirow{6}{*}{\rotatebox{90}{0.9899}} &\multirow{6}{*}{\rotatebox{90}{0}}
&\multirow{6}{*}{\rotatebox{90}{0}} &\multirow{6}{*}{\rotatebox{90}{2.3241}} &\multirow{6}{*}{\rotatebox{90}{0}}
&\multirow{6}{*}{\rotatebox{90}{0}} &\multirow{6}{*}{\rotatebox{90}{0.3261}} &\multirow{6}{*}{\rotatebox{90}{0}}
&\multirow{6}{*}{\rotatebox{90}{0}}  \\
&VGGNet16 \cite{SimonyanZ14aVGG16}                 & && &   && &    &&  \\
&VGGNet19 \cite{SimonyanZ14aVGG16}                 & && &   && &    &&  \\
&InceptionV3 \cite{SzegedyLJSRAEVR15}           & && &   && &    &&  \\
&MobileNet  \cite{HowardZCKWWAA17corr}          & && &   && &    &&  \\
&ResNet50 \cite{HeZRS16}                        & && &   && &    &&  \\\hline

\end{tabular}
\label{AverageRanking2DCNN}
\end{center}
\end{table*}
DenseNet121~\cite{HuangLMW17} achieved the best scores from the UCF-Crime~\cite{SultaniCS18} dataset and comparable results from the UCSD Ped1~\cite{ChanLV08} and CUHK-Avenue \cite{LuSJ13} datasets, whereas ResNet50~\cite{HeZRS16} attained the best scores from the UMN~\cite{UMNdataset2021} dataset.
Using the Friedman test \cite{Friedman1937}, the ranking scores of DenseNet121 \cite{HuangLMW17} from six different datasets were clustered within $\sigma=1.5217$ of the mean, whereas for ResNet50 \cite{HeZRS16}, the same was $\sigma=1.0755$. Specifically, the scores of different datasets in ResNet50 \cite{HeZRS16} were 29\% less spread out than those of DenseNet121 \cite{HuangLMW17}.
Similarly, ResNet50 \cite{HeZRS16} acquired 26\% and 22\% better condensed ranking score distributions compared with its counterpart DenseNet121 \cite{HuangLMW17} considering the aligned Friedman~\cite{Hodges1962} and Quade tests \cite{Quade1979}, respectively.
DenseNet121 \cite{HuangLMW17} concentrates on making the deep learning networks move even deeper as well as simultaneously making them well-organized for training by applying shorter connections among the layers. It requires fewer parameters and allows feature reuse.
The key base element of ResNet50~\cite{HeZRS16} is the residual block. ResNet50 \cite{HeZRS16} adopts summation, whereas DenseNet121 \cite{HuangLMW17} deals with concatenation. Nevertheless, the dense concatenation of DenseNet121 \cite{HuangLMW17} creates the challenges of a demanding high-GPU memory and more training time \cite{ZhangBALKRBK21}. On the other hand, the identity shortcut that balances training in ResNet50 \cite{HeZRS16} curbs its representation dimensions \cite{ZhangBALKRBK21}. Compendiously, there is a dilemma between ResNet50 \cite{HeZRS16} and DenseNet121 \cite{HuangLMW17} for many applications in terms of the performance and GPU resources \cite{ZhangBALKRBK21}.

InceptionV3 \cite{SzegedyLJSRAEVR15} is 48 layers deep. It reached comparable scores with the UCSD Ped1 \cite{ChanLV08}, UCSD Ped2 \cite{ChanLV08}, UMN~\cite{UMNdataset2021}, and ShanghaiTech Campus~\cite{LuoLG17}  datasets.
MobileNet~\cite{HowardZCKWWAA17corr} was based on a streamlined architecture that applied depth-wise separable convolutions to build a lightweight deep neural network. It gained the best scores from the CUHK-Avenue \cite{LuSJ13} dataset, whereas it performed excellently for the ShanghaiTech Campus \cite{LuoLG17} dataset.
InceptionV3 \cite{SzegedyLJSRAEVR15} received a 41\% and 47\% better condensed distribution of ranking scores compared with MobileNet  \cite{HowardZCKWWAA17corr} after taking the Friedman \cite{Friedman1937} and Quade tests \cite{Quade1979}, respectively.
However, MobileNet  \cite{HowardZCKWWAA17corr}  attained a 2\%  better condensed distribution of ranking scores compared with InceptionV3 \cite{SzegedyLJSRAEVR15} from the aligned Friedman test \cite{Hodges1962}.

\begin{figure*}
\centering
\includegraphics[width=0.934\textwidth]{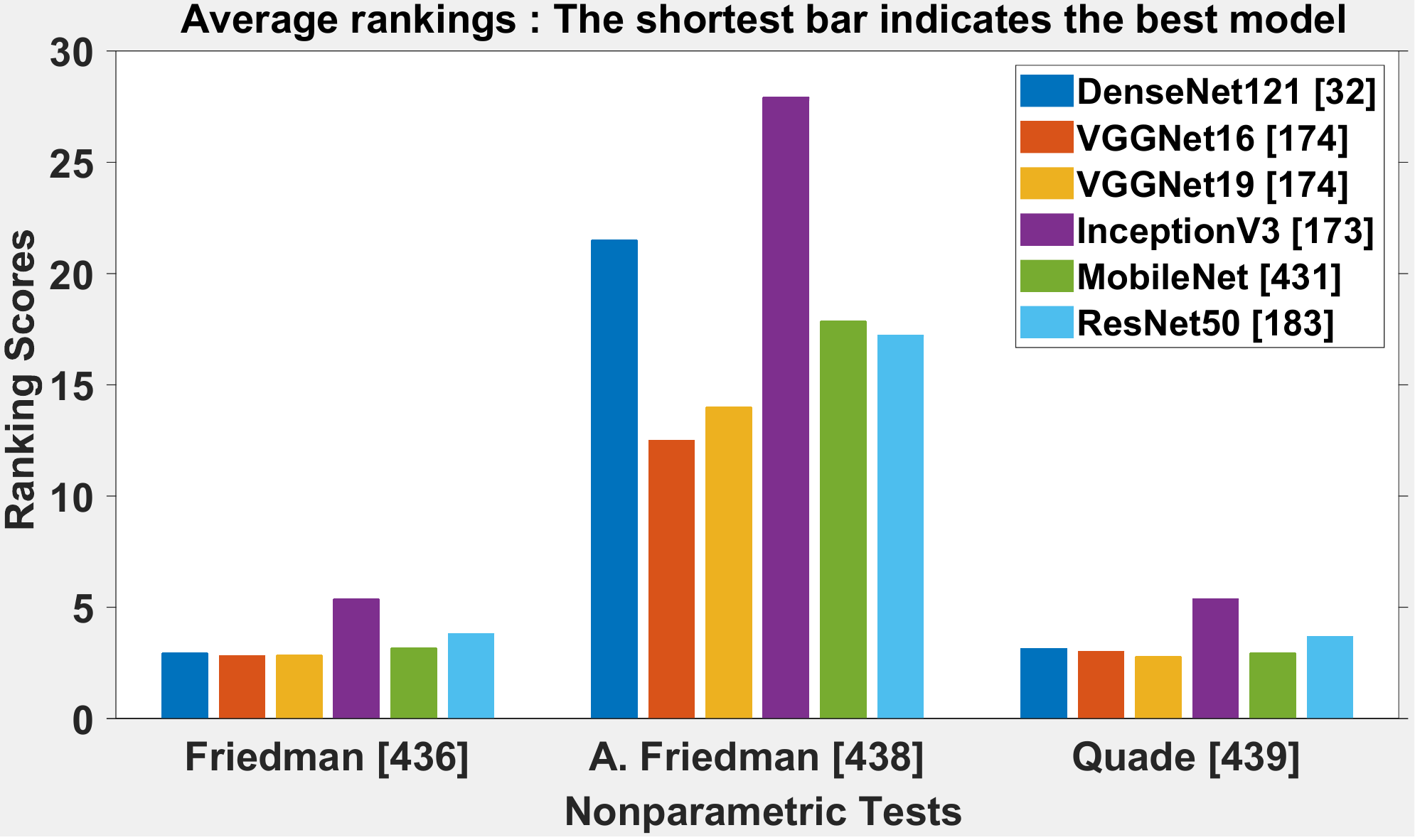}
\caption{Average ranking using the \textit{mean} values in Table \ref{AverageRanking2DCNN}.}
\label{MeanRankings}
\end{figure*}

There are various variants of VGGNet \cite{SimonyanZ14aVGG16} (e.g., VGG16 and VGG19) that differ exclusively in the total number of layers in the network. VGGNet16 \cite{SimonyanZ14aVGG16} did not come into the best scores from  any dataset, whereas VGGNet19~\cite{SimonyanZ14aVGG16} achieved the best scores from  the UCSD Ped1~\cite{ChanLV08} and UCSD Ped2~\cite{ChanLV08} datasets. However, VGGNet16 \cite{SimonyanZ14aVGG16} performed better than VGGNet19~\cite{SimonyanZ14aVGG16} using the UCF-Crime~\cite{SultaniCS18}, CUHK-Avenue~\cite{LuSJ13}, and UMN~\cite{UMNdataset2021} datasets.
VGGNet16~\cite{SimonyanZ14aVGG16} achieved 5\%, 48\%, and 14\% higher condensed distributions of ranking scores compared with VGGNet19 \cite{SimonyanZ14aVGG16} in the Friedman \cite{Friedman1937}, aligned Friedman~\cite{Hodges1962}, and Quade tests~\cite{Quade1979}, respectively.
With an increasing number of layers in the CNN model, the potential for the model to fit more sophisticated functions rises. Accordingly, more layers result in better CNN performance. The pre-trained 2DCNN models of VGG16 and VGG19 accommodate 16 and 19 layers, respectively. Compared with VGGNet16 \cite{SimonyanZ14aVGG16}, VGGNet19 \cite{SimonyanZ14aVGG16} is a considerably large neural network in terms of the number of parameters to be trained. Nevertheless, Figure \ref{MeanRankings} shows that the VGG16 and VGG19 models demonstrated razor-thin performance differences. Although VGGNet is painfully slow to train, many studies were carried out on VGG16 in lieu of VGG19 (see Table \ref{SummaryOfLiteratureReview2020to2021}).

Figure \ref{MeanRankings} depicts the average ranking upon applying the mean values from Table \ref{AverageRanking2DCNN}.
The Friedman \cite{Friedman1937} and aligned Friedman tests \cite{Hodges1962} hinted that VGGNet16~\cite{SimonyanZ14aVGG16} would be the best model among its counterparts, whereas the Quade test~\cite{Quade1979} preferred VGGNet19 \cite{SimonyanZ14aVGG16}. In practice, the differences in the various model are not very significant.
Furthermore, it is noticeable that none of the aforementioned 2DCNN models were able to show their continuous superiority over all datasets.

From the outset, the $\sigma$ of mean value 0 stipulates that all employed 2DCNN models should present identical performance scores from all datasets. Practically, the $\sigma$ of the mean values, i.e., 0.9899, 2.3241, and 0.3261 secured by the Friedman~\cite{Friedman1937}, aligned Friedman \cite{Hodges1962}, and Quade tests \cite{Quade1979}, respectively, cumulatively specify that the used 2DCNN models showed non-identical but almost similar performance scores from different datasets. Consequently, pre-trained 2DCNN architectural variations play an unimportant role in detecting crowd video anomalies.
Furthermore, the computation of a higher number of parameters is time-consuming in the sophisticated model. A less-sophisticated model uses comparatively fewer numbers of parameters. However, it gains nearly the same accuracy as many more sophisticated models. Based on the applications of 2DCNN models, there is a trade-off between two options: computational complexity and accuracy.

\subsection{Our Observations}
CNNs are the preferred option for computer vision applications. They undertake pixel values to output numerous visual features. Does it matter if the simple pre-trained CNN models provide slightly less accuracy by computing a lesser number of parameters than the complex pre-trained 2DCNN models? The aforesaid rigorous statistical analysis evinces that the architectural dissimilarities of the pre-trained 2DCNN models have an insignificant influence on crowd video anomaly detection. As a consequence, a lesser-layered CNN model can achieve approximately the same performance
as a more-layered CNN model and reduce computational time. For example, VGGNet16 \cite{SimonyanZ14aVGG16} can be used instead of VGGNet19 \cite{SimonyanZ14aVGG16}, saving 3.7\% of parameters.
Although VGGNet19 \cite{SimonyanZ14aVGG16} can be utilized to extract motion features from complex and noisy surveillance scenes, and showed the best mean performance in our experimental set-up, we do not recommend its use for small-size datasets as it demands a large set of training samples~\cite{Al-DhamariSM20}.

\section{Open Challenges and Future Prospects} \label{FutureProspectResearchChallenge}
The subjectivity of anomaly definition, the rarity of anomalies, big datasets, and high computational power make crowd anomaly detection a critical and challenging task. Deep learning enables the possibility of automatic feature extraction. Thus, deep learning-based solutions for crowd anomaly detection have performed significantly better than traditional solutions in terms of complexity and accuracy. However, they encounter difficulties when devising reliable approaches that can be applied to real-world problems  \cite{MohammadiEtAl2021corr}. Explicitly, various challenges exist in this research area. Therefore, here, we briefly review several remarkable facets, which could be minimized in the future.

\begin{itemize}
  \item \textbf{Lack of anomaly definition} $\Rightarrow$  Anomaly definition is fully subjective. Based on the time and place, the same event can be either normal or abnormal. The most popular datasets (e.g., UCSD \cite{ChanLV08}, CUHK-Avenue \cite{LuSJ13}, and ShanghaiTech Campus \cite{LuoLG17}) assume that whatever is unseen in the training data is marked as anomalous. In effect, this causes a very restricted nominal class and an apparent depiction of an anomaly \cite{DoshiEtAlwacv2022a}. As anomaly detection is a fault-finding task, the definition of anomaly should be flawlessly rational in the corresponding contexts.
      Can anomalies in crowds be reliably identified from CCTV footage without a prior definition of anomalies and the need for the extraction of handcrafted features? This is one of the biggest challenges for crowd anomaly detection.

  \item \textbf{Lack of realistic datasets availability} $\Rightarrow$  Deep learning methods demand a large number of datasets for training. However, the existing datasets are not sufficient to perform accurate training or testing.
   It is important to build a dataset with a larger amount of data and a wider behavior categorization. In addition to data collection, ensuring correct annotation is a challenging task for a bigger video dataset. The availability of video datasets is also a big issue. Few datasets are publicly accessible, while many datasets are still not available for open research. In addition, publicly available crowd datasets suffer from certain restrictions.  For example, the UMN~\cite{UMNdataset2021} dataset is straightforward where the performance of methods is saturated on it. Furthermore, videos in Ped1--Ped2 of the UCSD \cite{ChanLV08} datasets are captured in just one location, and hence the camera is fixed during the training procedure. Additionally, the resolution of the video frames is extremely low.
   In addition, abnormal scenarios are sensitive to obtain and use.
   For example, the usage of crime-related videos is strictly regulated by governments or authorities.

  \item \textbf{Lack of ground truth annotation} $\Rightarrow$
  Some existing datasets only provide a frame-level ground truth annotation. Henceforth, without performing P-AUC and P-EER, we have to satisfy F-AUC and F-EER. For example, the UMN \cite{UMNdataset2021} dataset does not provide the pixel-level ground truth annotation. Therefore, P-AUC and P-EER cannot be performed \cite{HanWYWKDL20}. Furthermore, due to the unavailability of labeled data, it is very arduous to create a benchmark for anomaly explanation \cite{SzymanowiczCC21}.

  \item \textbf{Lack of hardware applications} $\Rightarrow$ In addition to methodology breakthroughs and accessible big-data training, successful solutions for video anomaly detection are also due to recent advances in hardware applications
  . Crowd anomaly detection demands the processing of a large amount of video data, which needs a powerful GPU. The accessible video dataset is unlimited, whereas the accessible hardware processing volume is in short supply. Therefore, there are challenges regarding hardware constraints. Currently, big artificial intelligence systems are notoriously difficult and expensive to train because the underlying hardware applications are not fast enough.

  \item \textbf{Lack of computing power} $\Rightarrow$
   The demand for artificial intelligence models is increasing  rapidly. Ten years ago, the largest models were some 10 million parameters, which might be trained in a few hours on a single GPU. However, today, the largest models are over 10 trillion parameters, which can take up to a year for training across tens of thousands of machines. Soon, highly sophisticated artificial intelligence systems could become the exclusive domain of companies and administrations. For various deep learning and large-scale analytics applications, power dissipation across the many components of the computing infrastructure is expected to be an order of magnitude higher than in the current systems. Data are sent along electrical wires in traditional hardware. Nevertheless, electrical wires consume far more energy and transfer far less data over longer distances. A solution to this problem is luminous computing, which is developed on light-based artificial intelligence accelerator chips \cite{Angelini2022}. In theory, such light-based chips may lead to higher performance levels because light produces less heat than electricity. Furthermore, light can propagate faster and is less susceptible to alternations in temperature and electromagnetic fields. However, light-based chips are physically larger than their electronic counterparts. In addition, their architectures still mainly rely on electronic control circuits, which can produce bottlenecks.

  \item \textbf{Lack of unsupervised learning} $\Rightarrow$  Similar to machine learning, deep learning methods can be grouped into supervised, semi-supervised, and unsupervised categories. Supervised learning methods are widely used in video anomaly detection. However, they entail the manual labeling of a large number of datasets. The exponential growth of data makes manual labeling full of challenges. Unsupervised and deep reinforcement learning both require more attention to realize automatic learning from unlabeled videos. However, training deep neural networks should be humanly understandable.

 \item \textbf{Lack of static backgrounds} $\Rightarrow$ Surveillance cameras can capture videos in various dynamic backgrounds, which are subject to illumination variation, occlusions, and viewpoint alternation. Analyzing such videos is a challenging task. Rainy, sunny, and snowy conditions can often cause anomalies. These environmental conditions are closely related to background features.
     Model accuracy can affect the neglected background features. Ideally, both foreground and background features can be considered by the anomaly detection models.

  \item \textbf{Lack of high-quality videos} $\Rightarrow$  Religious events, airport arrivals, and departure terminals are busy places where occlusions happen very frequently. Due to the long distances of cameras, the subjects are relatively small, producing poor-quality videos. The relatively low quality of long-distance videos makes the detection process more challenging.

 \item \textbf{Lack of frequency of normal events} $\Rightarrow$ For crowd anomaly detection, the false positive rate should be as low as possible. To discover a method with a high detection rate along with a low false positive rate is a considerable area of research.

\item \textbf{Lack of model direct applicability} $\Rightarrow$ Some deep learning models cannot be directly applicable enough for real-world applications. For example,  GAN \cite{GoodfellowPMXWOCB14}  is used for generating anomalous data but its inability to generate anomalies has been highlighted in the literature. It assumes that a previously unseen activity creates a higher prediction error \cite{DoshiEtAlwacv2022a}. Most of its generated data are in imitation of random noises. Accordingly, instead of directly using it, the anomaly detection problem can be transferred into a binary classification problem \cite{PourRezaMKBSS21}.

\item \textbf{Lack of direct training capability} $\Rightarrow$
   Some GAN-based deep learning models are dedicated to one-class classification tasks. The GAN discriminator can be utilized as a deformity detector. However, such solutions need trial and error to decipher the potential awkwardness during the training methods.

\item \textbf{Lack of fairness} $\Rightarrow$  The factors of skewed samples, limited features, tainted examples, and disparities of sample sizes and proxies can bias the training sets \cite{BolukbasiCZSK16}, resulting in an unfair output from a trained deep neural network. For example, Zhang et al. \cite{ZhangD21} investigated fairness for the anomaly detection problem and reported that the deep-support-vector data description model \cite{RuffGDSVBMK18} failed in fairness evaluation.

\item \textbf{Lack of assurance} $\Rightarrow$ Deep neural networks are vulnerable to adversarial attacks, which do not signify detecting abnormal events \cite{MohammadiEtAl2021corr}.

\item \textbf{Lack of lengthy video segments} $\Rightarrow$
The contemporaneous state-of-the-art methods inherently assume that each test video segment comprises anomalous activity. To meet this assumption, the length of video segments would need to be exceedingly long because, in real-life scenes, abnormal events only occasionally take place. In practice, the video segments in most of the existing benchmark datasets are a few minutes long and consistently labeled by abnormal frames, which cannot undoubtedly keep in touch with the real-life abnormalities. Therefore, they make false alarms in real-world synopses \cite{DoshiEtAlwacv2022a}.

\item \textbf{Lack of negative value count} $\Rightarrow$
The ReLU cuts off negative values. Hence, it can limit diverse feature portrayals~\cite{ParkNH20}. For example,  Park et al. \cite{ParkNH20} and Szymanowicz et al.~\cite{SzymanowiczEtAlwacv2022} provided solutions to this problem by removing the last BN~\cite{IoffeS15} and ReLU layers \cite{KrizhevskySH12} in their encoders to minimize this effect.

\item \textbf{Lack of temporality richness} $\Rightarrow$
A crystal-clear abnormal event in one scene cannot be recognized as an alike activity that can take the shape of a normal event in a completely dissimilar scene of a significantly larger and more complex multi-scene dataset. The sheer heterogeneity in the multi-scene dataset creates mismatched issues~\cite{RamachandraJ20}. Instead of comprehensive data over various scenes, comprehensive data over time can minimize this problem~\cite{DoshiEtAlwacv2022a}.

\item \textbf{Lack of model understandability} $\Rightarrow$
The ability of a system to automatically detect anomalous events and to recover humanly readable explanations for detected anomalies is very important~\cite{RadSJD21}. Researchers interpreted deep models to explain the outcomes of deep neural networks. In reality, refraining from taking the deep neural network as a black-box approach brings about a better understandability and reliability~\cite{HolzingerKWT18}. 

\item \textbf{Lack of solvable mathematical paradox} $\Rightarrow$
As a comparison to the limitations of intelligence for both humans and machines, we observe that humans are frequently effective at recognizing when they get things erroneous. However, deep learning models are not. Although deep learning is the leading artificial intelligence technology for pattern recognition, many deep learning models are untrustworthy and easy to fool. Many deep models do not realize when they make errors.
Sometimes, it is even more laborious for a deep model to know when it is making an error than to generate a true result. Alan Turing and Kurt G\"{o}del, two notable mathematicians, discovered a paradox at the heart of mathematics, namely that it is impracticable to show whether certain mathematical statements are true or false; in addition, some computational problems cannot be addressed with algorithms. Regardless of the data accuracy, we cannot obtain perfect information for constructing a required neural network, and, generally, deep learning models suffer from inherent limitations due to this century-old mathematical paradox \cite{CambridgeUni2022}.

\item \textbf{Lack of anonymization} $\Rightarrow$
Some jurisdictions have very strict privacy laws which preclude the application of any automated video analytics on raw video data.
Anonymization is a practical solution to preserve privacy \cite{MajeedL21} as it removes or modifies personally identifiable information suitable for utilizing in research and data mining.
Generally, anonymization removes any identifying features such as faces \cite{WereszczynskiMS17,DingHLNW022}, clothing, accessories, objects carried by individuals, etc. from video feeds is thus imperative prior to any further processing. However, it must be performed in such a way that both preserves mandated privacy and does not discard salient video features necessary for anomaly detection. The extent to which a trade-off between anomaly detection performance and level of anonymization may exist remains an open question. Since both anonymization and determining salient features are essentially abstraction operations, the ideal approach would directly use the result of anonymization as input features to automatic crowd anomaly detection. Although anonymization may be an imperative, it does not, on its own, guarantee general acceptance of real-time surveillance in private and public spaces.

\end{itemize}

\section{Conclusion}\label{Conclusion}
We were motivated to produce this comprehensive survey of deep learning parameters by the lack of contemporary research regarding advanced deep learning-based crowd anomaly detection methods for video clips. 
 Our review revealed several novel facts regarding datasets, taxonomy, and deep model architectures, along with their performances. We found that CNNs were the de facto model of choice for computer vision procedures. 
  Furthermore, we explored open research challenges to explore prospects for future study.

\section*{Acknowledgments}
This work is a part of the AI4CITIZENS research project (number 320783) supported by the Research Council of Norway.


\bibliographystyle{IEEEtran}
\bibliography{main}

\vfill

\end{document}